\documentclass{article}

\usepackage[final]{neurips_2025}

\usepackage[utf8]{inputenc} 
\usepackage[T1]{fontenc}    
\usepackage{hyperref}       
\usepackage{url}            
\usepackage{booktabs}       
\usepackage{amsfonts}       
\usepackage{amsmath}        
\usepackage{nicefrac}       
\usepackage{microtype}      
\usepackage{xcolor}         
\usepackage{natbib}
\usepackage{graphicx}
\usepackage{subcaption}
\usepackage{float}     
\usepackage{placeins}  
\usepackage{mathtools}

\title{Emergent World Beliefs: Exploring Transformers in Stochastic Games}

\author{
\begin{tabular}{ccc}
{\bfseries Adam Kamel}\thanks{* Equal contribution.} &
{\bfseries Tanish Rastogi}\footnotemark[1] &
{\bfseries Michael Ma}\footnotemark[1] \\
{\normalfont University of Waterloo} &
{\normalfont Independent Researcher} &
{\normalfont Independent Researcher} \\
\texttt{atkamel@uwaterloo.ca} &
\texttt{tanishrastogi2020@gmail.com} &
\texttt{michaelma0311@gmail.com}
\end{tabular}
\\[1.5em]
\begin{tabular}{cc}
{\bfseries Kailash Ranganathan}\thanks{\dag\ Senior author.}\normalfont &
{\bfseries Kevin Zhu}\footnotemark[2]\normalfont \\
University of California, Berkeley &
Algoverse AI Research \\
\texttt{kranganathan@berkeley.edu} &
\texttt{kevin@algoverse.us}
\end{tabular}
}

\begin{document}
\maketitle

\begin{abstract}
Transformer-based large language models (LLMs) have demonstrated strong reasoning abilities across diverse fields, from solving programming challenges to competing in strategy-intensive games such as chess. Prior work has shown that LLMs can develop emergent world models in games of perfect information, where internal representations correspond to latent states of the environment. In this paper, we extend this line of investigation to domains of incomplete information, focusing on poker as a canonical partially observable Markov decision process (POMDP). We pretrain a GPT-style model on Poker Hand History (PHH) data and probe its internal activations. Our results demonstrate that the model learns both deterministic structure, such as hand ranks, and stochastic features, such as equity, without explicit instruction. Furthermore, by using primarily nonlinear probes, we demonstrated that these representations are decodeable and correlate with theoretical belief states, suggesting that LLMs are learning their own representation of the stochastic environment of Texas Hold'em Poker.
\end{abstract}

\section{Introduction}

Current transformer-based large language models (LLMs) have achieved breakthrough results across various tasks, ranging from answering industry programming questions to solving olympiad-level problems. \citep{PhysicsOly, livecodebench}. The ability of LLMs to complete these tasks lies in their advanced reasoning capabilities, which are extremely evident when playing reasoning-intensive games such as chess \citep{1788EloChessLLM}.

Despite these achievements in LLM reasoning capabilities, the internal execution of their strategies remains a "black-box". Recently, research on LLMs with \textit{internal world representations} has grown to demonstrate higher-level LLM understanding in games as seen in \citet{ChessFine-Tuned} and \citet{OthelloAnalysis}. The findings of \citet{OthelloAnalysis} demonstrate the OthelloGPT model's ability to develop its own internal representation of the game states and rules of Othello from move strings in a strictly deterministic, perfect-information setting, with the hope that natural-language models are learning broader semantic "world representations". In this paper, we extend this analysis to world models in games of incomplete information, in particular Poker, to explore how LLMs intrinsically model uncertainty in a Bayesian fashion and provide new insights into their decision-making process. 
\newpage
Our paper's main contributions include: 
\begin{itemize}
   \item We extend LLM internal world representation to games of incomplete information.
   \item We quantify the understanding of circuits/features that underlie an LLM's "belief state" for partially observable Markov Decision Processes (explored through Poker).
\end{itemize}
We defer a theoretical analysis of the second focus to appendix section \ref{subsec:theory}, and a further discussion on LLM world models to \ref{subsec:worldmodels}.

\section{Related Works}

This paper builds upon the work of \citet{ChessFine-Tuned}, \citet{OthelloAnalysis} and \citet{nanda2023emergentlinearrepresentationsworld} who trained language models to play complete information games such as Chess and Othello. \citet{OthelloAnalysis} demonstrated that OthelloGPT, a LLM trained on sequences of legal moves in Othello spontaneously developed an internal representation of the game board that could be extracted with nonlinear probes. Later it was demonstrated by \citet{nanda2023emergentlinearrepresentationsworld} that linear probes could extract this representation as well. \citet{ChessFine-Tuned} extended these findings from Othello to the game of Chess as well, and also demonstrated models that developed these world representations also had the capability to understand and estimate latent variables such as player skill. 

\section{Poker Model}

As a foundation for our studies on LLMs in POMDPs/stochastic games, we pretrain a GPT-style architecture on Poker games, using the Poker Hand History (PHH) format \citep{kim2024recordingdescribingpokerhands} (see Appendix \ref{phh-format}).
\subsection{Dataset}
As noted by past papers exploring emergent world representations such as \citet{ChessFine-Tuned}, dataset size plays a large role in probe results. Due to the unavailability of large and complete poker hand datasets, we opted to generate our own. We did this by utilizing a large number of game simulations to determine poker equity\citep{Billings1999SelectiveSampling} which then drives decision making. Our simulation script generates legal six-player No-Limit Texas Hold'em hands in PHH format. To start each hand, we give each player fresh stacks and randomly initialize their playing style to ensure that there is variation in agent behavior. Agents use a combination of simulation equity estimates and heuristics based on their randomly initialized playing style to make decisions, driving realistic and diverse poker games. 

\subsection{Training}

We fine-tuned a causal transformer language model based on GPT-2 \citep{GPT-2} using a PHH-formatted dataset comprising over two million synthetically generated poker trajectories, as described above. The model retained the GPT-2 base configuration, with 12 attention heads and a hidden dimensionality of 768. The GPT-2 base configuration was chosen due to it being lightweight to work with our limited compute resources while being well understood and robust.   Each hand was tokenized using a custom \texttt{PreTrainedTokenizerFast} vocabulary. During preprocessing, for a subset of tokens, we insert a reserved special \texttt{<GAP>} token in the input and shift the replaced original token to appear following a special \texttt{<ANS>} token later in the sequence. In training, we compute loss exclusively on the tokens that follow \texttt{<ANS>} to ensure proper loss calculation.

Optimization used AdamW with \(\beta_1=0.9\), \(\beta_2=0.95\), and \(\epsilon=10^{-8}\), with a commonly used learning rate of \(5\times 10^{-5}\). We trained for 13 epochs (training was paused after validation loss stopped improving), using an effective batch size of 128 (minibatch size 64 with gradient accumulation of 2). Checkpoints were saved every 5{,}000 steps and the best-performing checkpoint by validation loss was stored separately. We used a 95-5 train-test split for model training. See Appendix \ref{compute} for more training details.

\section{Probing Internal Representations}

We distinguish between two types of internal representations in our analysis: deterministic representations, which capture absolute aspects like hand rank and actions, and stochastic representations- such as equity- to extract the model’s internal belief state of the underlying Poker POMDP (\citet{shai2024beliefgeometry}). To capture these results, we probe internal activations using a linear classifier probe and a two-layer multilayer perceptron (MLP), a technique frequently used (\citep{OthelloAnalysis, HernandezAndreas2021LowDim}). The function of a linear probe used for our deterministic model is \( p_\theta(x_t^{l}) = \text{argmax}(W x_t^{l}) \), where \( \theta = \{ W \in \mathbb{R}^{C \times F} \} \), where \( F \) is the number of dimensions of the input activation vector \( x_t^{l} \). The function used for our two-layer MLP probe is \( p_\theta(x_t^{l}) = \text{argmax}(W_1 \, \text{ReLU}(W_2 x_t^{l})) \), where \( \theta = \{ W_1 \in \mathbb{R}^{C \times H}, W_2 \in \mathbb{R}^{H \times F} \} \), \(H\) is the number of hidden dimensions for the nonlinear probes, where \(C\) denotes the number of classes under consideration for identification (typically \(C = 4\) in our context). For our stochastic representation probes, \texttt{argmax} was not used as the output was continuous. Overall, the MLP probe achieved higher accuracy than the linear probe, consistent with the findings of \citet{OthelloAnalysis}.

\subsection{Deterministic World Model}


\begin{figure}[H]
    \centering

    \begin{subfigure}[t]{0.47\textwidth}
        \centering
        \includegraphics[width=\linewidth]{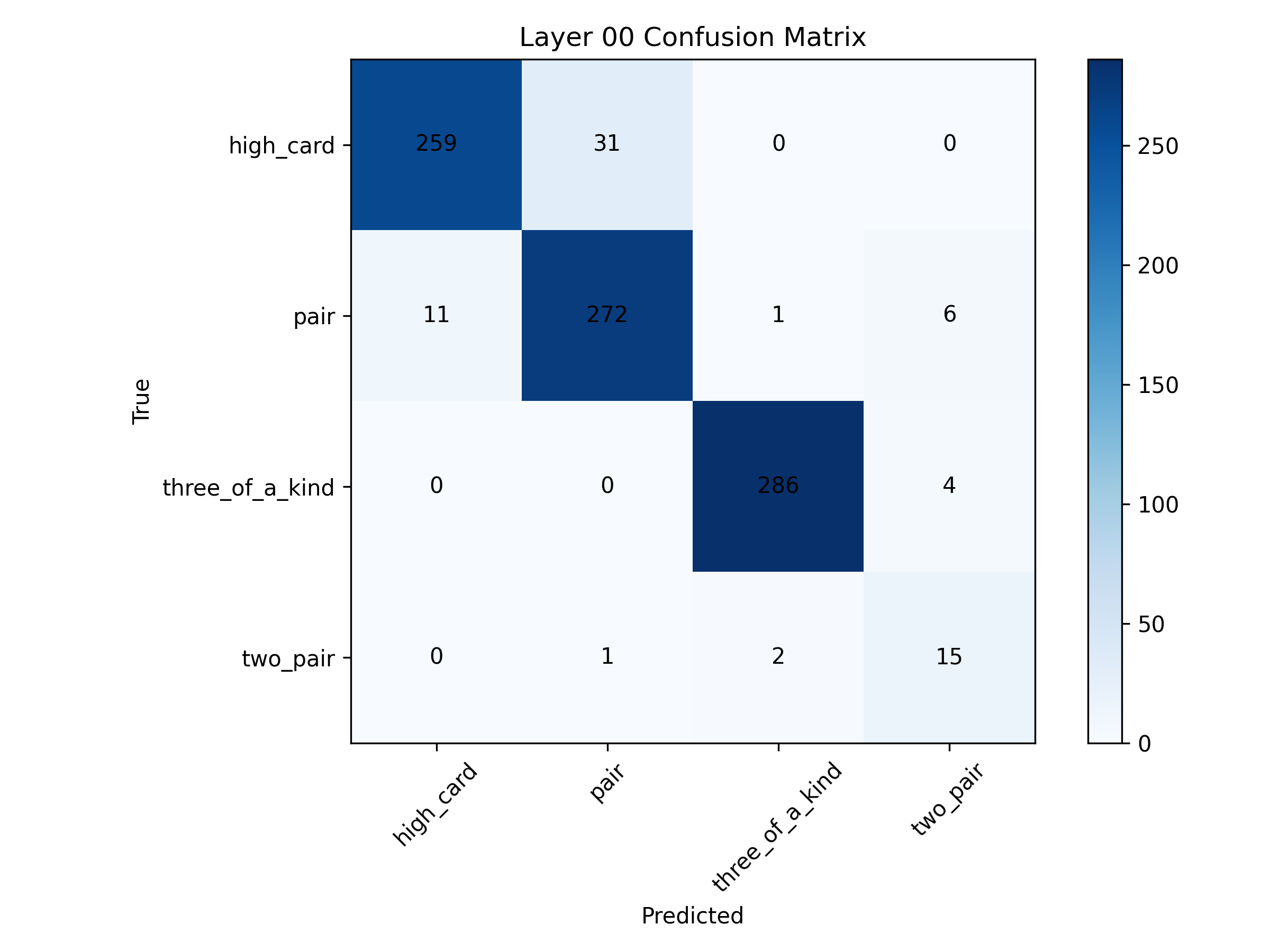}
        \caption{Layer 0 – 30th percentile equation.}
        \label{fig:mlp_30th}
    \end{subfigure}
    \hfill
    \begin{subfigure}[t]{0.47\textwidth}
        \centering
        \includegraphics[width=\linewidth]{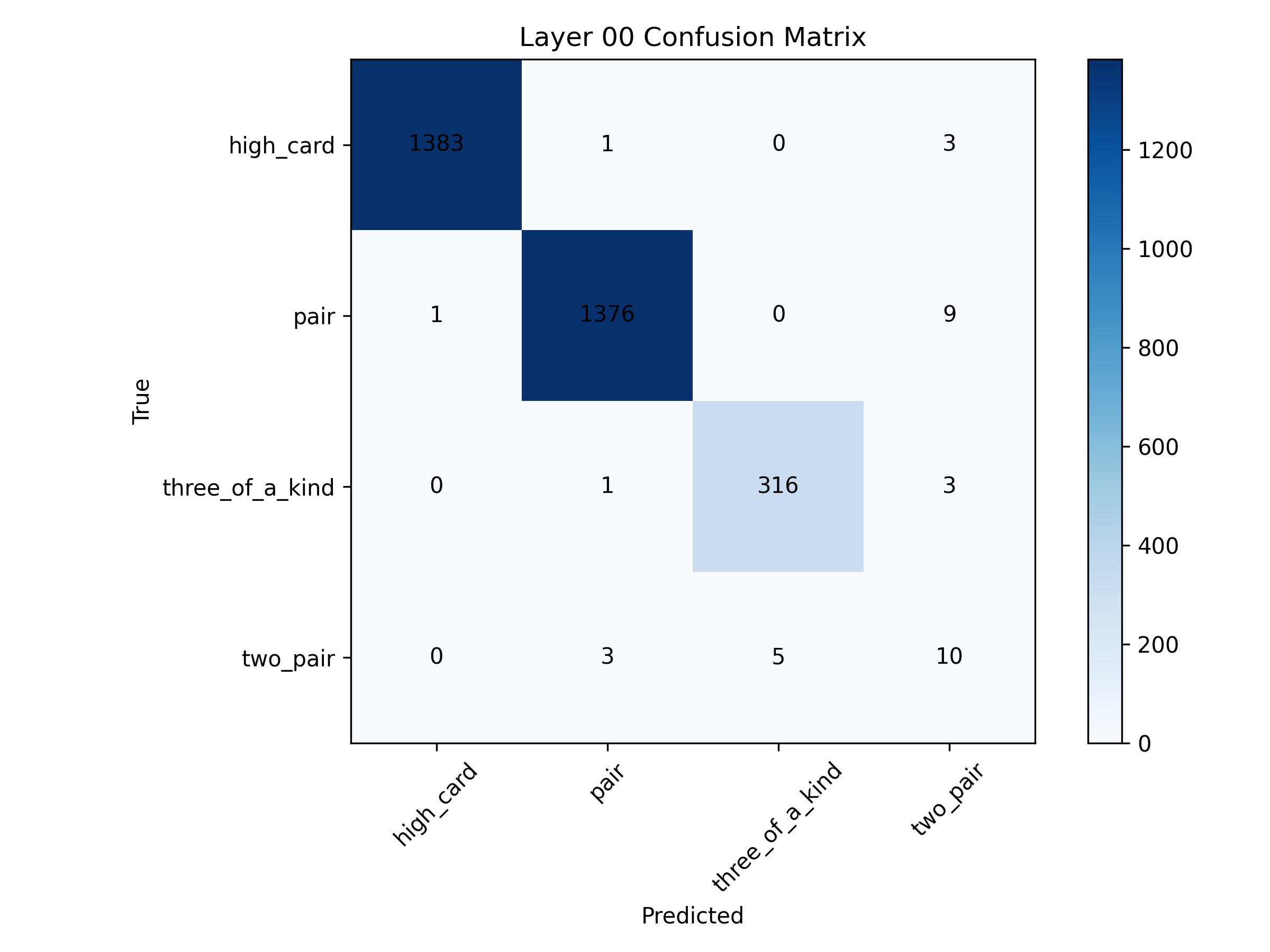}
        \caption{Layer 0 – 35th percentile equation.}
        \label{fig:mlp_35th}
    \end{subfigure}

    \vspace{0.8em}

    \begin{subfigure}[t]{0.47\textwidth}
        \centering
        \includegraphics[width=\linewidth]{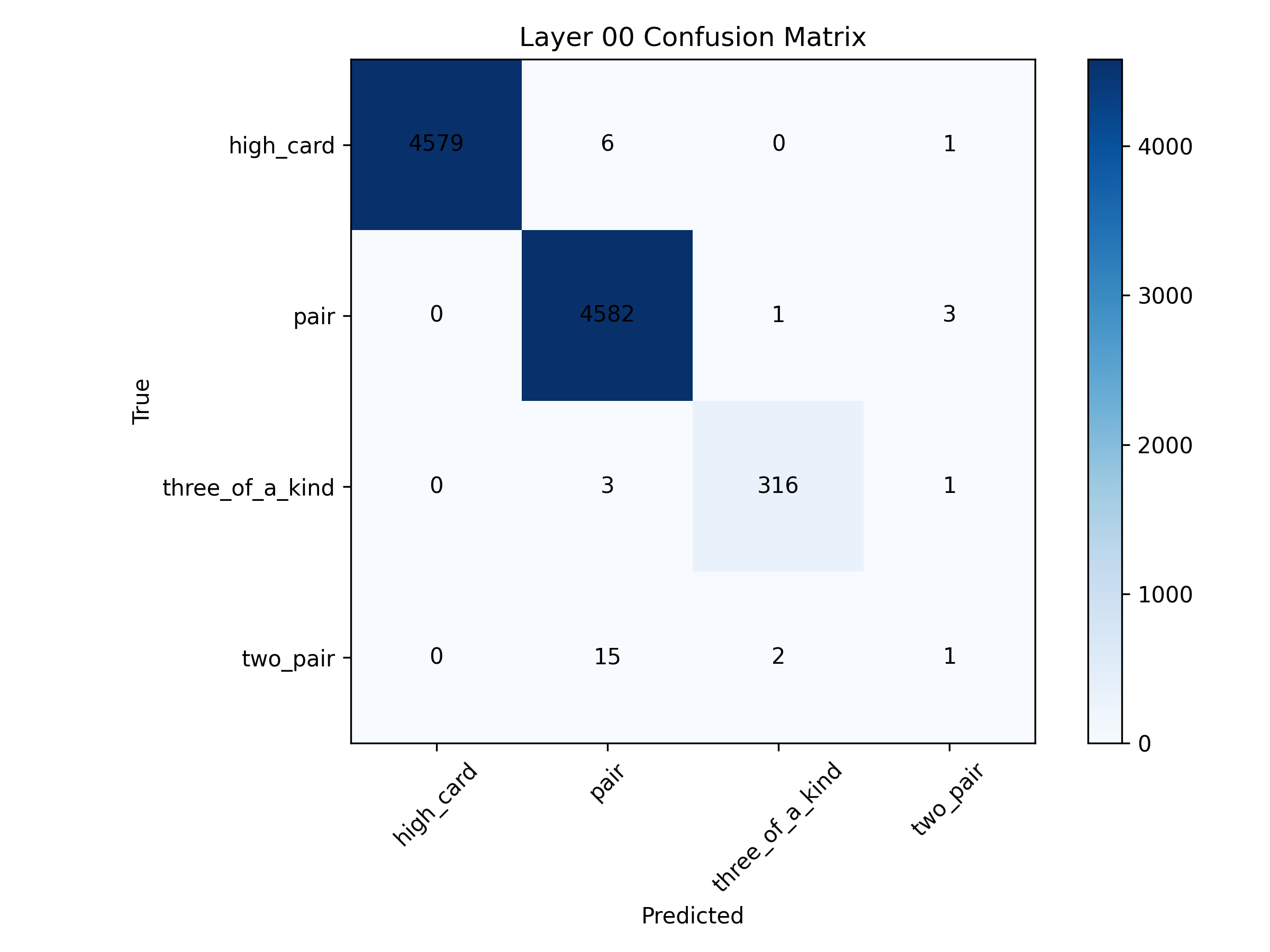}
        \caption{Layer 0 – 40th percentile equation.}
        \label{fig:mlp_40th}
    \end{subfigure}
    \hfill
    \begin{subfigure}[t]{0.47\textwidth}
        \centering
        \includegraphics[width=\linewidth]{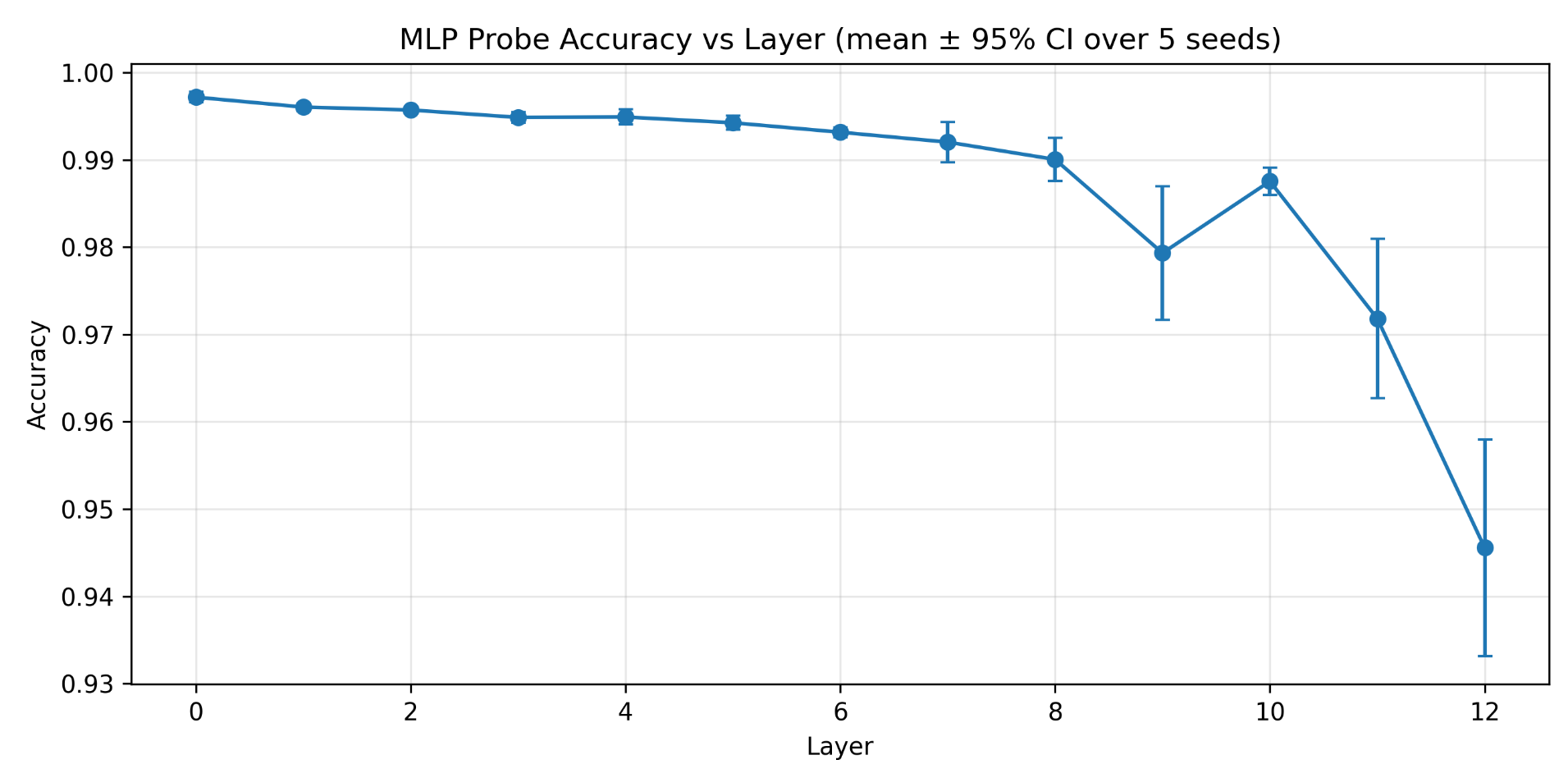}
        \caption{MLP probe accuracy across model layers (40th percentile).}
        \label{fig:Error_bars}
    \end{subfigure}

    \caption{
        MLP probe performance for hand-rank identification. Panels (a-c) show confusion matricies for Layer 0 using datasets balanced by limiting each hand-rank class to the 30th, 35th, and 40th percentile of its unique sample count. Darker diagonal cells indicate more accurate predictions. Representation of rarer hand ranks, such as two pairs, is improved with lower percentiles. Panel (d) shows probe accuracy across all layers using the 40th percentile dataset, with 95\% confidence intervals across five random seeds. Together, results indicate that hand-rank information is encoded strongly and consistently. See Appendix \ref{actionIdentification} for additional deterministic experimental results.
    }
\end{figure}
Expanding on \citet{OthelloAnalysis}, we extract basic deterministic game features such as hand-rank through activation probing. Hand-rank represents the categorical value of a player's hole cards in the context of board cards and is strictly deterministic in our setting. 
We train a separate probe on each layer of the model to note any possible variations between layers that may signify that layer's responsibility in output generation. To prevent over-representation of frequent hand-ranks (e.g. high\_card and pair) and misidentification of internal representational states, we balanced the dataset by capping each class at the 40th percentile of unique hand-rank counts (Appendix \ref{percentileCompute} for more details on this). The linear probe achieves \textasciitilde{}80\% accuracy for identifying hand-rank, while the MLP reaches \textasciitilde{}98\% accuracy (\ref{fig:mlp_40th}, similar to results shown in \citet{OthelloAnalysis}. These results are measured on a dataset excluded from training of the GPT-based model as well as separate from probe training (extended results in Appendix \ref{probes}). As shown in Figure \ref{fig:mlp_40th}, the prominent diagonal in the confusion matrix indicates high class-wise accuracy, demonstrating that the internal activations of the model reliably encode the rank of the hand, which implies that the model is developing an internal representation of poker hand states, rather than just memorizing statistics. Furthermore, small error bars reflect low variance across various seeds in Figure \ref{fig:Error_bars} in the first several layers, demonstrating that a strong consistency of the model's internal representations that reflects the \textasciitilde{98}\% accuracy is present.

\subsection{Stochastic World Model}
\begin{figure}[H]
  \centering
  \begin{subfigure}{0.45\linewidth}
    \includegraphics[width=\linewidth]{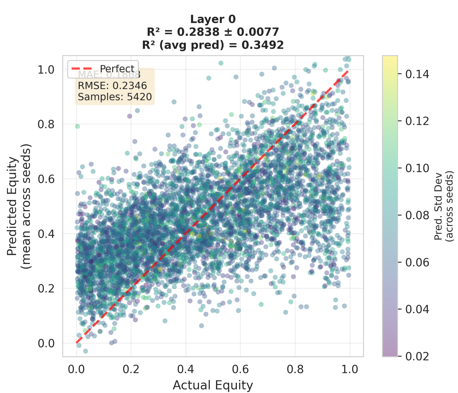}
    \caption{Predicted vs. true equity (layer 0).}
    \label{fig:mlp_equity}
  \end{subfigure}
  \hfill
  \begin{subfigure}{0.45\linewidth}
    \includegraphics[width=\linewidth]{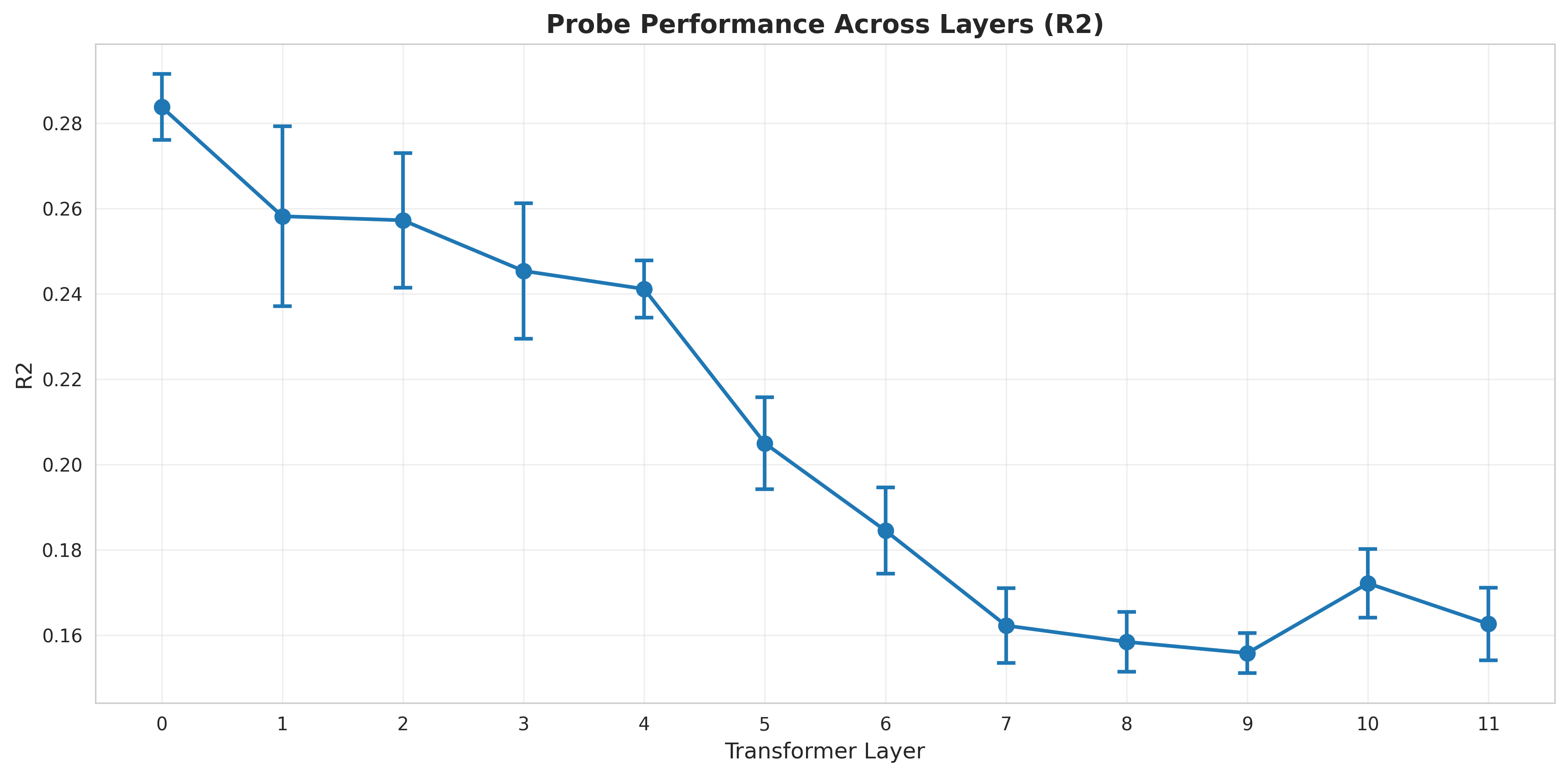}
    \caption{Comparison of $R^2$ across layers.}
    \label{fig:mlp_r2}
  \end{subfigure}
  \caption{Probe performance on stochastic representations. Panel (a) shows predicted versus true hand equity for Layer 0, demonstrating that model activations contain information about winning probability despite incomplete information. Panel (b) shows the $R^2$ value of equity prediction across layers, showing that equity information is strongest in earlier layers of (0-5) and progressively diminishes deeper in the network, consistent with information-bottleneck–style compression.}
  \label{fig:mlp_combined}
\end{figure}

To demonstrate our hypothesis of the language model having internal representations corresponding to the internal representation of the belief state over the POMDP, we trained a two-layer MLP using simulation based equity estimations \citep{Billings1999SelectiveSampling} as our label. For this dataset, we intentionally masked out all hole cards except for those belonging to player one. From our trained probe, we were able to achieve a correlation coefficient of 0.59 on our test dataset predictions, averaged across seeds (Figure \ref{fig:mlp_equity}). This correlation between model activations and the predicted winning potential of a given hand demonstrates that our GPT model has spontaneously developed some internal representation of the hand strength. Observing the $R^2$ across layers, the understanding of equity is most strongly encoded in the early layers, becoming diluted after layers 0-4 (Figure \ref{fig:mlp_r2}). This decrease in $R^2$ across layers is consistent with information bottleneck style compression \citep{Tishby2015DeepLA}, with deeper layers retaining information that is more relevant for token prediction, leading to weaker representations of input variables such as hand equity as the representation becomes more focused on the prediction task. This result is similar to what is observed in our deterministic world model evaluations (Figure \ref{fig:Error_bars}), with hand recognition experiencing the same trend. 
\subsection{Activation Maps}
As a validation of the LLM’s world representation, we observe its ability to discern hands and patterns of different strategic value in poker. We visualized activations using PCA, t-SNE, and UMAP (see Appendix \ref{plots} for extended results). Figure \ref{fig:actplot} reveals distinct clusters, indicating that the model organizes its representations in accordance with hand rank, pairs, and three-of-a-kind clusters closely, thus demonstrating its ability to learn game-level concepts from unsupervised data. Note that the presence of multiple clusters for hand-ranks such as pair indicates that the model is learning a broader representation of pair, where each cluster likely refers to a subset of pairs (the types of cards encompassed in the pair).

\begin{figure}[H]
\centering
    \begin{subfigure}[t]{0.4\textwidth}
        \centering
        \includegraphics[width=\textwidth]{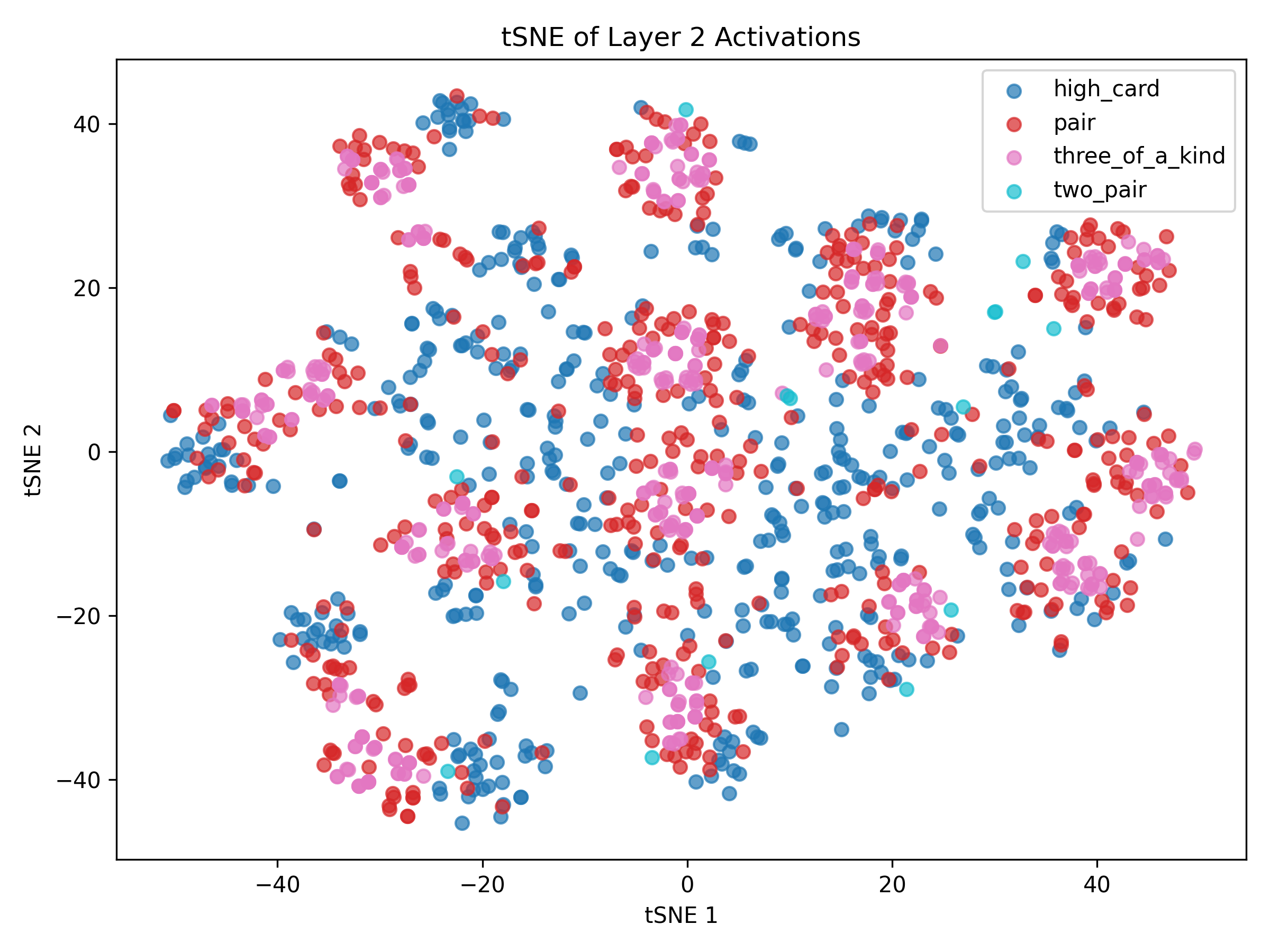}
        \caption{
        t-SNE visualization of Layer 2 activations. Points represent a single hand colored by rank. }
        \label{fig:actplot}
    \end{subfigure}
    \hfill
    \begin{subfigure}[t]{0.4\textwidth}
        \centering
        \includegraphics[width=\textwidth]{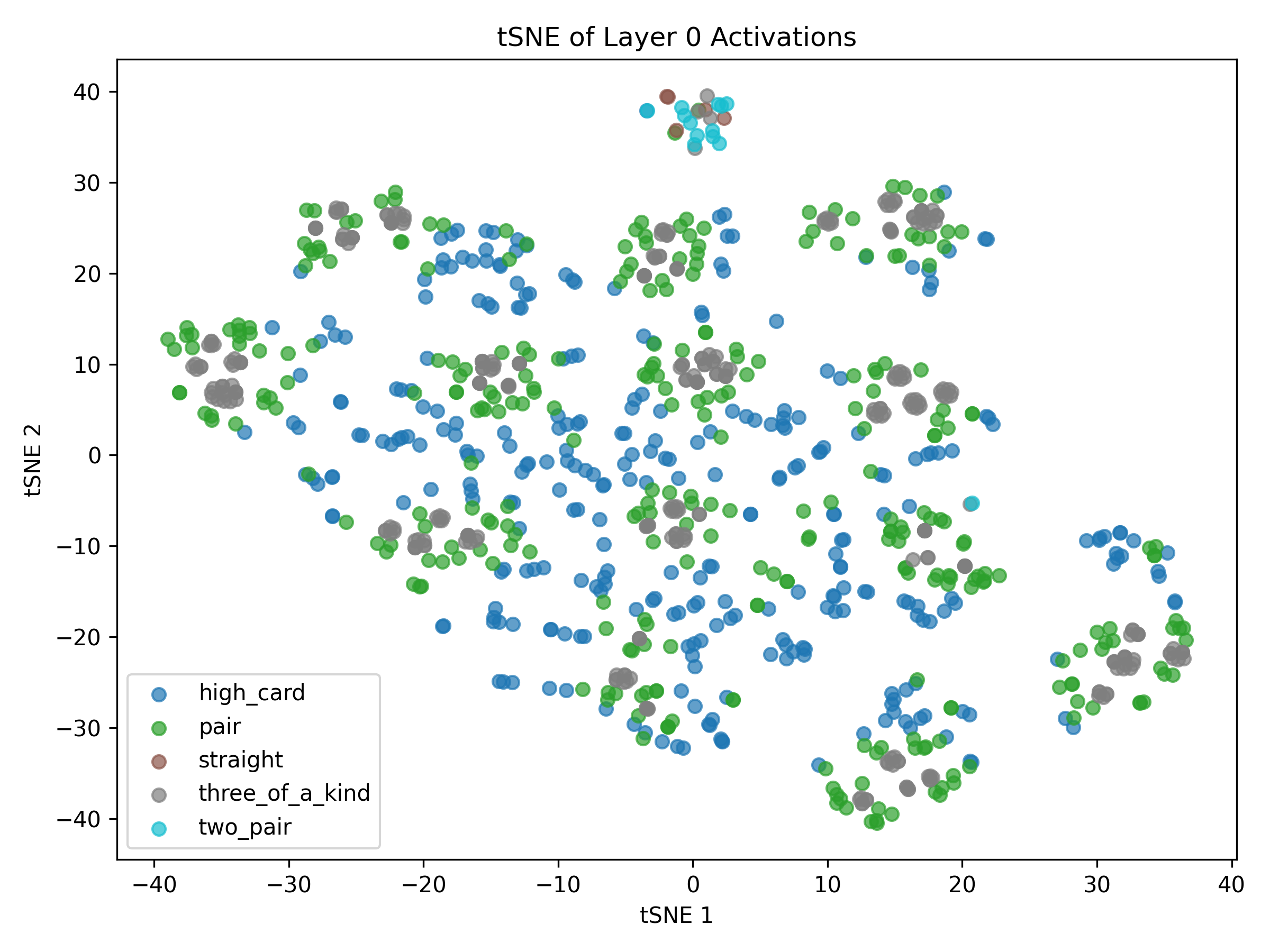}
        \caption{
        t-SNE visualization of Layer 0 activations. Points represent a single hand colored by rank. Note the two\_pair cluster near the top.}
        \label{fig:actplot}
    \end{subfigure}
    \caption{t-SNE visualized activation plots. Activations are clustered by hand-rank and conceptual similarity. Distinct clusters indicate that the model internally organizes hands according to rank or equity strength that follows. Multiple clusters for ranks such as “pair” suggest that the model learns more detailed sub-categories (e.g., types of pairs such as J and K) rather than treating each rank as a single class.}
\end{figure}

\section{Limitations}
Dataset size remains a core limitation of our experiments, as with our generation method it becomes computationally expensive to generate extremely large amounts of data. The size of our dataset impacts the model accuracy on its task as well as results obtained from probes \citep{ChessFine-Tuned}. Our deterministic probing analysis is also limited by this factor, as there are insufficient examples of rarer hands such as straights or flushes in our generated datasets for us to accurately probe for them. Additionally, the process for generating data may be overly simplified in relation to the complexity of poker interactions, potentially impacting how the model understands the game and it's simple and complex variables. This is an unfortunate consequence of being forced to use synthetic data, due to a lack of fully documented poker hand datasets. Due to this, we are unable to validate our results with additional datasets. Finally, there are no guarantees that current results regarding LLM beliefs of the Poker POMDP are able to be extended to analysis of new stochastic poker variables, or to novel domains.

\section{Conclusion and Future Works}
We demonstrated that a GPT-2-based model trained on PHH-style data can develop a deterministic understanding of the game state as well as an understanding of stochastic game elements. This brings us closer to extending the emergent world model hypothesis to games characterized by incomplete information. To extend our work, we hope to further scale our base LM and formalize our intuitions of LLM Bayesian behavior---in particular, extracting LLM beliefs of the Poker POMDP (see Appendix \ref{subsec:theory} for theory)---and better understand the structure of LLM predictive world representations through SAEs and further probing experiments. 

\section{Acknowledgments}
We are extremely grateful to the Algoverse research program for computational resources and extensive mentorship. We also thank the anonymous reviewers of our paper for their helpful feedback.

\bibliography{custom}

\appendix

\section{Appendix}

    Our code can be found at:
    \begin{itemize}
        \item \href{https://anonymous.4open.science/r/poker-interp-4653/}{https://anonymous.4open.science/r/poker-interp-4653/}
    \end{itemize}
\section{Theoretical Justification of Bayesian World Models}
\label{subsec:theory}
In this section, we motivate LLMs as MLE learners in a POMDP setting, and formalize the connection between these LLM probabilistic "belief states" and concrete residual stream activations that are separated through linear probes. 
Past work such as \cite{shai2024beliefgeometry} demonstrates that geometry of a Hidden Markov Model's belief states can be recovered in the residual stream of a transformer. In this work, we formalize the meaning of an LLM "belief state" for next-token prediction that represents the trajectory of partially observable Markov processes. 

\subsection{Belief State and Poker MDP Definitions}
Consider $Z$ as some unobserved latent world state in the LLM (i.e. the board state in Othello, hidden cards in Poker, semantic topics of conversation in NLP) along with a history of tokens up to time t as $h = (x_1, \cdots, x_t)$. The true next token distribution is given as 
\begin{align*}
    p(x_{t+1}| h_t) = \sum_z p(x_{t+1} | z, h_t)p(z|h_t)
\end{align*}
So the next token depends on our distribution over latent states $p(z|h_t)$, which we encode as our \textbf{LLM belief state}. For an ideally trained LLM in deterministic games of chess/Othello, this distribution is just an indicator of the deterministic board state given a sequence of moves, but for games such as poker we uncover a nontrivial distribution over latent space. 

In short, \textit{for next-token prediction in a POMDP setting, the LLM must carry information at least as strong as $p(z|h_t)$}. We view Poker as such a partially observable Markov Decision Process (POMDP), defining states $S$ as the full game specification with a partially observable subset $O$ representing outside cards and behaviors, actions $\mathcal{A}$ as a single player's options (raising, folding, calling, etc), and the transition dynamics $\mathcal{T}$ representing stochasticity over dealt cards as well as other player's actions. 

From standard POMDP analysis, we recall that the belief state is a sufficient statistic for decision-making (in our setting, next-token prediction) \cite{Kaelbling1998POMDP}.

\subsection{Linearity of Predictions in the Belief State}
For POMDPs over deterministic games as explored in \cite{OthelloAnalysis}, we can justify the use of linear probing through a theoretical analysis as predictions of the future as linear functions on the LLM's belief state.

\textbf{Linearity Lemma:} Given our Poker POMDP $\mathcal M=(\mathcal S,\mathcal A,\mathcal O,T,\Omega)$ and a policy $\pi$ (our sequence model trained on poker games), let $H_t$ be the history up to $t$ and $b_t \in \Delta(\mathcal S)$ the belief $b_t(s)=\Pr(S_t=s\mid H_t)$.
For any finite-horizon future event $F$ measurable w.r.t.\ $(S_{t+1:T},O_{t+1:T},A_{t:T-1})$ under $\pi$,
there exists a vector $v_F \in \mathbb R^{|\mathcal S|}$ such that
\[
\Pr_\pi(F \mid H_t) \;=\; \sum_{s\in\mathcal S} b_t(s)\, v_F(s) \;=\; \langle b_t, v_F \rangle .
\]
In other words, our prediction of future events given our current history under the policy is \textit{linear} in our defined belief state. In particular, then for each observation $o\in\mathcal O$,
\[
\Pr_\pi(O_{t+1}=o \mid H_t) \;=\; \langle b_t, v_o\rangle ,
\]
where $v_o$ is a per-observation vector of weights (in practice, learned by a linear probe) so the entire next-observation distribution is an affine linear map of $b_t$.

\textit{Proof:}

Let $f=\mathbf 1_F$ be the indicator of $F$. By the tower rule,
\[
\mathbb E_\pi[f\mid H_t]
= \sum_{s\in\mathcal S} \Pr_\pi(S_t=s\mid H_t)\, \mathbb E_\pi[f\mid S_t=s, H_t].
\]
In a POMDP, the controlled Markov property implies that, given $S_t=s$ (and with policy $\pi$ fixed), the distribution of
$(S_{t+1:T},O_{t+1:T},A_{t:T-1})$ does not depend on the specific past $H_t$ (i.e. our token histories); hence
$\mathbb E_\pi[f\mid S_t=s,H_t]=\mathbb E_\pi[f\mid S_t=s] \eqqcolon v_F(s)$ by conditional independence.

Then, this directly gives us the probabilities as 
$\Pr_\pi(F\mid H_t)=\sum_s b_t(s)\, v_F(s)=\langle b_t, v_F\rangle$.
Taking $F=\{O_{t+1}=o\}$ yields the desired linearity of observation result.

So the PODMP belief state "automatically" gives us linearity! Intuitively, if the transformer trained on next-token prediction does in fact hold its belief state in the residual stream, then any predictive probe should be \textit{linear} in these activations. Consider the following toy example: 

Suppose we have a simple binary hypothesis testing $Z = \{0, 1\}$ to denote which coin is in use among two coins with probabilities $p_1, p_2$. Conditional on $Z$, our observations $x_1, \cdots, x_t \in \{H, T\}$ are id with $P(x_i = H | Z = z) = p_z, P(x_i = T | Z = z) = 1 - p_z, z \in \{0, 1\}$. Our log likelihood ratio (LLR) of one hypothesis over the other evolves with ratios $\log(\frac{\theta_1}{\theta_0})$ and $\log(\frac{1 - \theta_1}{1 - \theta_0})$. Our log likelihood ratio $\eta_t$ by assumption exists in the residual stream of our sequence prediction model. 

\paragraph{How does the residual stream "provide" the belief coordinate?} Our connection between linear probes and POMDP belief states lies in the LLM's residual stream, motivated from the theory of transformer circuits \cite{Anthropic2021MathematicalFramework}. 
Let $r_t\in\mathbb R^d$ denote the residual stream at position $t$.
Assume the model stores $\eta_t$ along a direction $v\in\mathbb R^d$:
\begin{equation}
r_t \;\approx\; r_0 \;+\; \eta_t\, v \;+\; \varepsilon_t
\;=\; r_0 \;+\; \Big(\eta_{t-1}+\mathrm{LLR}(x_t)\Big)\,v \;+\; \varepsilon_t,
\end{equation}
so residual \emph{addition} implements evidence accumulation.
With a fixed unembedding $U\in\mathbb R^{d\times |\mathcal V|}$, the logits satisfy
\begin{equation}
\mathrm{logits}(x\mid x_{1:t}) \;=\; U_x^\top r_t + c_x
\;\approx\; (U_x^\top v)\,\eta_t \;+\; \text{const},
\end{equation}
and so we get a linear function of the belief coordinate. Ultimately, we get that a linear probe $w$ can recover $\eta_t$ from $r_t$ via $\hat\eta_t=w^\top r_t$.

Related works explore the theoretical justifications of linear probes in partially observable processes. Two complementary works make a precise claim in this direction: \emph{predictions are linear functionals of state}.

\textbf{POMDPs.}
Under a fixed policy, the \emph{belief} $b_t$ (posterior over latent state) is a sufficient statistic for prediction/control; for any finite-horizon event $F$,
$\Pr(F\mid H_t)=\langle b_t, v_F\rangle$ for some vector $v_F$ determined by the dynamics/observations \citep{Kaelbling1998POMDP,SmallwoodSondik1973}. Thus the entire next-observation distribution is an affine-linear map of $b_t$.

\textbf{PSRs.}
If the Hankel matrix of future-test probabilities has finite rank $k$, there exists a $k$-dimensional \emph{predictive state} $p(h)$ (probabilities of \emph{core tests}) such that the probability of \emph{any} test $\tau$ is linear: $\Pr(\tau\mid h)=c_\tau^\top p(h)$ \citep{Littman2001PRS,Singh2004PSRTheory}. Hence, \emph{if} a transformer stores an affine transform of $b_t$ or $p(h)$ in its residual stream, there \emph{exists} a linear probe (and the model's own unembedding) that recovers the relevant prediction

\subsection{Relation to poker (this paper)}
Poker is a canonical partially observable domain where the minimal predictive state is a \emph{belief over hidden hands} (often summarized as a range). Training a next-token model on poker strings (actions and reveals) creates direct pressure to maintain this belief internally, because the Bayes-optimal next-token distribution is the mixture over hypotheses weighted by the current belief. Our empirical program---linear probes for range log-odds, layerwise tuned-lens trajectories, and causal edits along probe directions---follows the Othello/chess playbook while grounding interpretation in POMDP/PSR sufficiency. This explains why (i) range features should be linearly decodable, (ii) updates should approximate additive log-likelihood ratios upon new evidence, and (iii) editing decoded belief directions should steer action logits in predictable ways.

\section{LLM World Models}
\label{subsec:worldmodels}

In this section we further explore the literature of LLM world models and discuss our contributions in the context of prior work. 

In the previous section, we formalize our definition of world models/belief state for POMDPs. In the case of OthelloGPT, this world model takes the form of a representation of the board state/dynamics in the residual stream, but in our Poker case, the relevant latent is instead a belief (range) over hidden information, such as player hands and strategies/deck cards. 
\subsection{What counts as a world model?}
Broader than POMDPs and games, a world model functions as an internal state whose evolution under the model approximates the latent state/dynamics of the data-generating process, such that predictions are a (typically affine-linear) functional of that state.
Early neural control work formalized this idea broadly as ``world models'' \citep{HaSchmidhuber2018WorldModels}. In LLMs, the clearest evidence comes from synthetic or structured domains where latent state is objectively defined and recoverable from strings. The primarily goal of LLM world model research in toy settings such as Poker and Othello is to find strong signals that LLMs can learn higher-order structure (translated to the language setting, higher-order emotions/rationalities) from sampled sequences of these unobserved transition dynamics. 

\subsection{Empirical evidence in trained sequence models}
\textbf{Board games.}
In \emph{Othello-GPT}, a small transformer trained only to predict legal moves (no board supervision) learns an internal representation of the full board: probes decode square occupancy; causal interventions flip squares and reliably change downstream moves \citep{OthelloAnalysis}. Follow-up work in chess reaches similar conclusions: linear decoders recover piece/square features and editing those features predictably shifts move probabilities, indicating persistent board-state coordinates in the residual stream \citep{Mei2025ChessWorldModel}.

\textbf{Space \& time.}
When trained on ordinary text corpora, LLMs encode geometric and temporal structure that is linearly recoverable: e.g., countries/cities embed into coherent low-dimensional coordinate systems and historical entities align along temporal axes \citep{Gurnee2023SpaceTime}. These are not merely lexical clusters but approximately \emph{metric} maps, suggesting latent factors aligned with world structure.

\subsection{Mechanistic lenses on storage and update}
Two families of tools consistently reveal how predictions depend on internal state.

\textbf{Layerwise decoding:}
The \emph{logit lens} and the calibrated \emph{tuned lens} linearly decode token distributions from intermediate residual streams, showing a monotone refinement of predictions across depth---consistent with iterative inference/update of a persistent state carried forward by residual addition \citep{Nostalgebraist2020LogitLens,Belrose2023TunedLens}. In practice, we observe this through the refinement of LLM confidence in poker games as more cards are dealt. 

\textbf{Feature decomposition:}
Sparse autoencoders (SAEs) and related dictionary-learning methods recover more monosemantic directions in residual space (e.g., individual board squares, entity features), addressing superposition and enabling targeted causal edits \citep{Cunningham2023SAE,regularText,Elhage2022Superposition}. These results support a picture in which a small set of task-relevant \emph{state features} are embedded (often nearly linearly) and read out by the fixed unembedding matrix.

\section{PHH Formatting}
\label{phh-format}
In PHH notation cards abbreviated to a rank followed by a suit (King of Hearts -> Kh). Table 1 shows how actions are represented in PHH notation.
\begin{table}[H]
\centering
\renewcommand{\arraystretch}{1.3} 
\setlength{\tabcolsep}{12pt} 
\begin{tabular}{|l|l|}
\hline
\multicolumn{2}{|c|}{\textbf{Player Actions}} \\
\hline
\textbf{Standard Representation} & \textbf{PHH Representation} \\
\hline
Hole Cards Dealt      & \texttt{d dh pN card(s)==s)} \\
Board Cards Dealt          & \texttt{d db card(s)} \\
Fold             & \texttt{pN f} \\
Check/Call             & \texttt{pN cc} \\
Bet/Raise               & \texttt{pN cbr amount} \\
Showdown          & \texttt{pN sm card(s)} \\

\hline
\end{tabular}

\caption{PHH-style representations of player actions}
\label{tab:phh_representation}

\end{table}

\section{Dataset Generation Details}
Our dataset generation process creates valid six player No Limit Texas Hold'em poker games. The games are generated as a result of six unique and independent game agents playing against each other. Agents use myopic heuristics, driven by simulated win equity estimates. Each agent is randomly initialized for the following impactful values on decision making, using a provided seed for reproducibility.
\begin{itemize}
    \item Propensity to raise
    \item Tightness in adhering to equity
    \item Bluff frequency
    \item Call willingness
    \item Initial bet scale
    \item Raise scale
    \item Bet continuation
\end{itemize}
This combination of heuristics with equity allows vast amounts of game data to be generated relatively quickly with a competent level of agent play. A considerable limitation of this method is that agents do not adapt over time or learn from others playing styles in a way that humans or more complex game playing agents could.

\section{Supplemental Figures and Tables}
Below we give a diagram of our overall training pipeline for our Poker-GPT model, including our data-masking procedure. 
\begin{figure}[H]
  \centering
  \includegraphics[width=\textwidth]{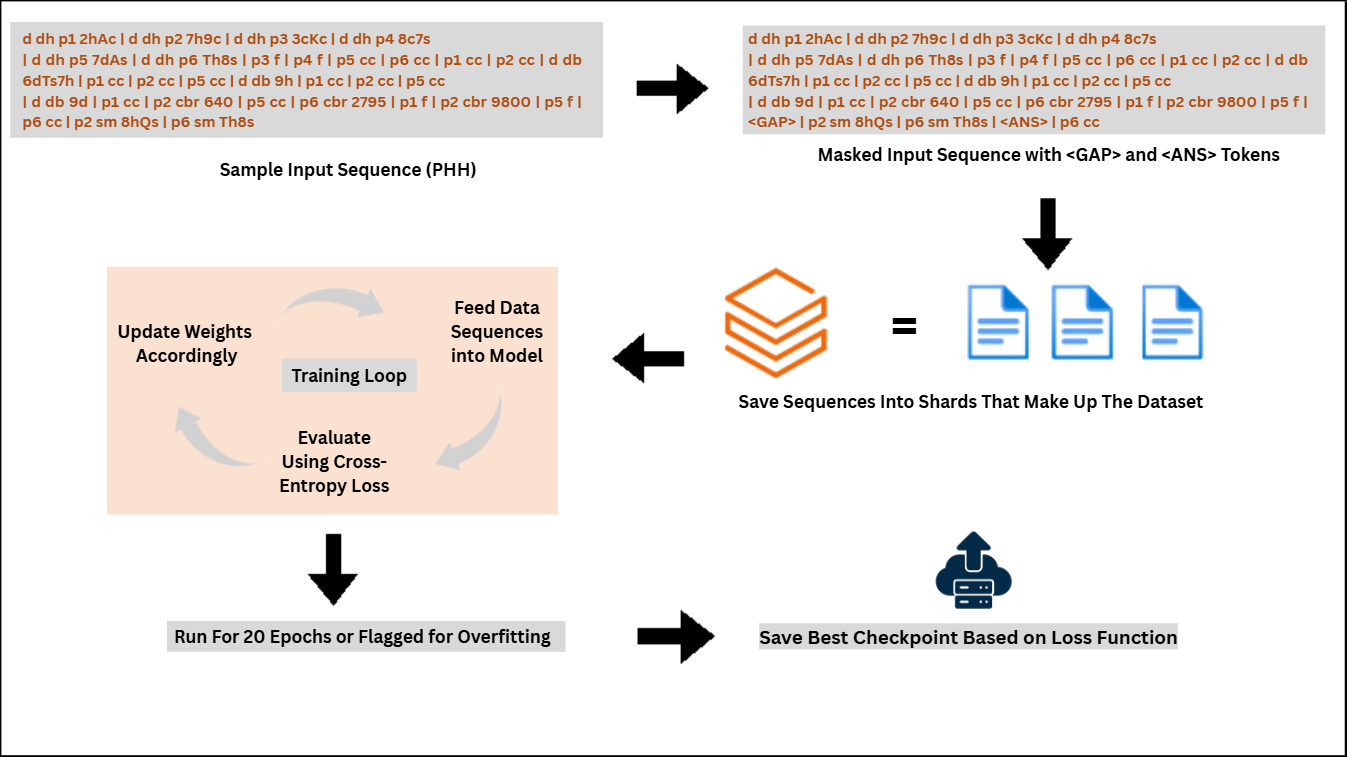}
  \caption{Training Pipeline. We train for up to 20 epochs, with early stopping if validation accuracy declines for three consecutive epochs to prevent overfitting.}
\end{figure}
\section{Compute and Memory Resources}
\label{compute}
We trained our GPT-2-based model, which comprises of 87 million parameters, on an NVIDIA H200 GPU for approximately seven hours. For the training of probes, we leveraged a diverse set of hardware: RTX 5090, NVIDIA H200, A10, and A100 GPUs.

\section{Percentile Computation}
\label{percentileCompute}
\begin{align*}
\text{Let } y &= \text{array of hand rank labels}, \\
\mathcal{L} &= \{ \ell_1, \ell_2, \dots, \ell_k \} = \text{unique labels in } y, \\
c_i &= \sum_{j} \mathbf{1}_{\{y_j = \ell_i\}} \quad \text{for } i = 1,\dots,k \quad \text{(counts per label)}, \\
\text{target\_count} &= \max \Big( \text{percentile}_{40}( \{ c_1, \dots, c_k \} ),\ 10 \Big)
\end{align*}
This balance in count helped us mitigate the impact of the excessive abundance of instances of hand ranks such as high\_card due to their high-frequency nature by chance. This calculation also helps us validate the probe is not just learning to output one hand rank and, instead, is forced to extract intricate information from the activations of the model.

\section{Action Identification}
\label{actionIdentification}
\begin{figure}[ht]
    \centering
    \begin{subfigure}[t]{0.40\textwidth}
        \centering
        \includegraphics[width=\textwidth]{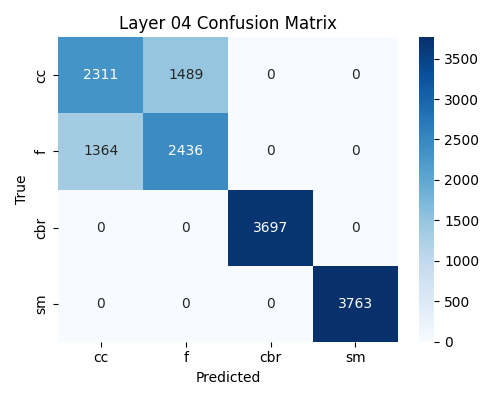}
        \caption{Confusion Matrix for MLP Probe Action Identification on Layer 4.}
        \label{fig:actionlp}
    \end{subfigure}
    \hfill
    \begin{subfigure}[t]{0.40\textwidth}
        \centering
        \includegraphics[width=\textwidth]{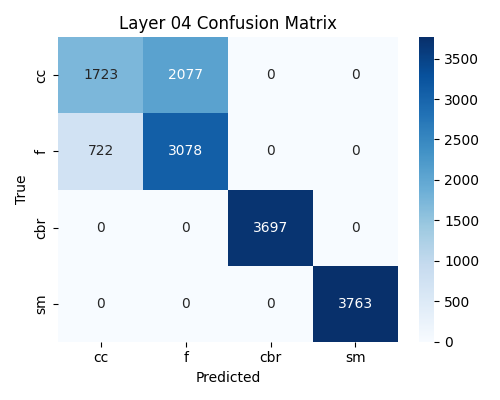}
        \caption{Confusion Matrix for MLP Probe Action Identification on Layer 4.}
        \label{fig:actionmlp}
    \end{subfigure}
    \caption{
        Action identification performance using linear and MLP probes when the action
        token (e.g., \texttt{f}, \texttt{cc}, \texttt{sm}) is masked out during inference, to prevent models from "cheating". Panels (a) and (b) show
    confusion matrices for the linear probe and two-layer MLP probe respectively,
    evaluated on Layer 4 of the transformer. Both probes achieve similar accuracy
    (\(\sim\)80\%), suggesting that the model’s internal activations already encode
    sufficient information about common actions and their situational context, but
    also show confusions between actions with similar local structure (e.g.,
    \texttt{cc} vs.\ \texttt{f}).}
    \label{fig:probe_side_by_side}
\end{figure}

To investigate how the model encodes the game state, we analyzed its ability to predict the action taken when the token corresponding to the action (\texttt{f}, \texttt{cc}, etc.) was masked out. This helped us prevent the model from 'cheating' and seeing the token within the playthrough. With this approach, the model was forced to determine the action taken based on the context of the playthrough. After running both a linear classifier probe and two-layer MLP (only one hidden layer), we noticed that the linear probe (see Figure \ref{fig:actionlp}) and MLP probe (see Figure \ref{fig:actionmlp}) both achieved \textasciitilde{}80\% accuracy for action identification. This implies that the model is learning to associate actions to certain contexts such as card reveals (as in the case of \texttt{sm}) and possibly learning how they fit in these contexts.

\section{Hand-Rank Probe Experiments}
\label{probes}

\begin{figure}[H]
    \centering

    \begin{subfigure}[b]{0.23\textwidth}
        \centering
        \includegraphics[width=\textwidth]{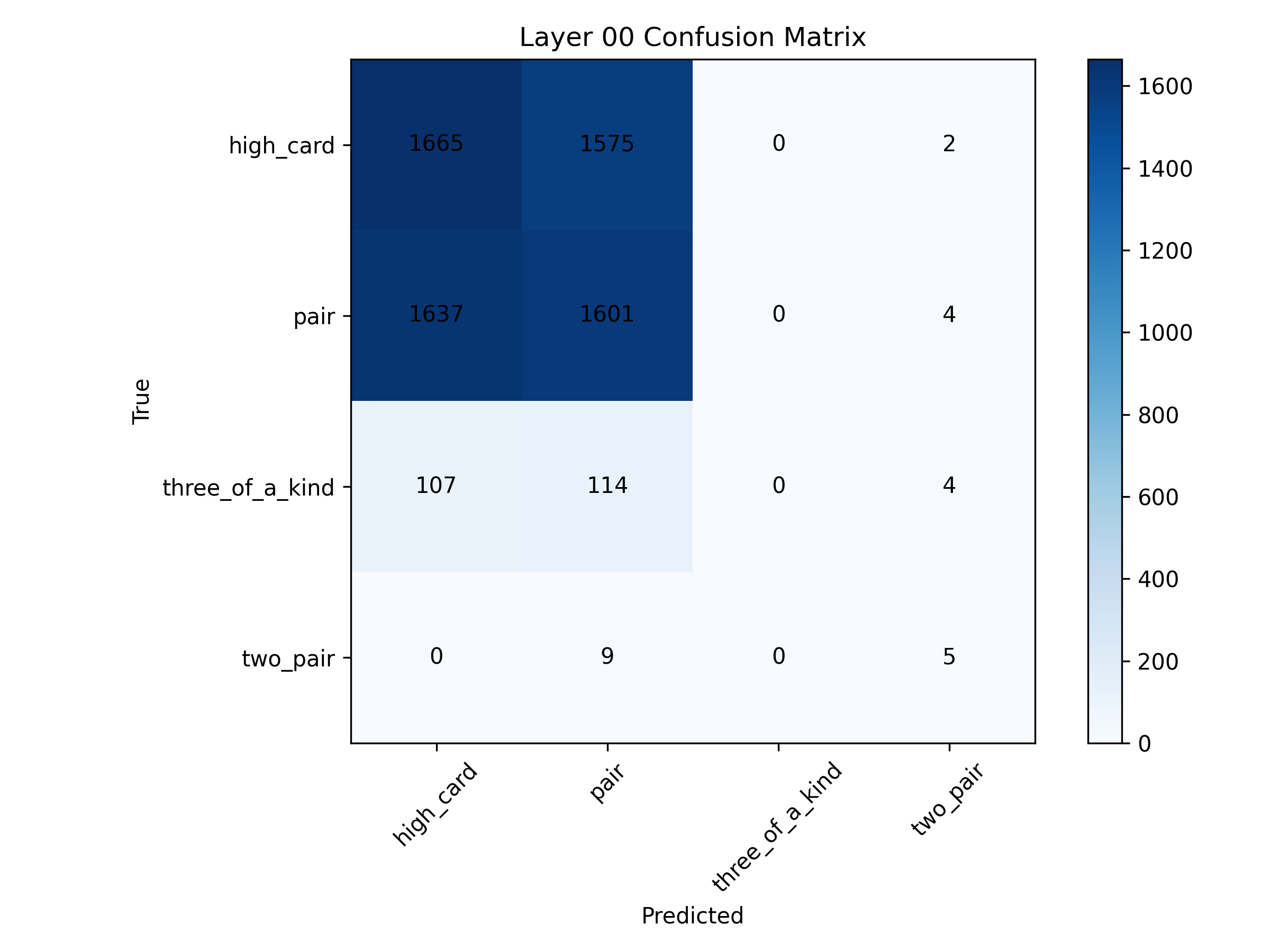}
        \caption{Layer 0}
        \label{fig:layer0_hr}
    \end{subfigure}
    \hfill
    \begin{subfigure}[b]{0.23\textwidth}
        \centering
        \includegraphics[width=\textwidth]{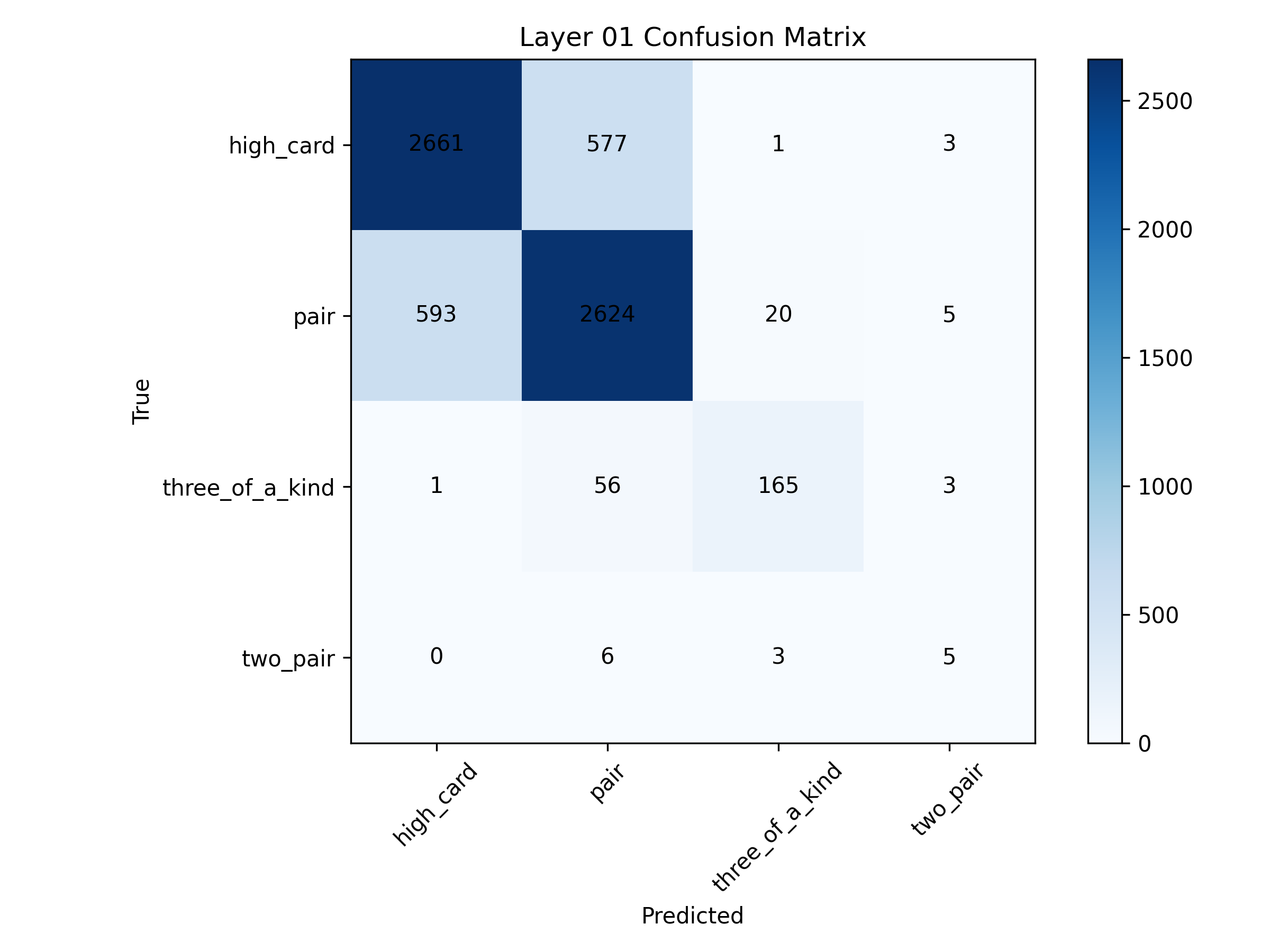}
        \caption{Layer 1}
        \label{fig:layer1_hr}
    \end{subfigure}
    \hfill
    \begin{subfigure}[b]{0.23\textwidth}
        \centering
        \includegraphics[width=\textwidth]{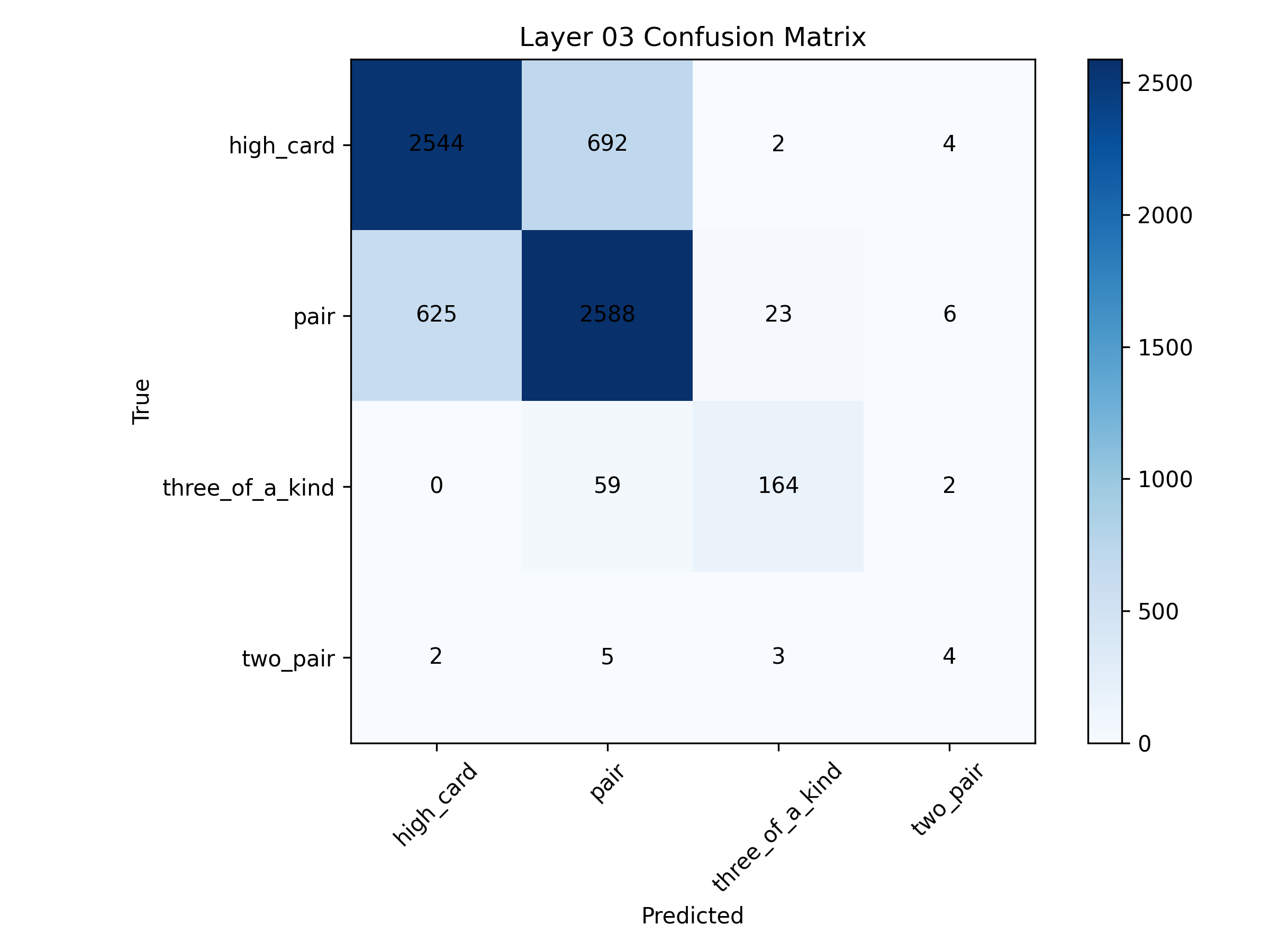}
        \caption{Layer 2}
        \label{fig:layer2_hr}
    \end{subfigure}
    \hfill
    \begin{subfigure}[b]{0.23\textwidth}
        \centering
        \includegraphics[width=\textwidth]{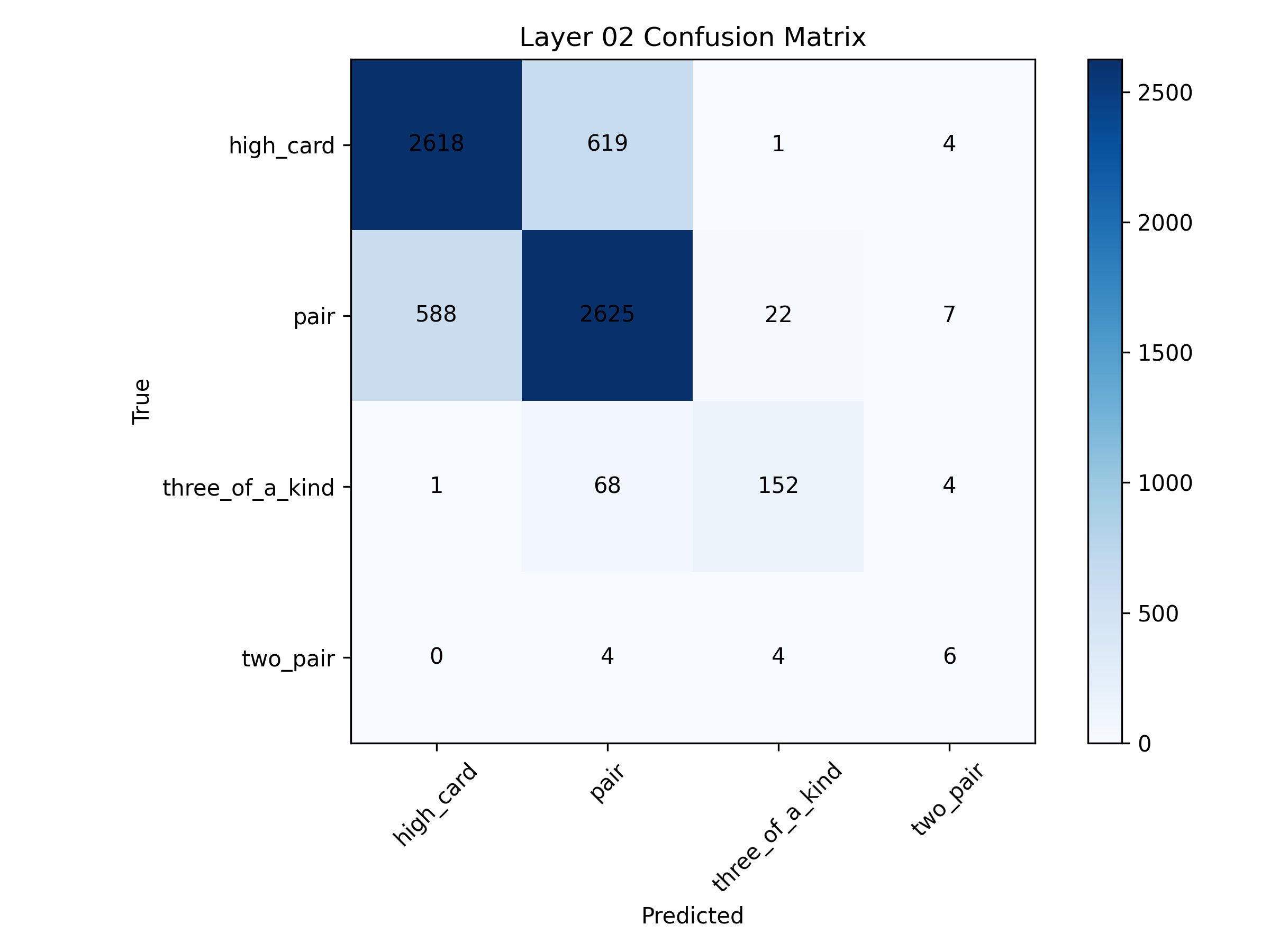}
        \caption{Layer 3}
        \label{fig:layer3_hr}
    \end{subfigure}

    \vspace{0.5em}

    \begin{subfigure}[b]{0.23\textwidth}
        \centering
        \includegraphics[width=\textwidth]{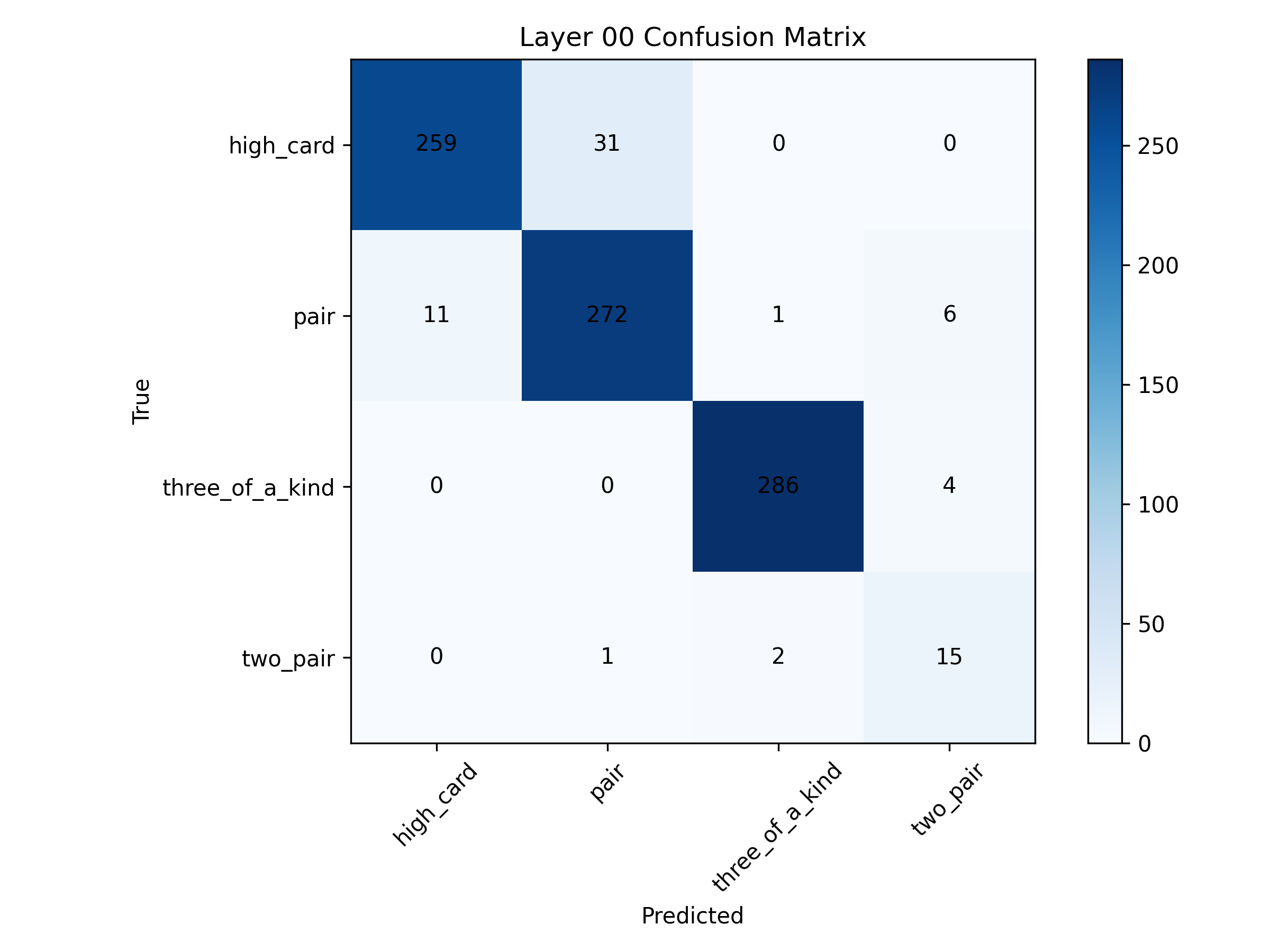}
        \caption{Layer 0}
        \label{fig:layer0_ac}
    \end{subfigure}
    \hfill
    \begin{subfigure}[b]{0.23\textwidth}
        \centering
        \includegraphics[width=\textwidth]{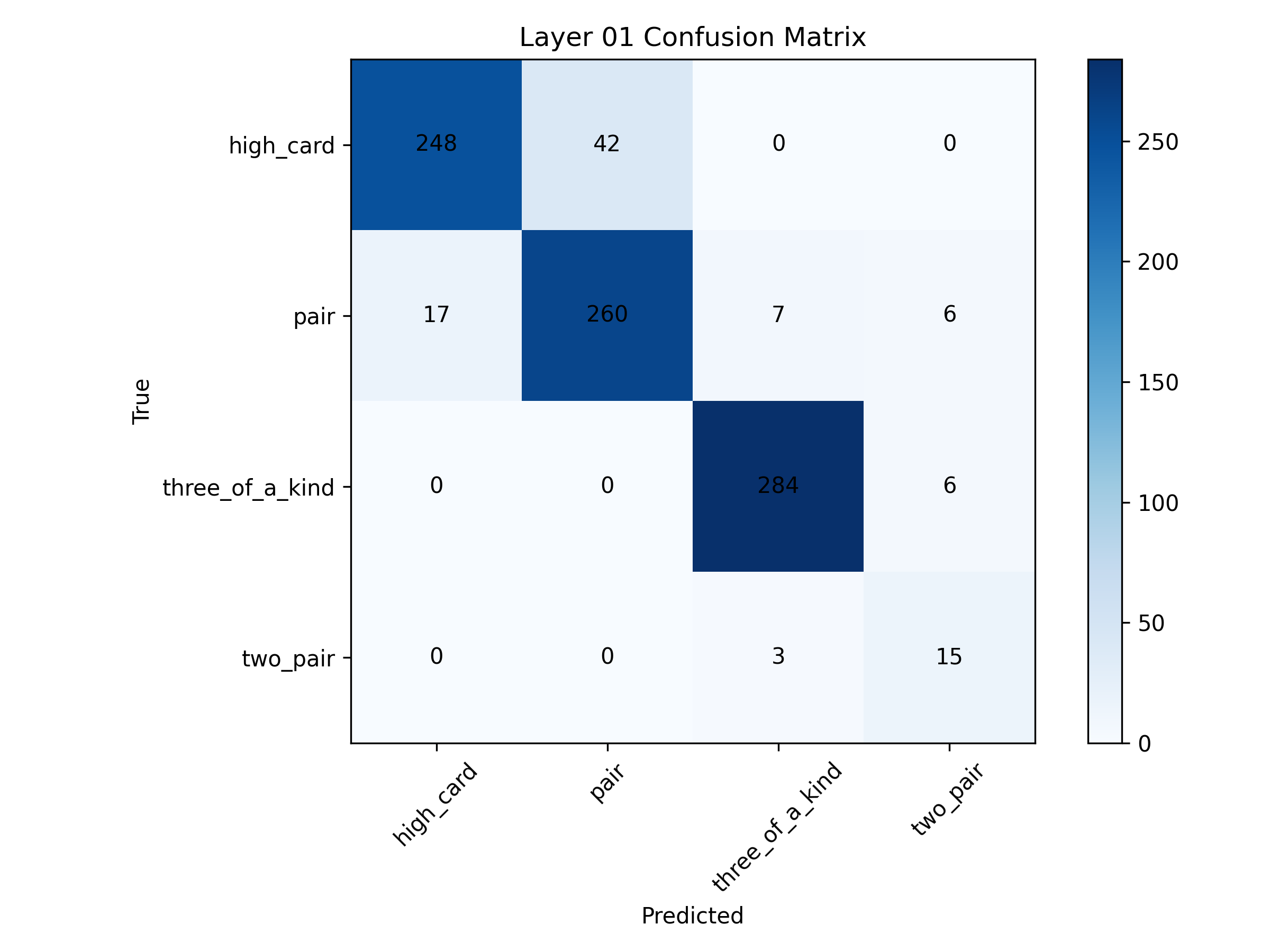}
        \caption{Layer 1}
        \label{fig:layer1_ac}
    \end{subfigure}
    \hfill
    \begin{subfigure}[b]{0.23\textwidth}
        \centering
        \includegraphics[width=\textwidth]{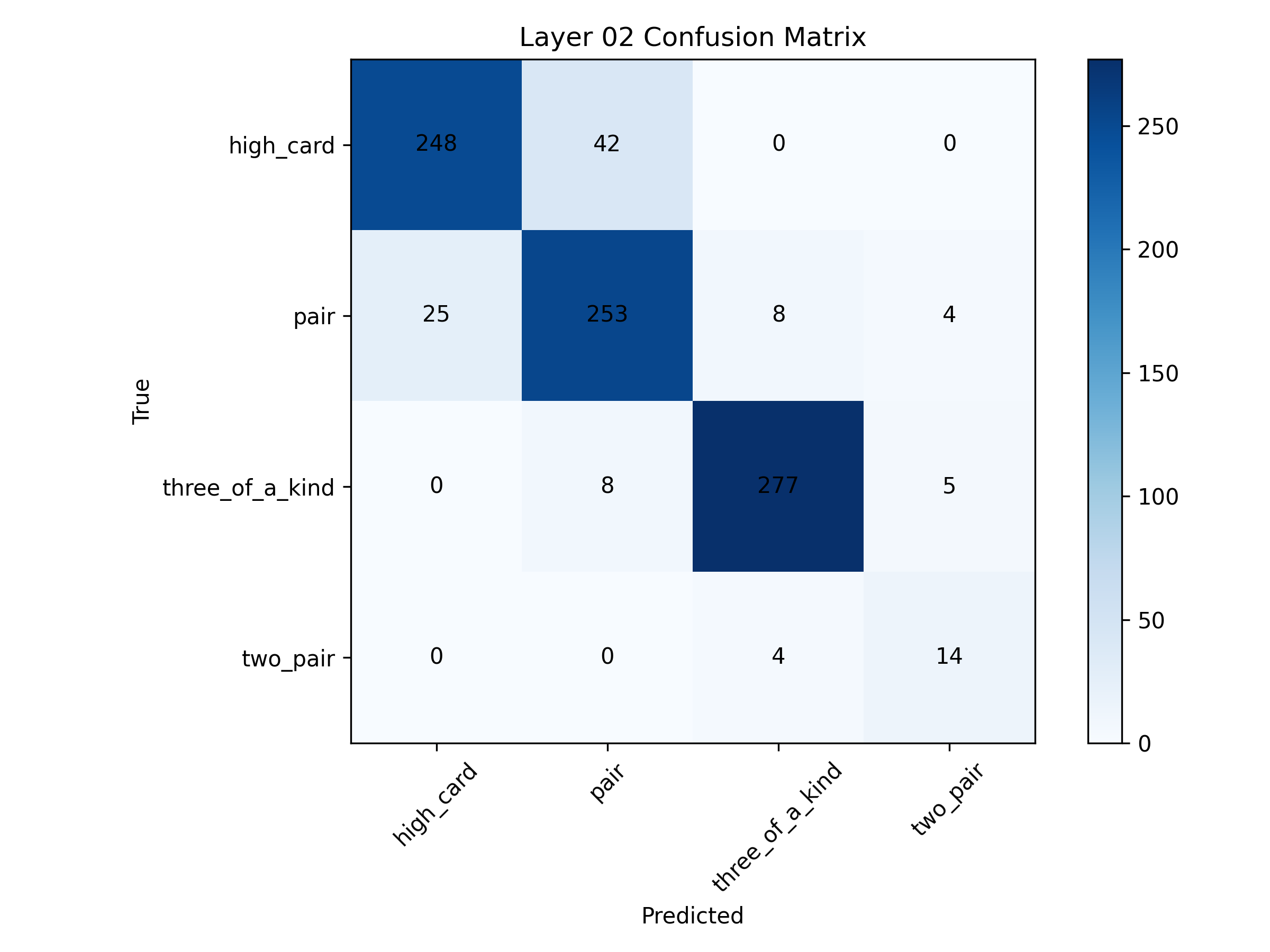}
        \caption{Layer 2}
        \label{fig:layer2_ac}
    \end{subfigure}
    \hfill
    \begin{subfigure}[b]{0.23\textwidth}
        \centering
        \includegraphics[width=\textwidth]{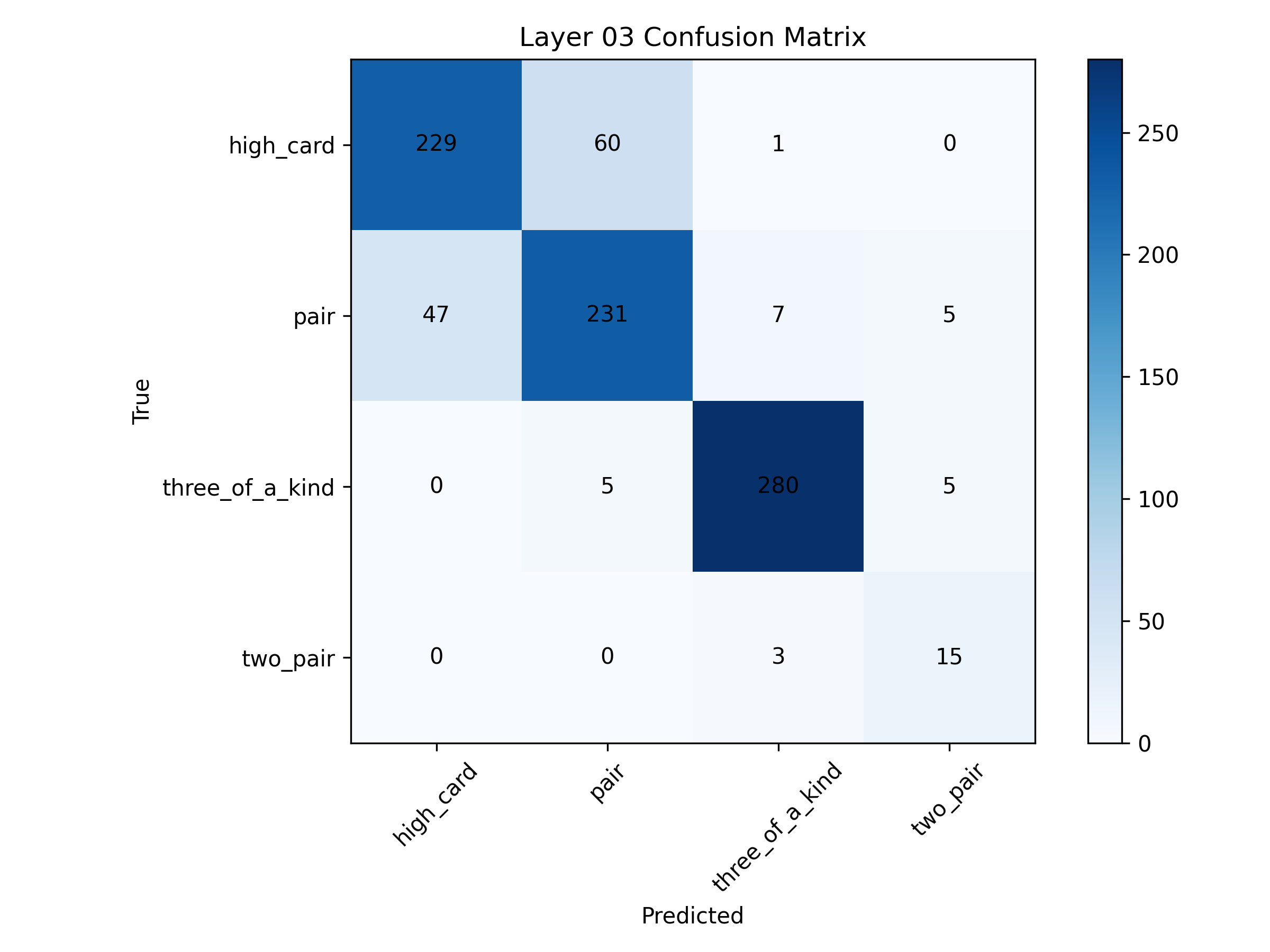}
        \caption{Layer 3}
        \label{fig:layer3_ac}
    \end{subfigure}

    \vspace{0.5em}

    \begin{subfigure}[b]{0.23\textwidth}
        \centering
        \includegraphics[width=\textwidth]{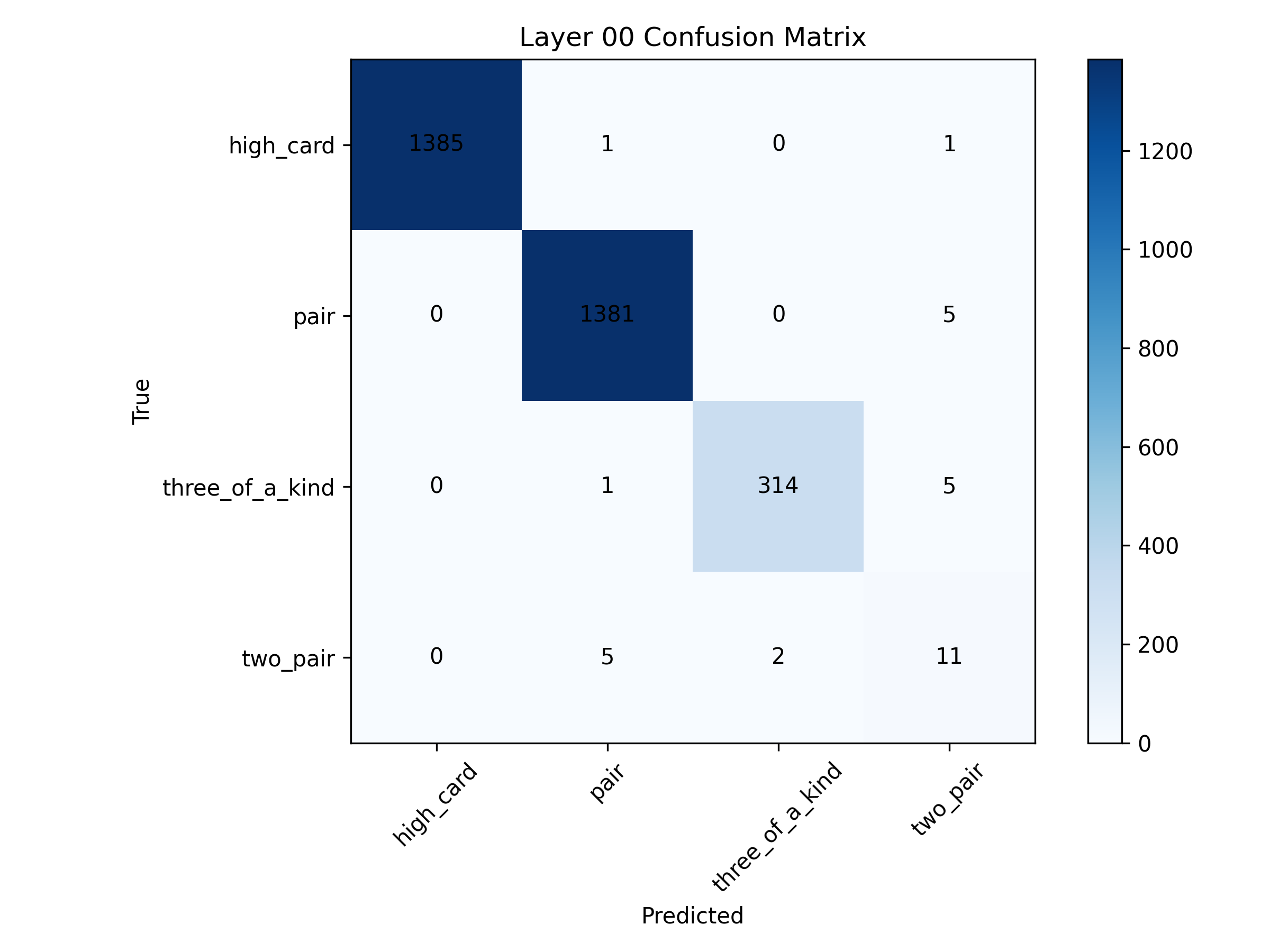}
        \caption{Layer 0}
        \label{fig:layer0_m1}
    \end{subfigure}
    \hfill
    \begin{subfigure}[b]{0.23\textwidth}
        \centering
        \includegraphics[width=\textwidth]{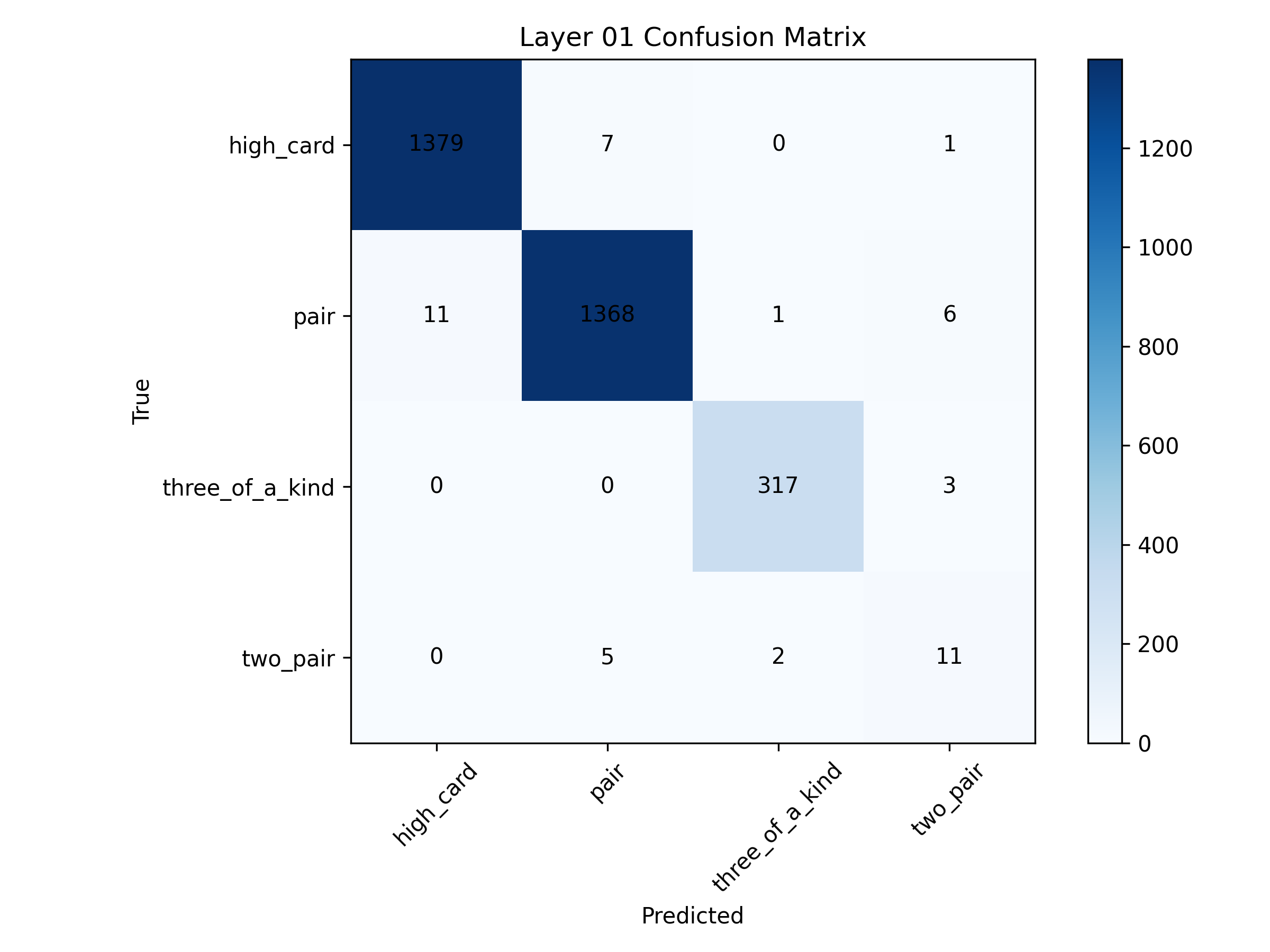}
        \caption{Layer 1}
        \label{fig:layer1_m1}
    \end{subfigure}
    \hfill
    \begin{subfigure}[b]{0.23\textwidth}
        \centering
        \includegraphics[width=\textwidth]{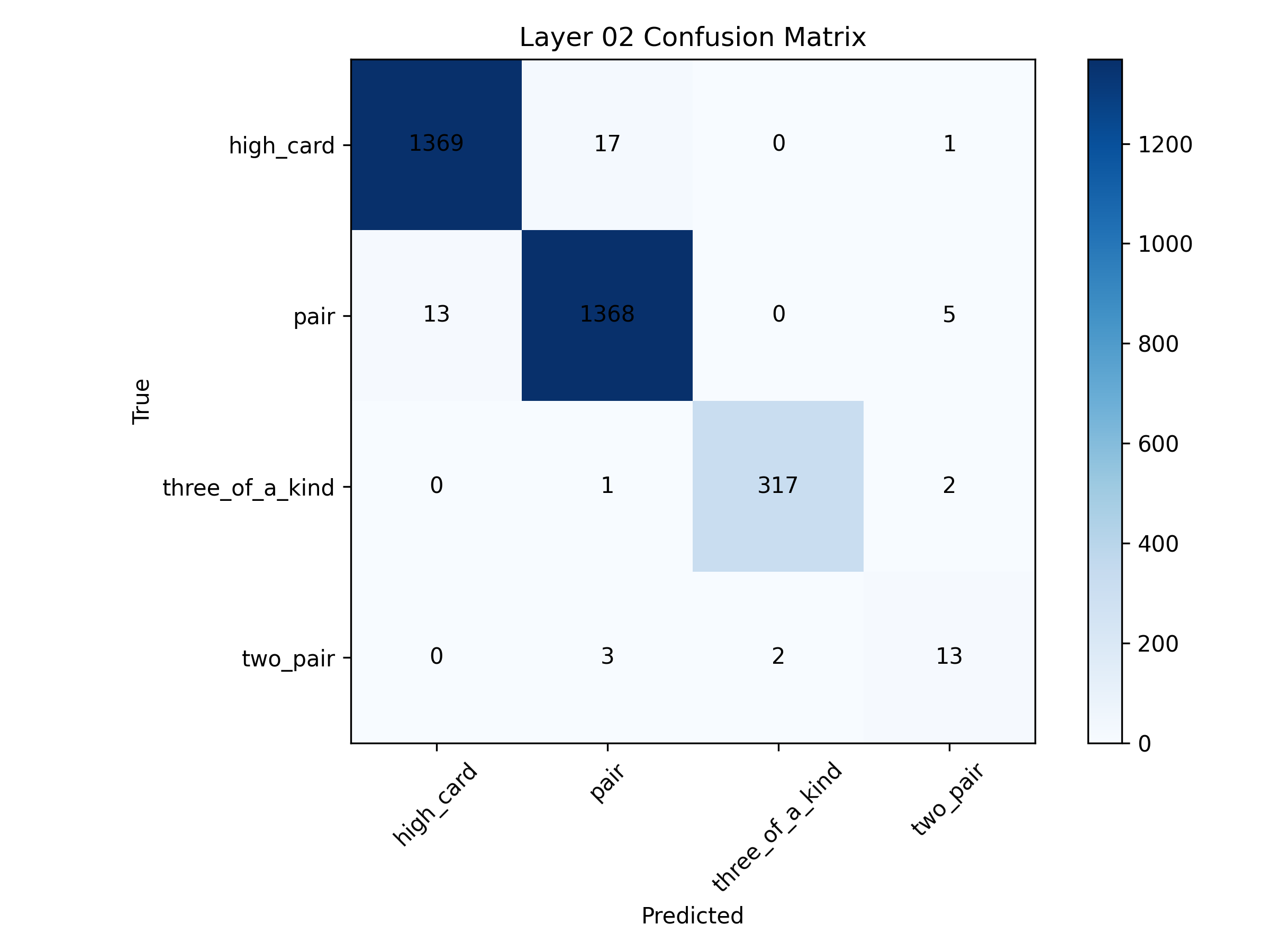}
        \caption{Layer 2}
        \label{fig:layer2_m1}
    \end{subfigure}
    \hfill
    \begin{subfigure}[b]{0.23\textwidth}
        \centering
        \includegraphics[width=\textwidth]{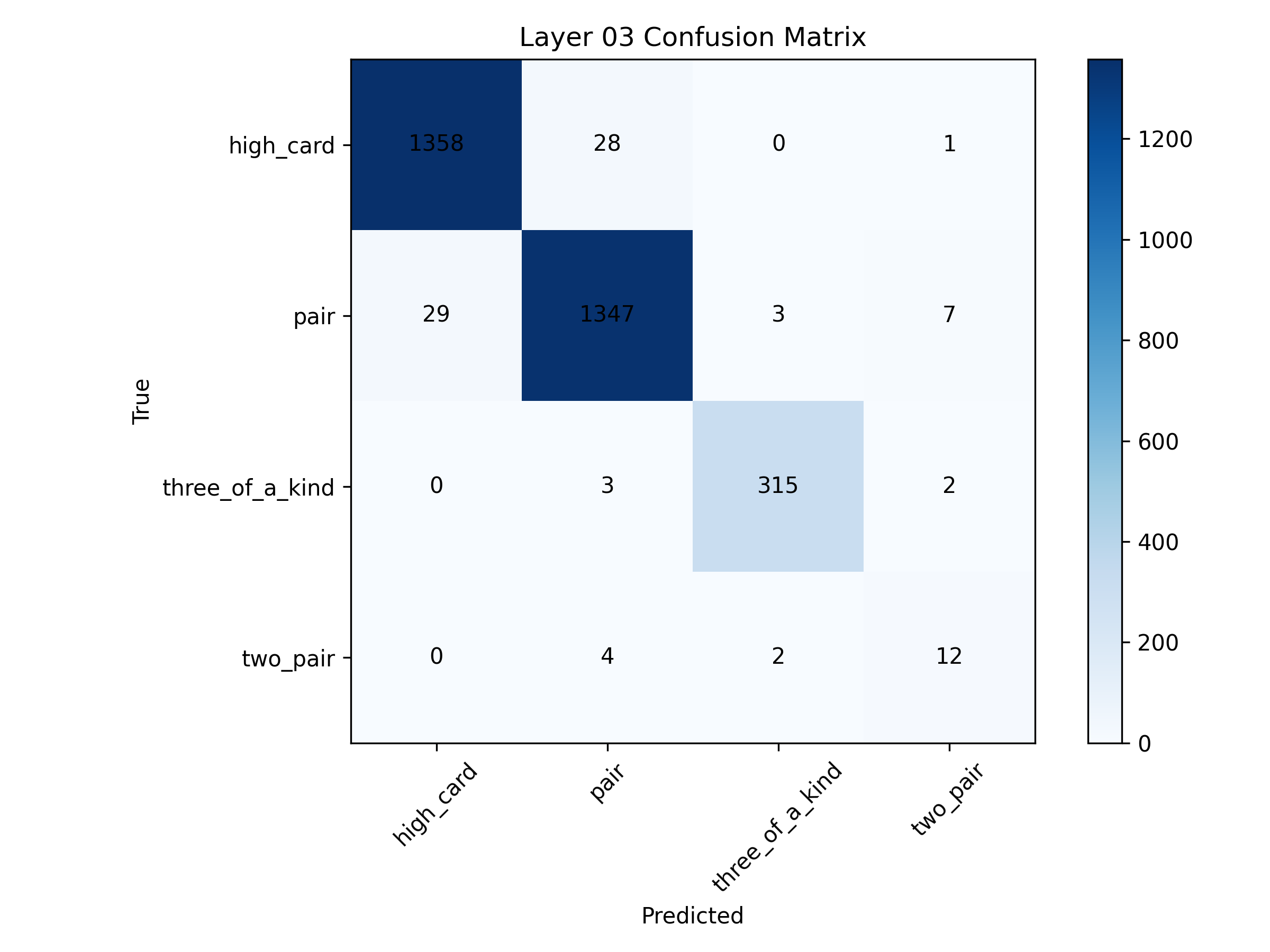}
        \caption{Layer 3}
        \label{fig:layer3_m1}
    \end{subfigure}

    \vspace{0.5em}

    \begin{subfigure}[b]{0.23\textwidth}
        \centering
        \includegraphics[width=\textwidth]{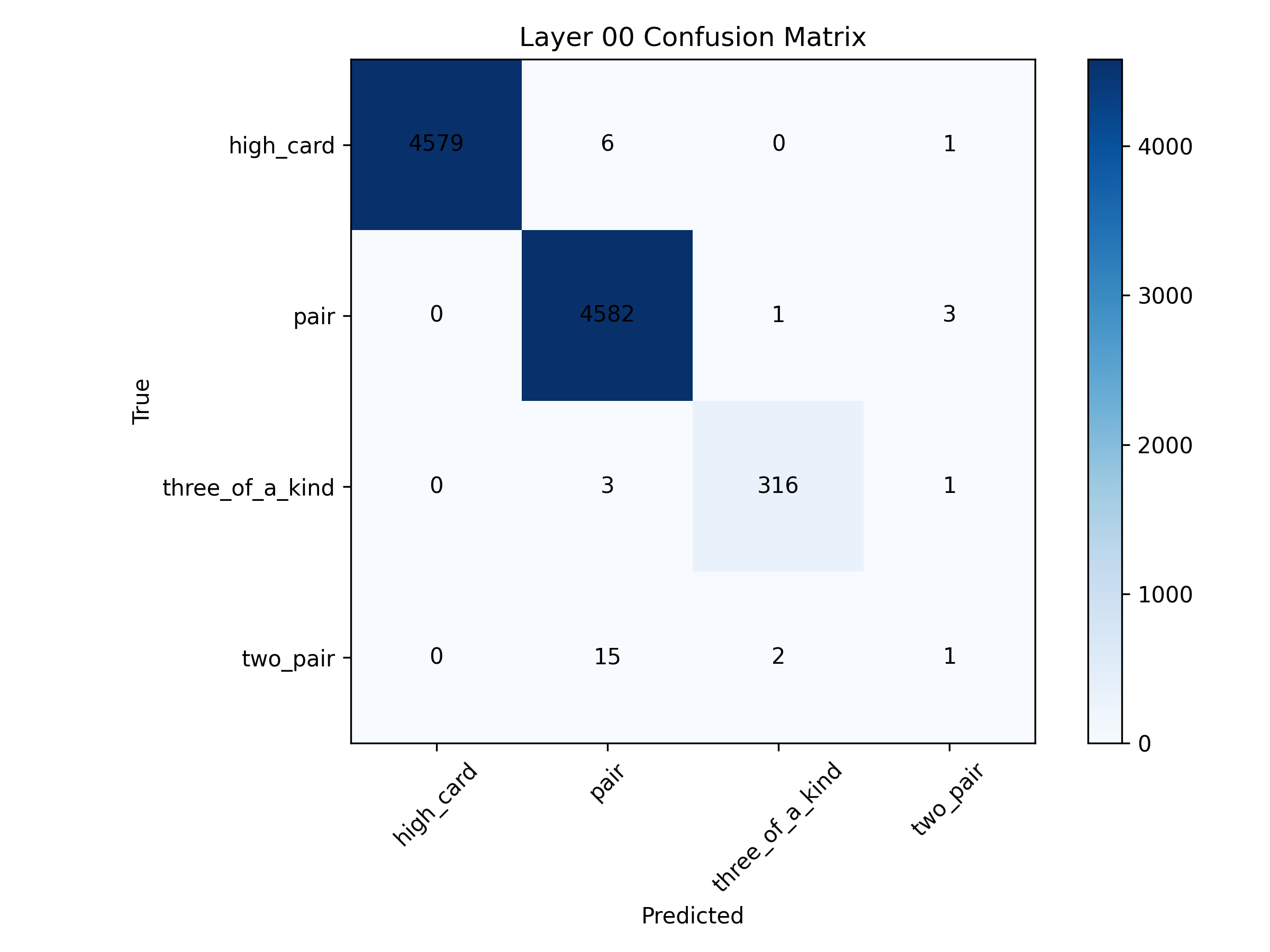}
        \caption{Layer 0}
        \label{fig:layer0_m2}
    \end{subfigure}
    \hfill
    \begin{subfigure}[b]{0.23\textwidth}
        \centering
        \includegraphics[width=\textwidth]{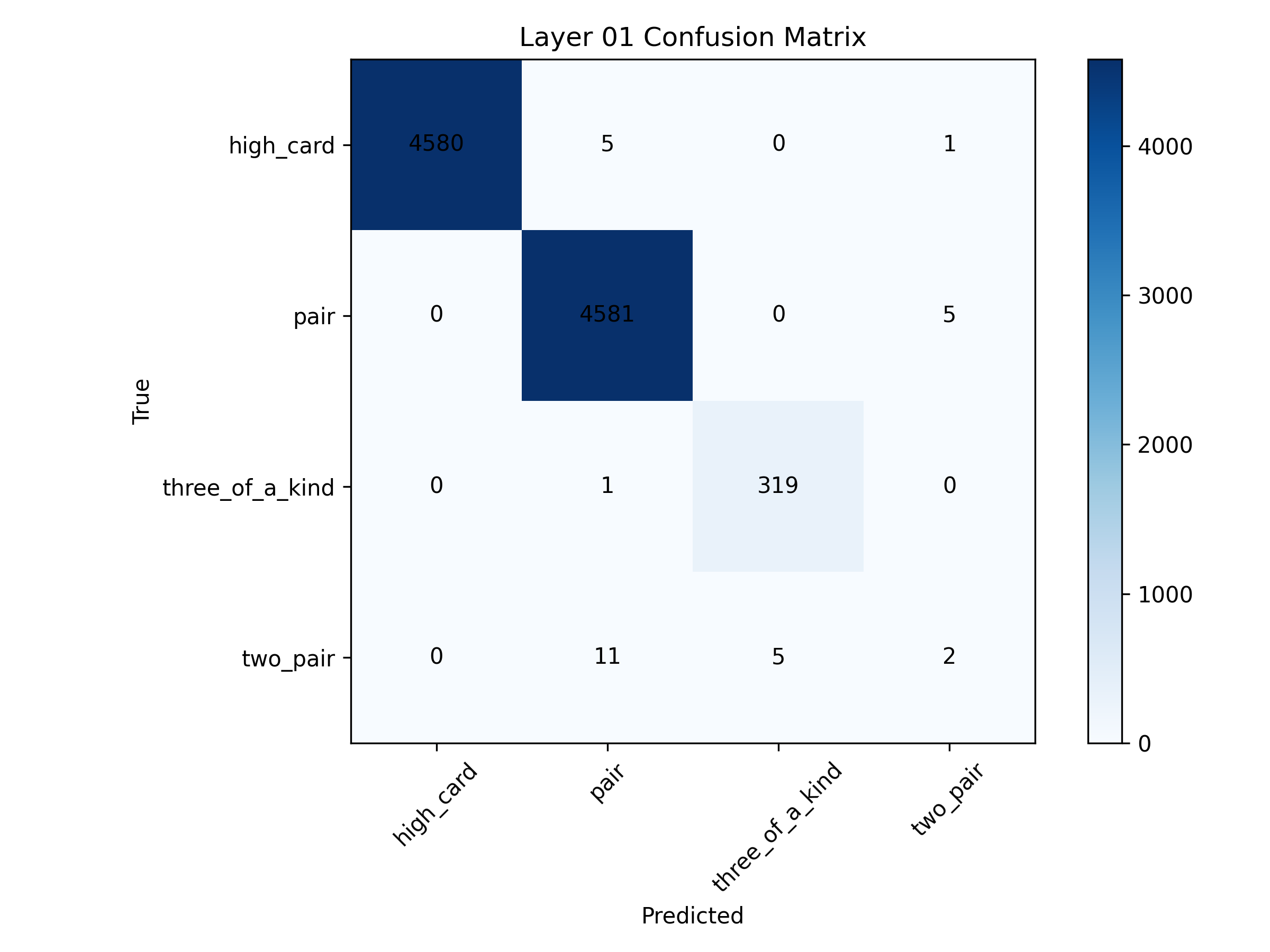}
        \caption{Layer 1}
        \label{fig:layer1_m2}
    \end{subfigure}
    \hfill
    \begin{subfigure}[b]{0.23\textwidth}
        \centering
        \includegraphics[width=\textwidth]{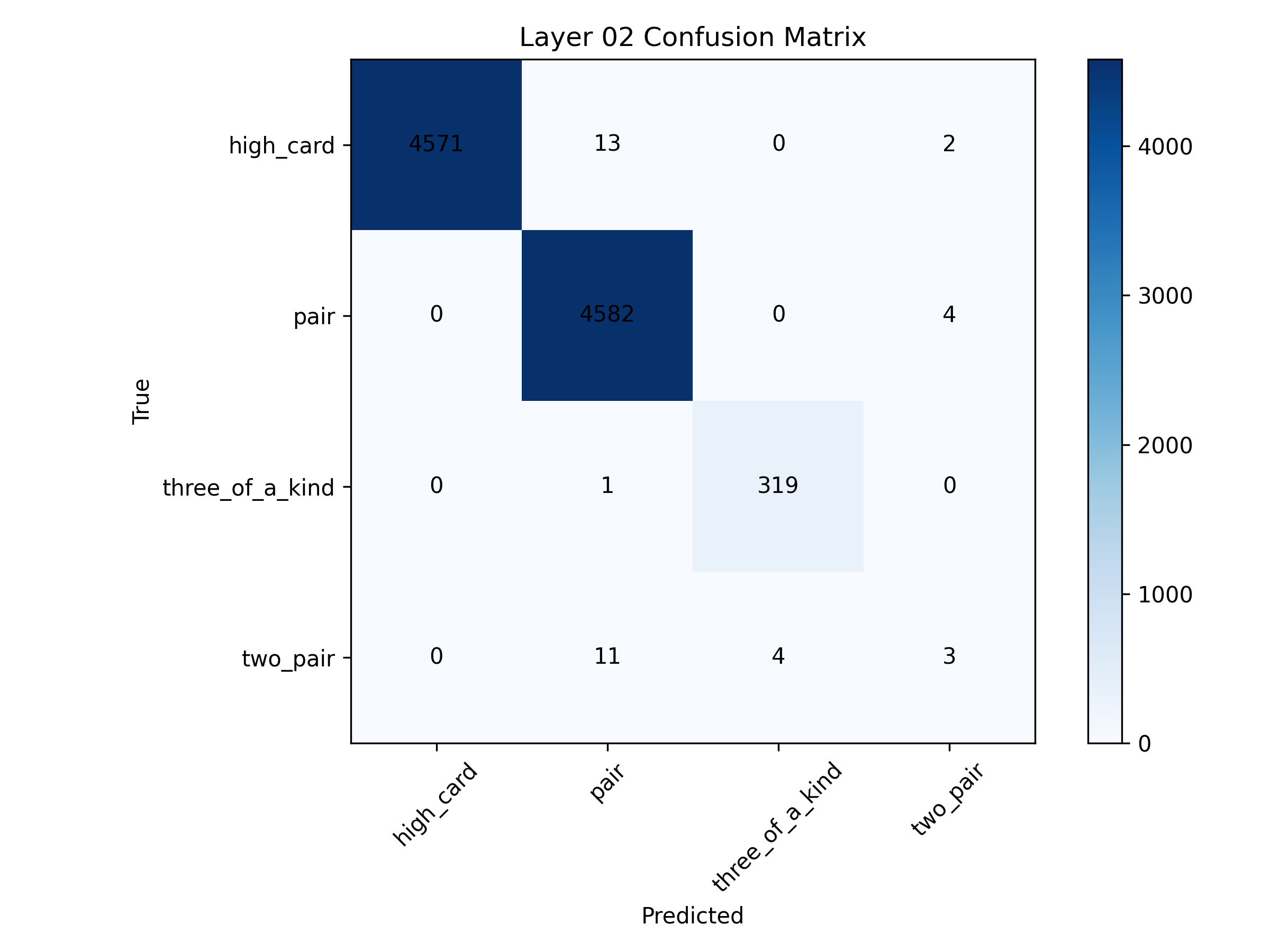}
        \caption{Layer 2}
        \label{fig:layer2_m2}
    \end{subfigure}
    \hfill
    \begin{subfigure}[b]{0.23\textwidth}
        \centering
        \includegraphics[width=\textwidth]{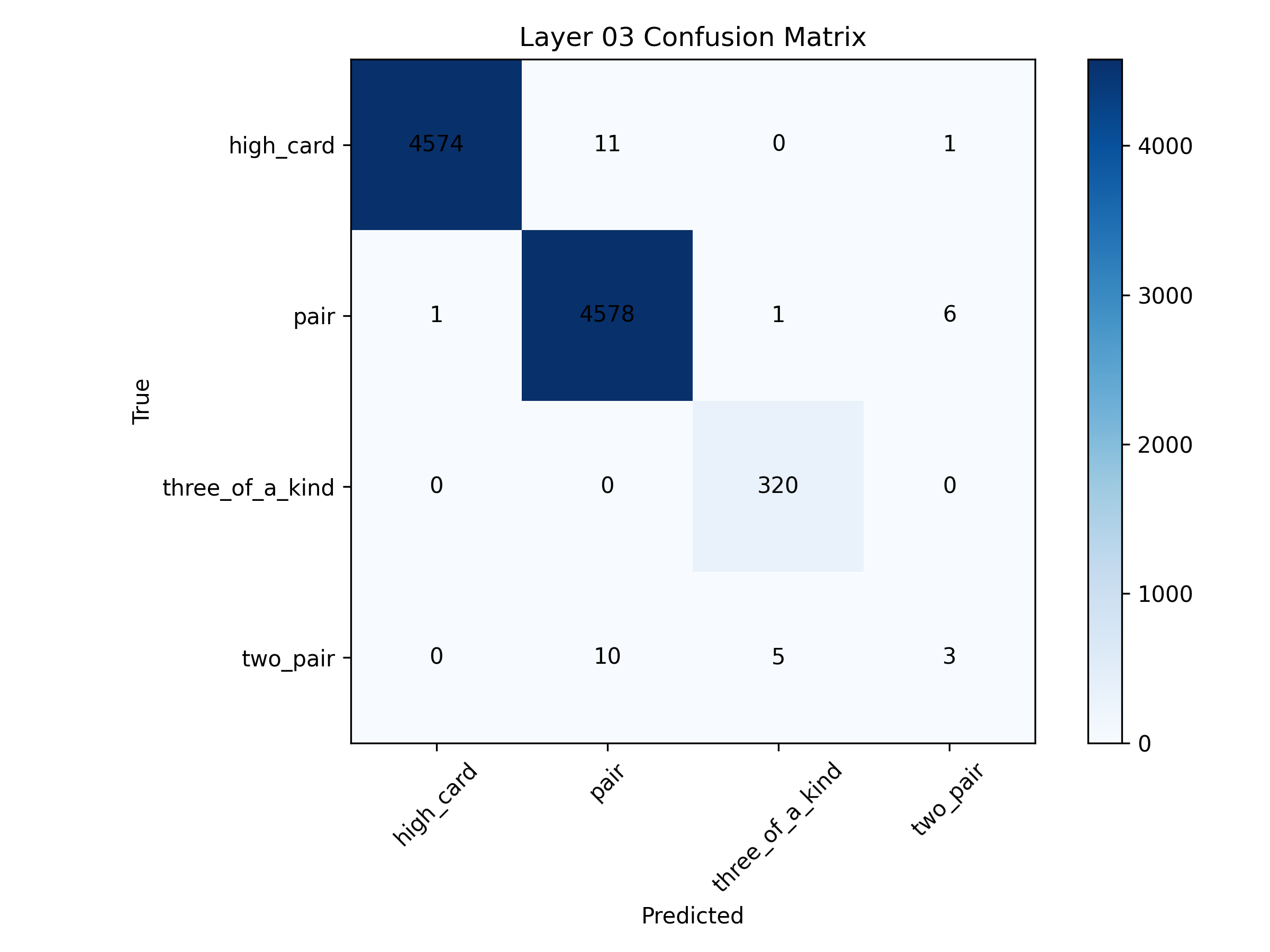}
        \caption{Layer 3}
        \label{fig:layer3_m2}
    \end{subfigure}

    \caption{Hand-rank identification results across transformer Layers 0–3
    using linear and MLP probes. Each confusion matrix shows probe predictions for the categorical
    poker hand-rank (e.g., high card, pair, two pair, etc.), evaluated on a
    held-out test set excluded from both model and probe training. Row 1 shows
    linear probe performance. Rows 2–4 show MLP probe performance when the
    training data is balanced at the 30th, 35th, and 40th percentiles of unique
    hand-rank frequencies, respectively, ensuring that rare hand-ranks are not
    underrepresented. Across all configurations, strong diagonals indicated by the dark coloring show consistent internal encoding of hand-rank information by early layers, while
    differences across percentiles displays the effect of dataset balancing for rarer hand-ranks.}
    \label{fig:activation_4x4}
\end{figure}

\section{Activation Plots}
\label{plots}
Below are our activation plots (compressed with PCA, t-SNE, and UMAP, respectively, to 2-D plots) across multiple model layers. We note the per-hand clusters (with each cluster representing a particular "type" of hand, ie. a pair with certain card values, or a three of a kind with a certain card type). We also note that conceptual similarity is also being represented by these clusters. Note: These plots were generated from a test set of  \textasciitilde{}200,000 samples. This set is separate from the training set used for the model and the one used for the probes.
PCA activation analysis revealed triangular activation structures that closely resemble 
the belief-state geometries described by \citet{shai2024beliefgeometry}. A natural 
direction for future work is to investigate whether these structures reflect the model’s 
implicit representation of belief states in a POMDP setting. In particular, the vertices 
of the triangle may correspond to pure beliefs, confident assignments to specific hand 
ranks, while interior points capture mixtures over multiple possibilities. This 
interpretation would suggest that the model has learned to encode uncertainty in a manner 
consistent with POMDP belief representations. 

\begin{figure}[ht]
    \centering
    \vspace{-0.75em}
    \begin{subfigure}[b]{0.24\textwidth}\centering
        \includegraphics[width=\linewidth]{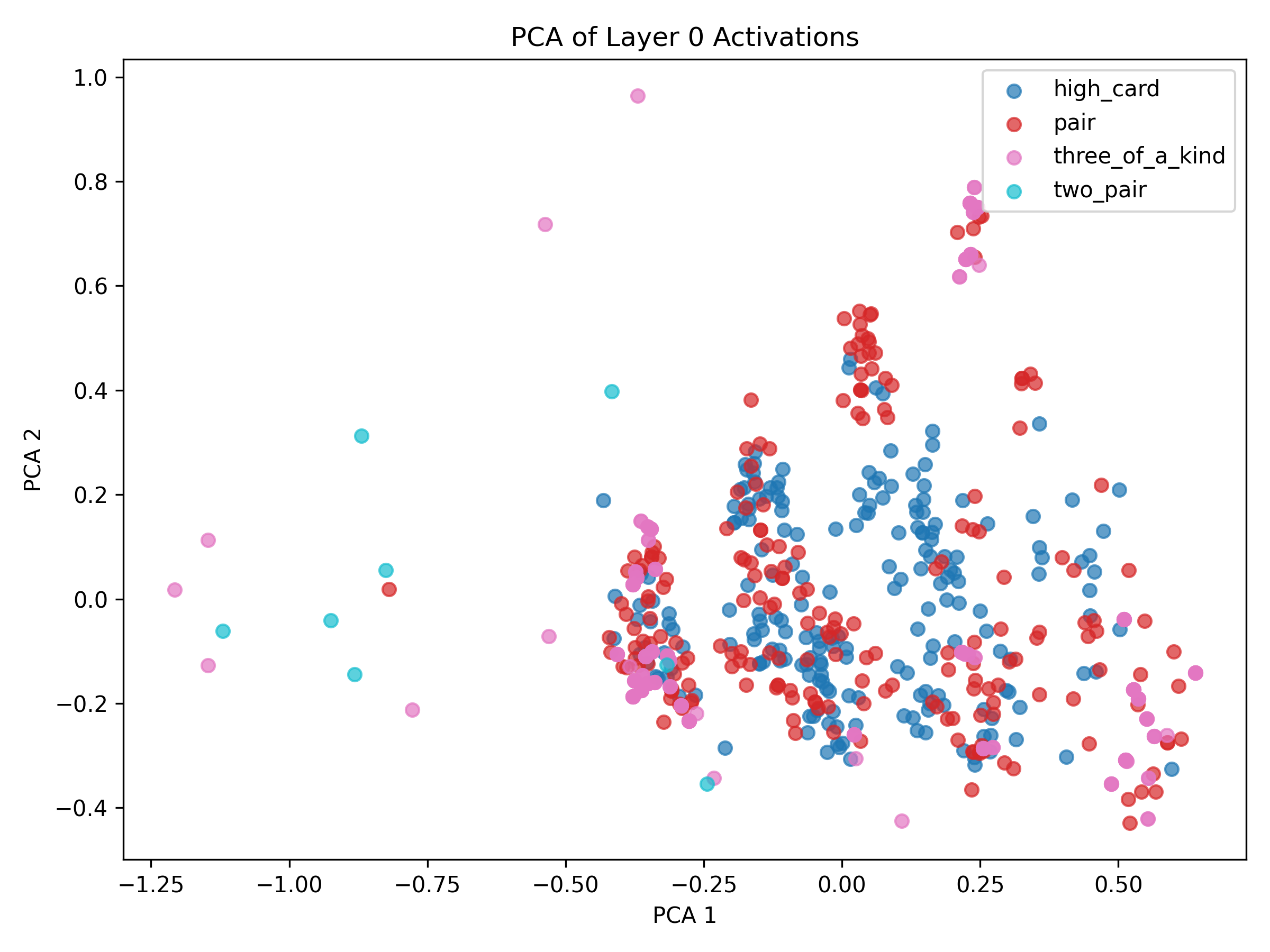}
        \caption{Layer 0}
    \end{subfigure}\hfill
    \begin{subfigure}[b]{0.24\textwidth}\centering
        \includegraphics[width=\linewidth]{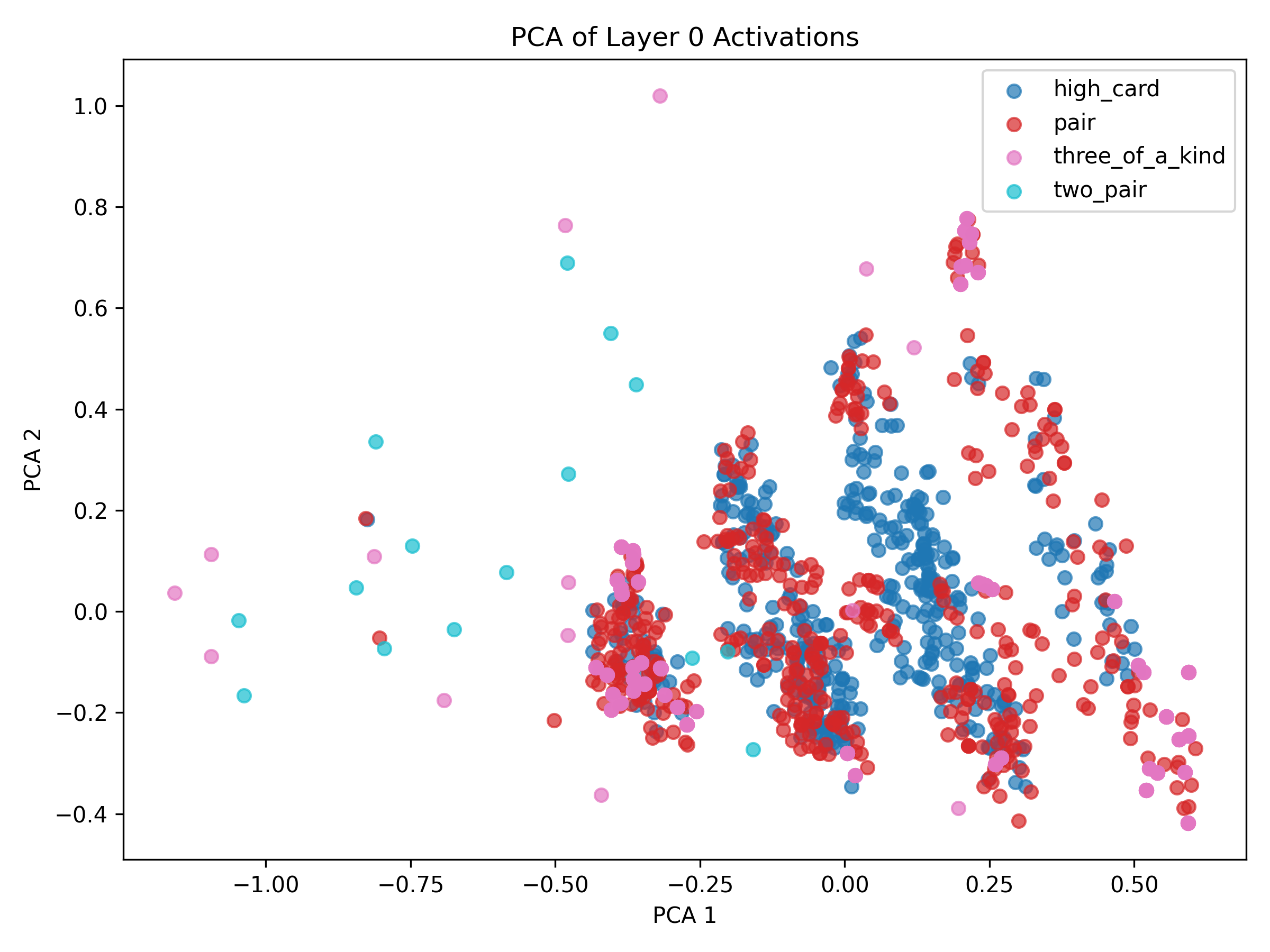}
        \caption{Layer 0}
    \end{subfigure}\hfill
    \begin{subfigure}[b]{0.24\textwidth}\centering
        \includegraphics[width=\linewidth]{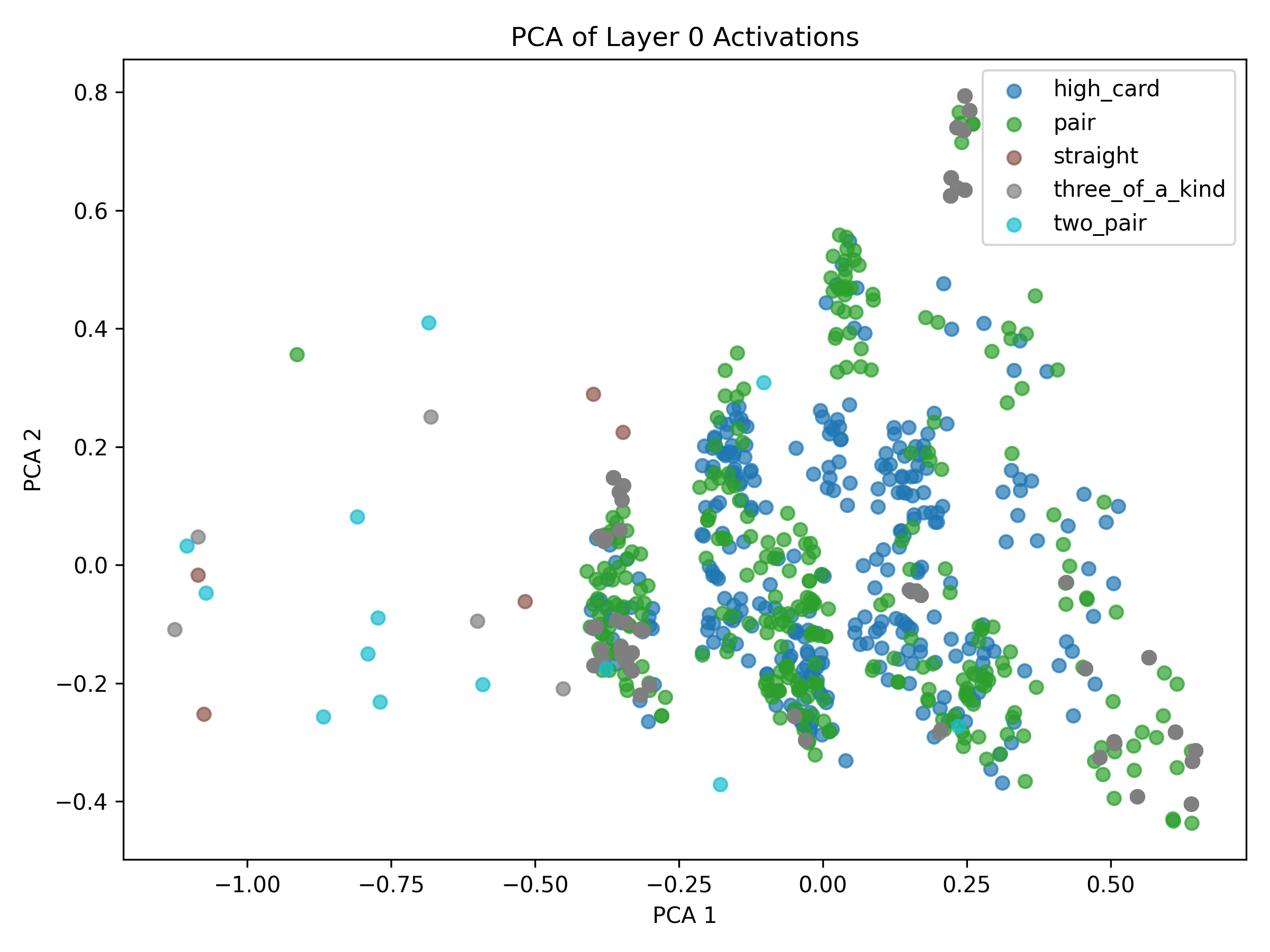}
        \caption{Layer 0}
    \end{subfigure}\hfill
    \begin{subfigure}[b]{0.24\textwidth}\centering
        \includegraphics[width=\linewidth]{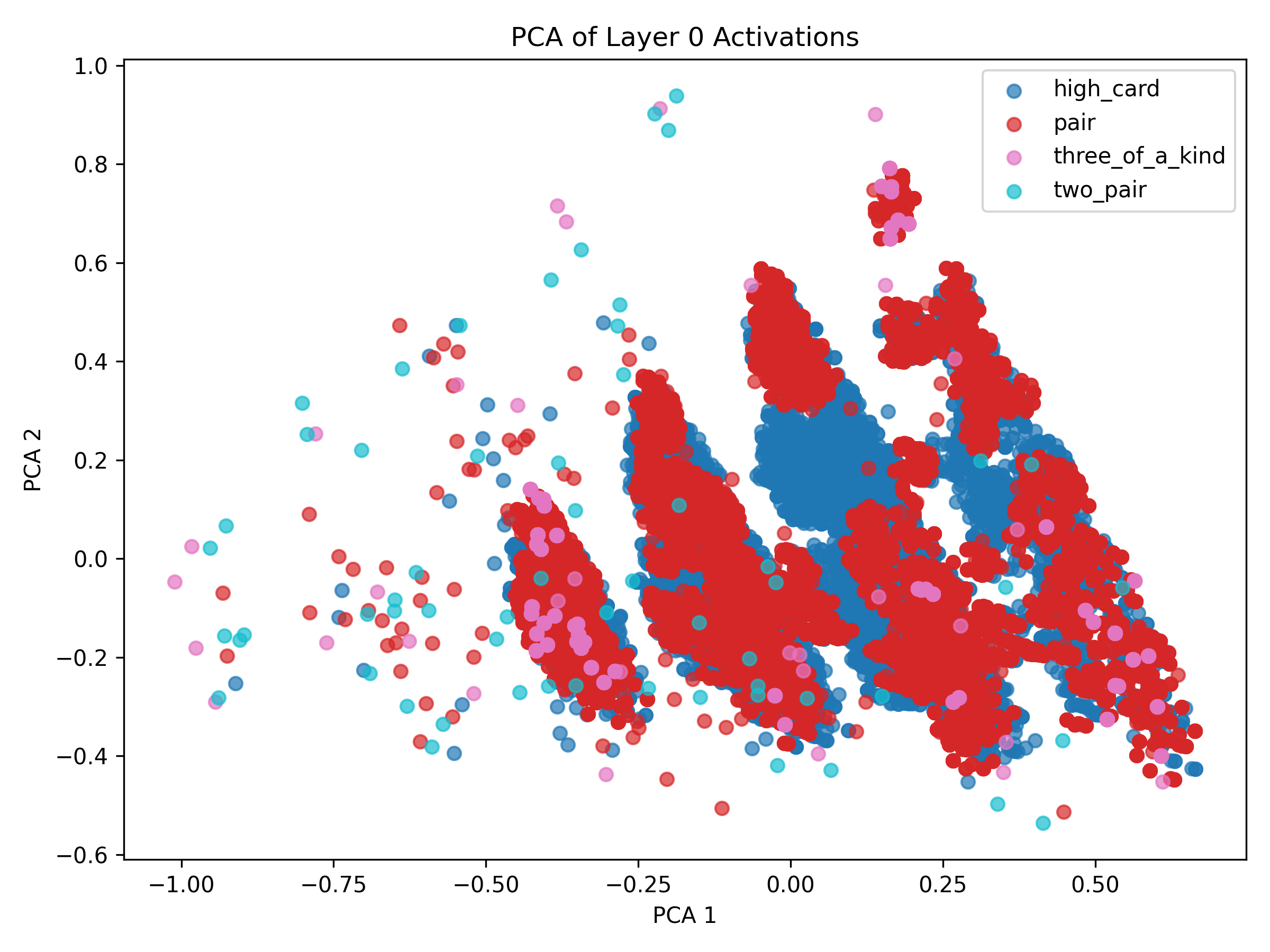}
        \caption{Layer 0}
    \end{subfigure}

    \par\medskip  

    \begin{subfigure}[b]{0.24\textwidth}\centering
        \includegraphics[width=\linewidth]{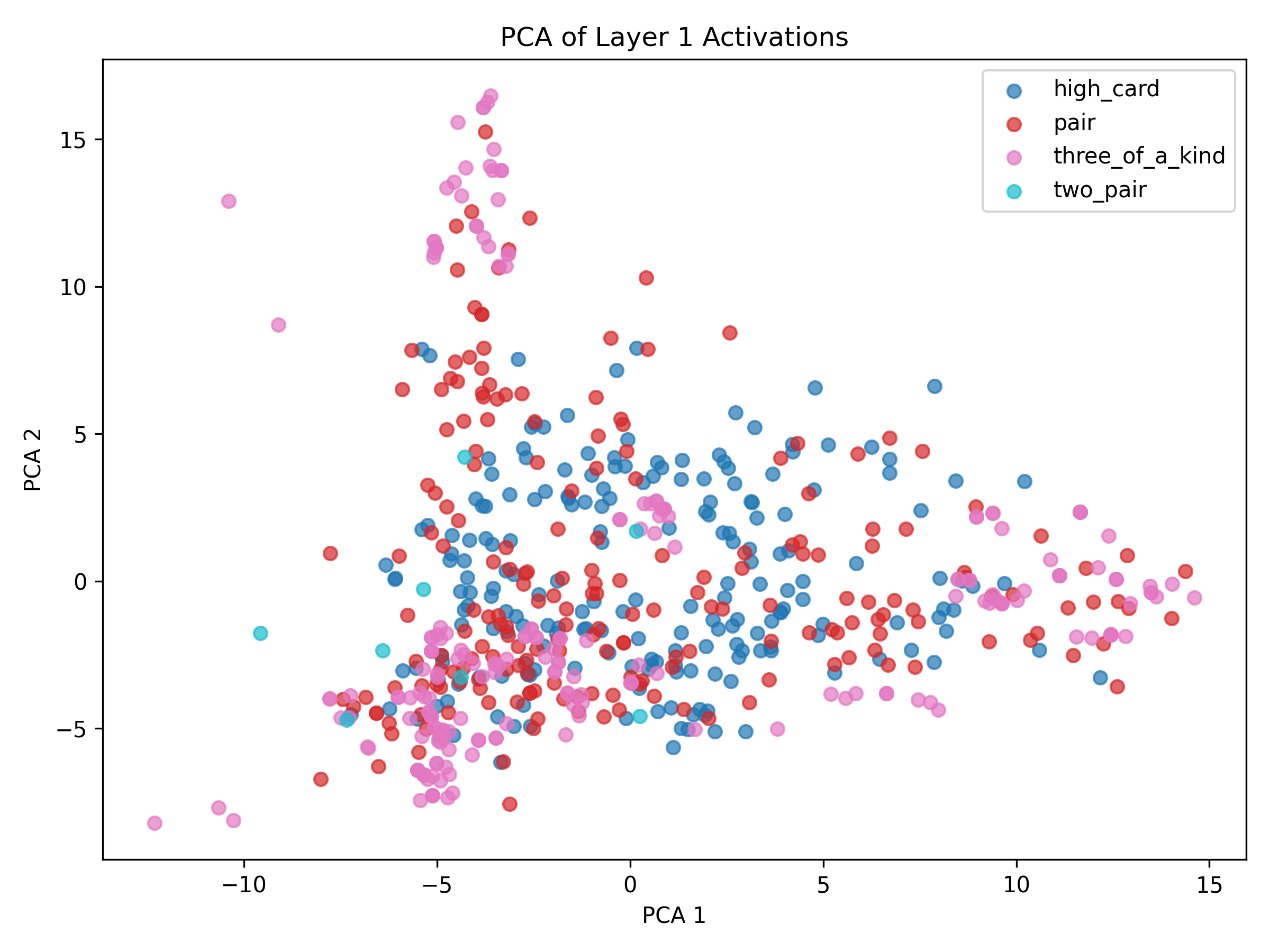}
        \caption{Layer 1}
    \end{subfigure}\hfill
    \begin{subfigure}[b]{0.24\textwidth}\centering
        \includegraphics[width=\linewidth]{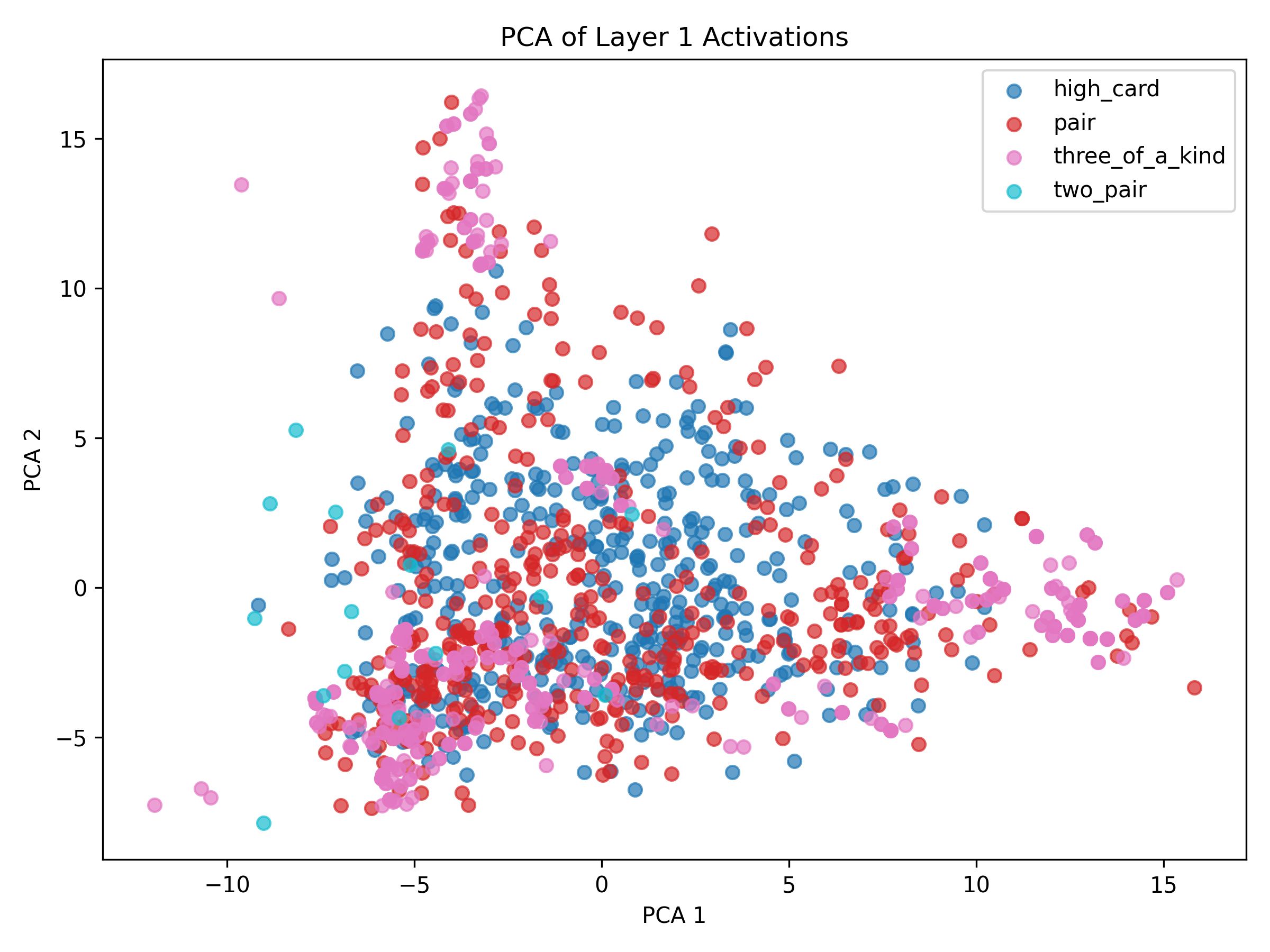}
        \caption{Layer 1}
    \end{subfigure}\hfill
    \begin{subfigure}[b]{0.24\textwidth}\centering
        \includegraphics[width=\linewidth]{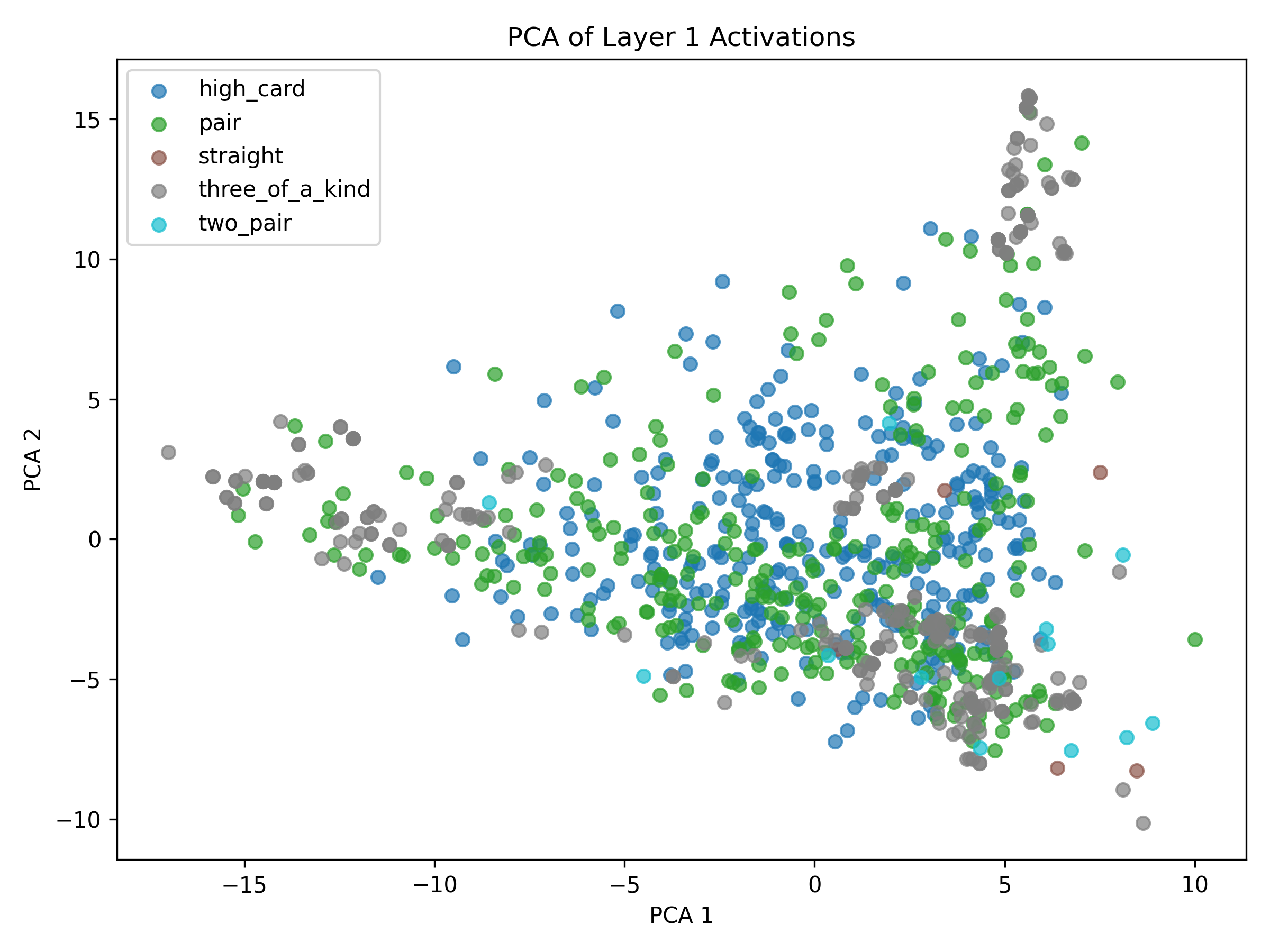}
        \caption{Layer 1}
    \end{subfigure}\hfill
    \begin{subfigure}[b]{0.24\textwidth}\centering
        \includegraphics[width=\linewidth]{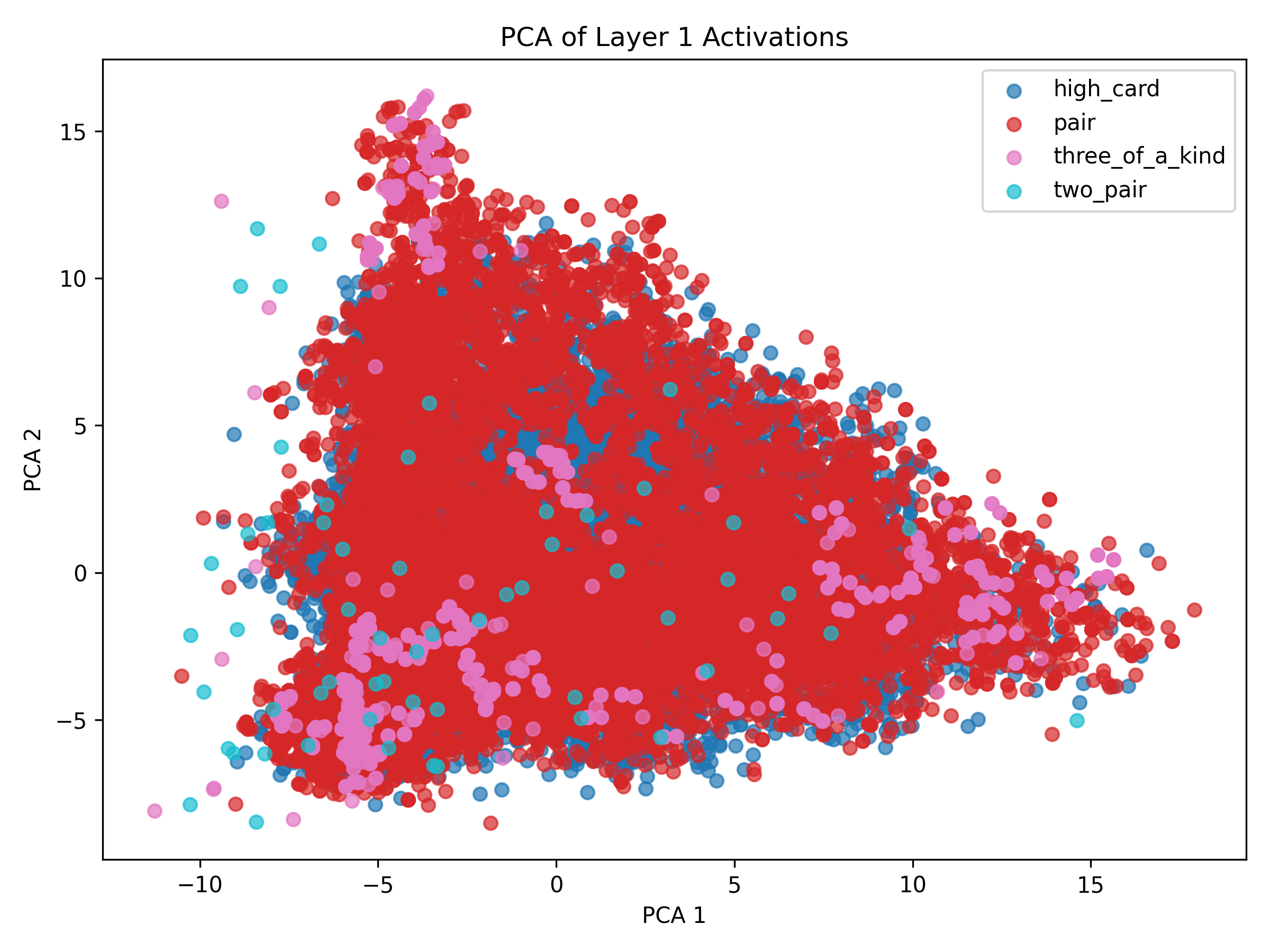}
        \caption{Layer 1}
    \end{subfigure}

    \par\medskip

    \begin{subfigure}[b]{0.24\textwidth}\centering
        \includegraphics[width=\linewidth]{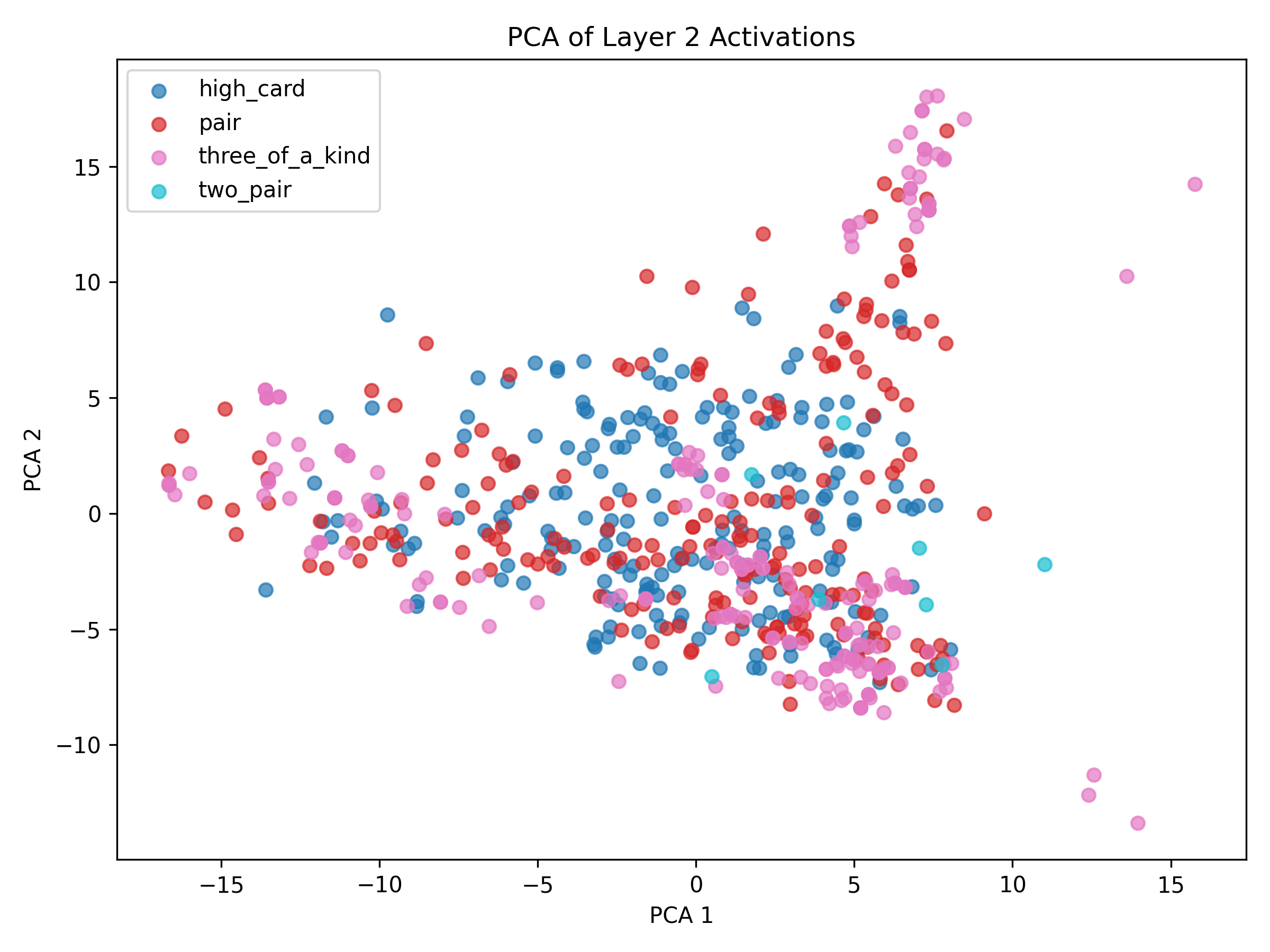}
        \caption{Layer 2}
    \end{subfigure}\hfill
    \begin{subfigure}[b]{0.24\textwidth}\centering
        \includegraphics[width=\linewidth]{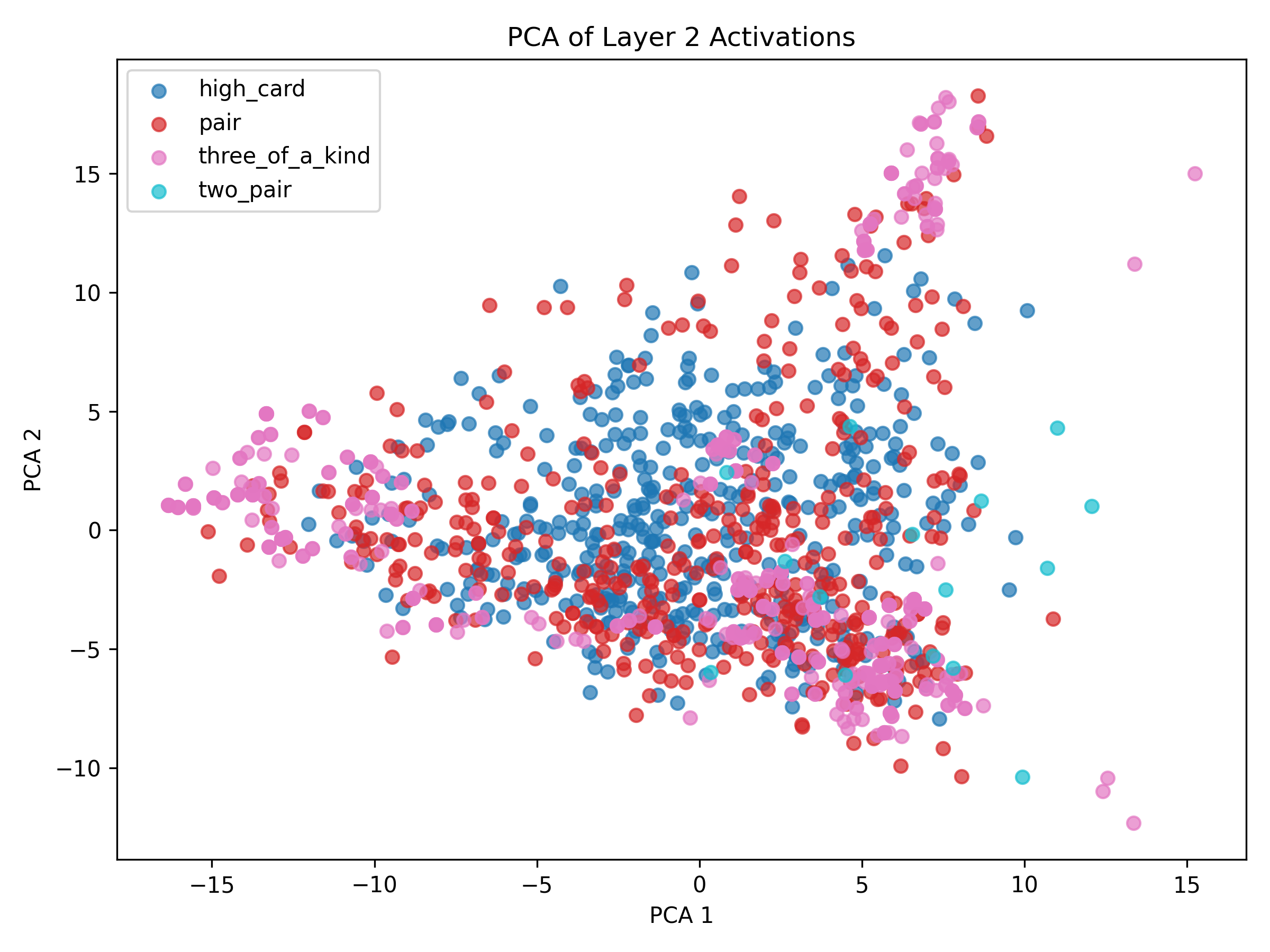}
        \caption{Layer 2}
    \end{subfigure}\hfill
    \begin{subfigure}[b]{0.24\textwidth}\centering
        \includegraphics[width=\linewidth]{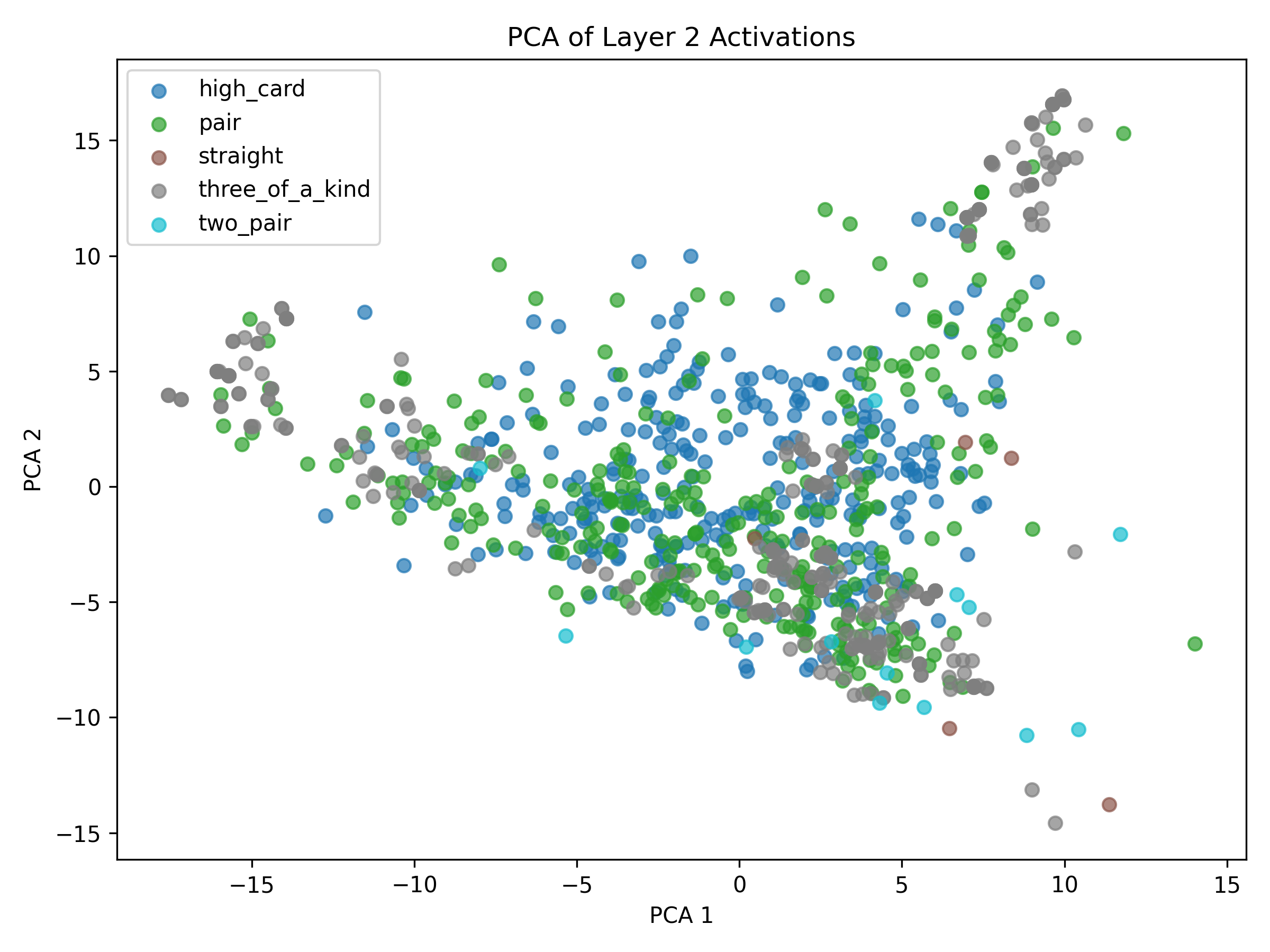}
        \caption{Layer 2}
    \end{subfigure}\hfill
    \begin{subfigure}[b]{0.24\textwidth}\centering
        \includegraphics[width=\linewidth]{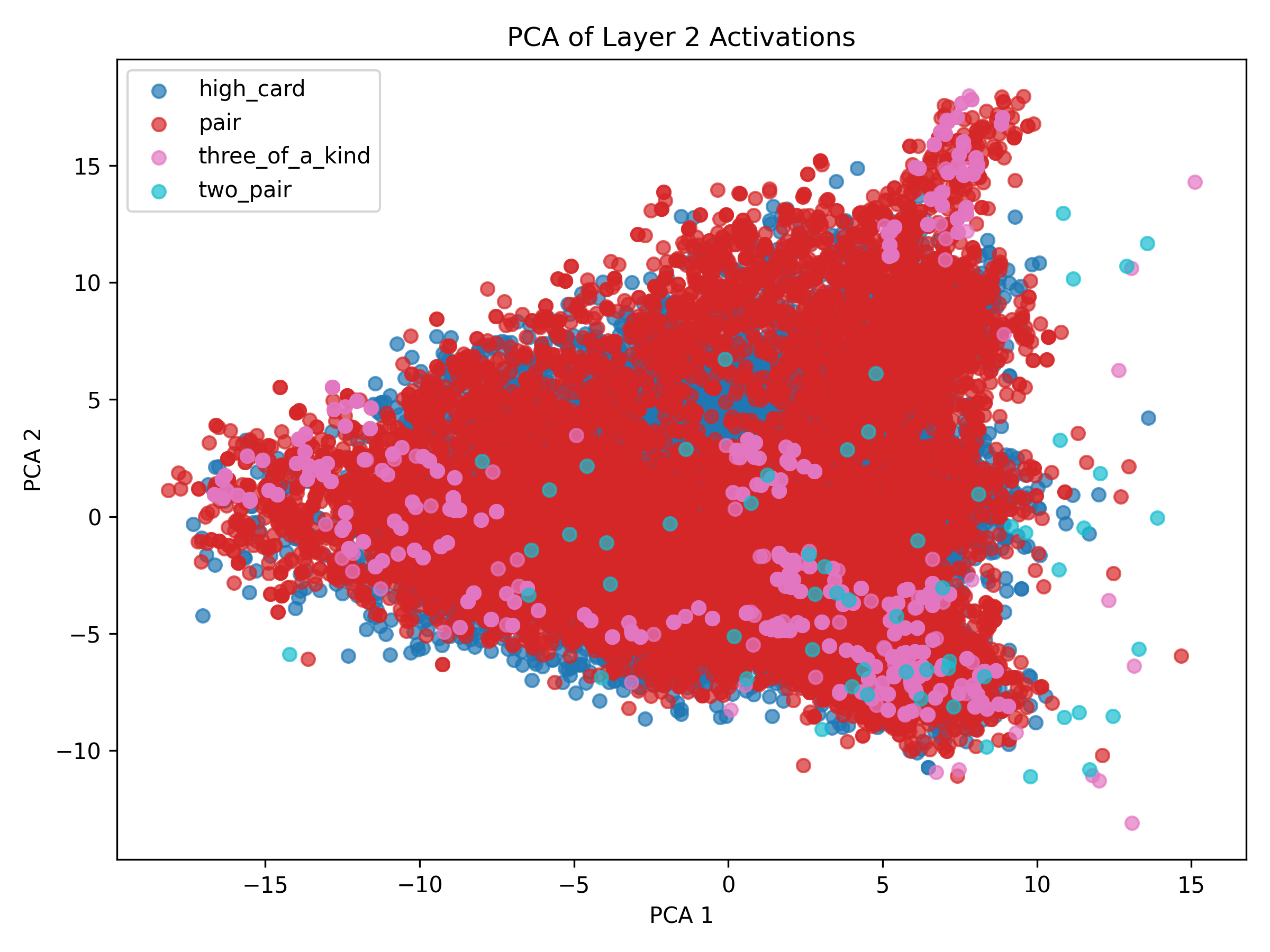}
        \caption{Layer 2}
    \end{subfigure}

    \par\medskip

    \begin{subfigure}[b]{0.24\textwidth}\centering
        \includegraphics[width=\linewidth]{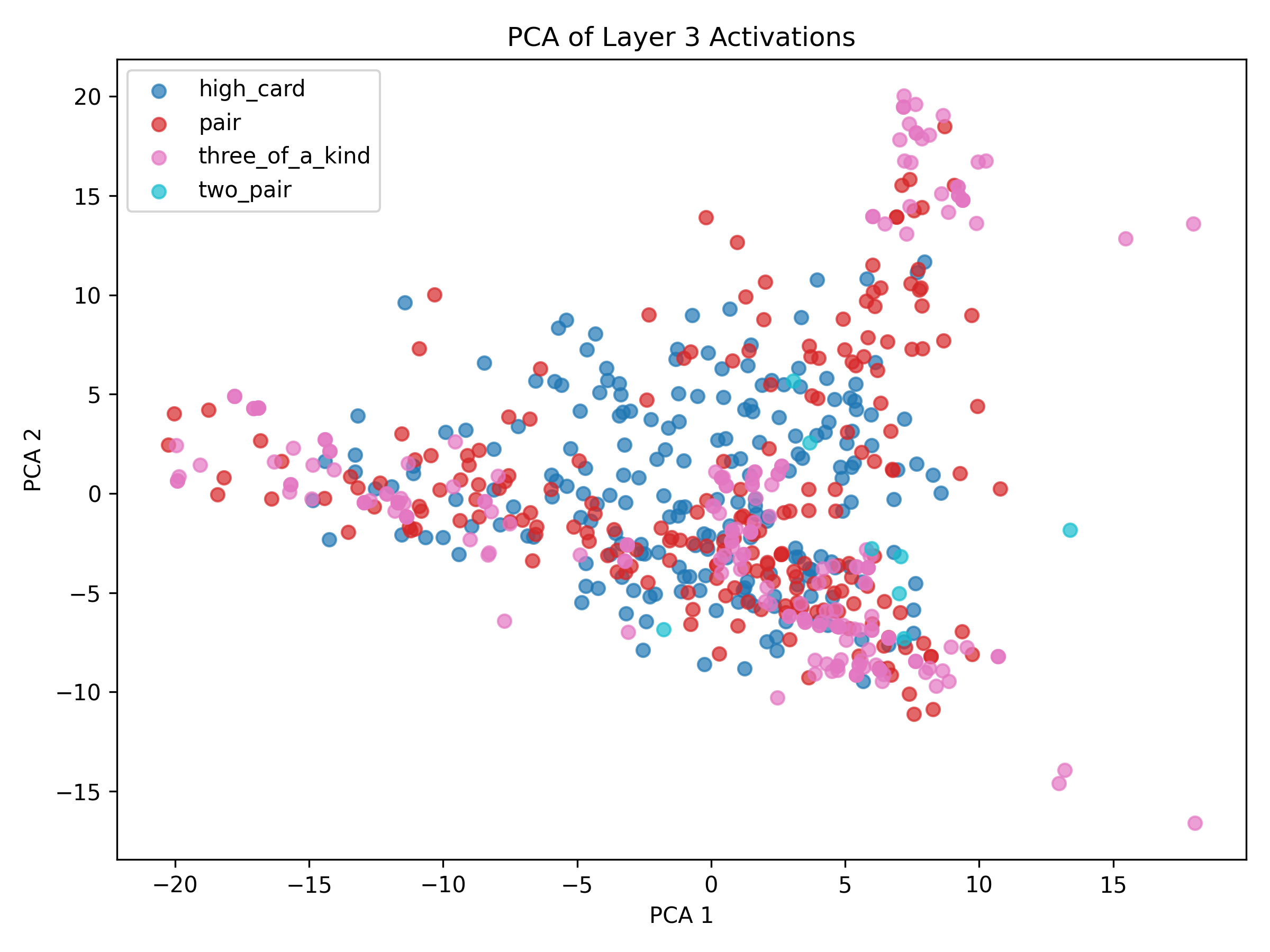}
        \caption{Layer 3}
    \end{subfigure}\hfill
    \begin{subfigure}[b]{0.24\textwidth}\centering
        \includegraphics[width=\linewidth]{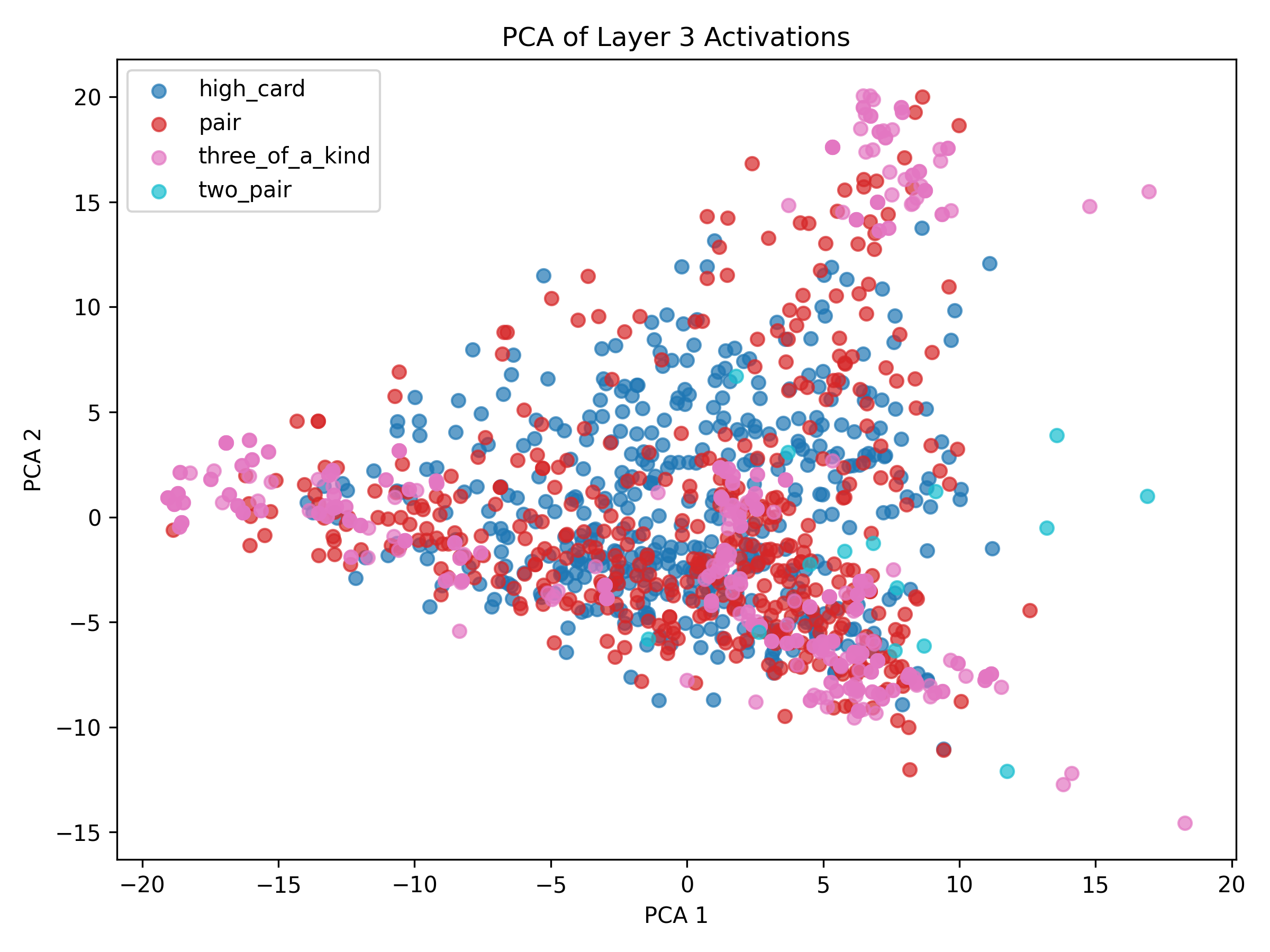}
        \caption{Layer 3}
    \end{subfigure}\hfill
    \begin{subfigure}[b]{0.24\textwidth}\centering
        \includegraphics[width=\linewidth]{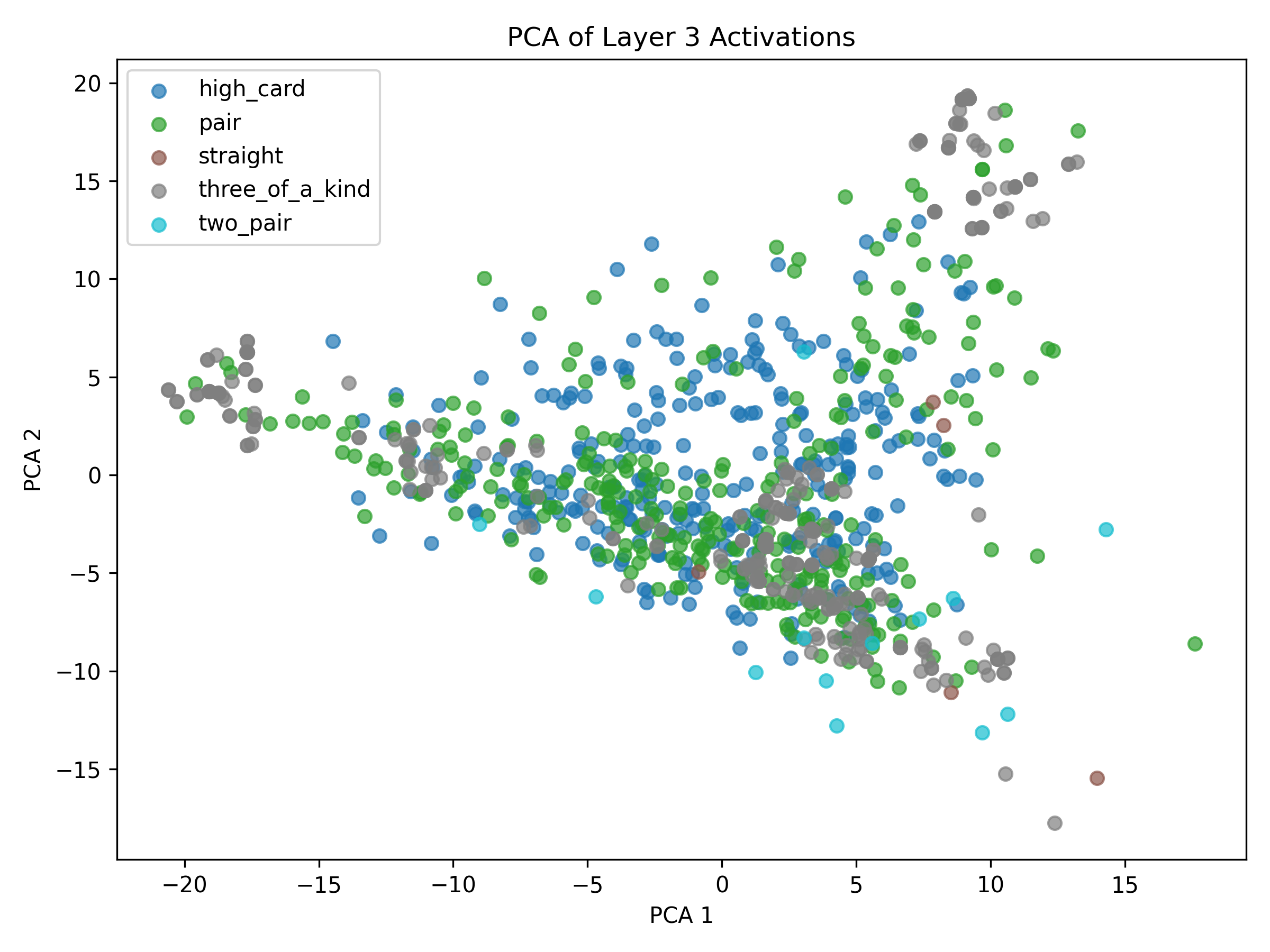}
        \caption{Layer 3}
    \end{subfigure}\hfill
    \begin{subfigure}[b]{0.24\textwidth}\centering
        \includegraphics[width=\linewidth]{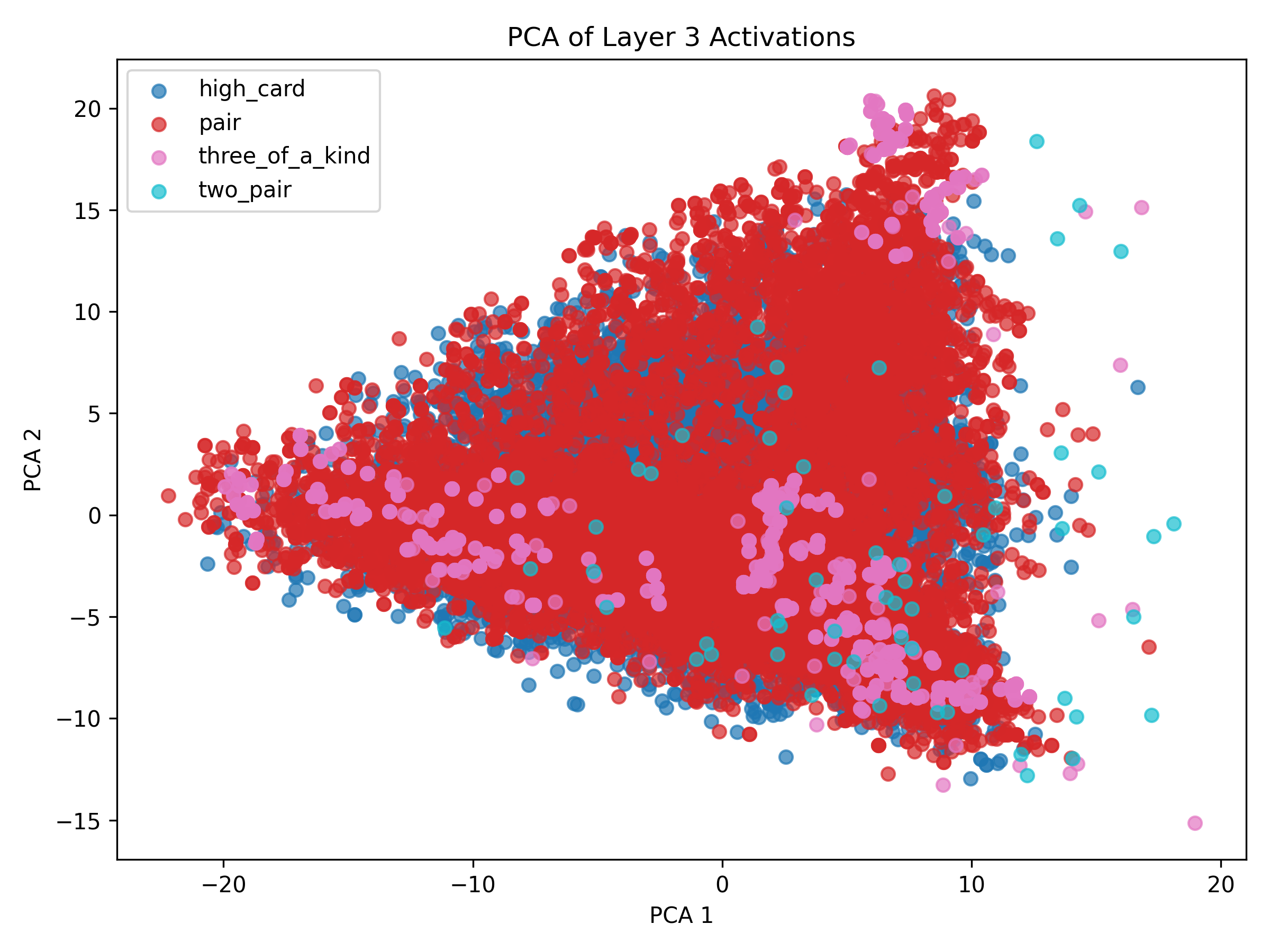}
        \caption{Layer 3}
    \end{subfigure}

    \caption{
        PCA projections of activation vectors across transformer Layers 0–3 and across four 
        different training set sizes (columns: 100k, 200k, 200k with modified test split, 
        and 430k samples). Each panel shows a 2D PCA embedding of per-token activations 
        colored by hand-rank class. The recurring triangular geometry resembles POMDP 
        belief-state manifolds and becomes increasingly well-separated at larger training 
        sizes, indicating stronger representational structure.
    }
    \label{fig:pca_grid}
\end{figure}
The t-SNE visualizations below further illustrate this structure. Each subplot shows a
two-dimensional embedding of activation vectors colored by hand-rank class. Compared to PCA,
t-SNE produces sharper and more clearly separated clusters, revealing that the model internally
organizes hands by both rank and conceptual similarity. In particular, pair and
three-of-a-kind categories form distinct, compact regions, while more ambiguous hands occupy
the intermediate space, reflecting graded internal beliefs about hand strength.
\medskip

\begin{figure}[H]
    \centering
    
    \label{fig:tsne_grid}

    \begin{subfigure}[b]{0.24\textwidth}\centering
        \includegraphics[width=\textwidth]{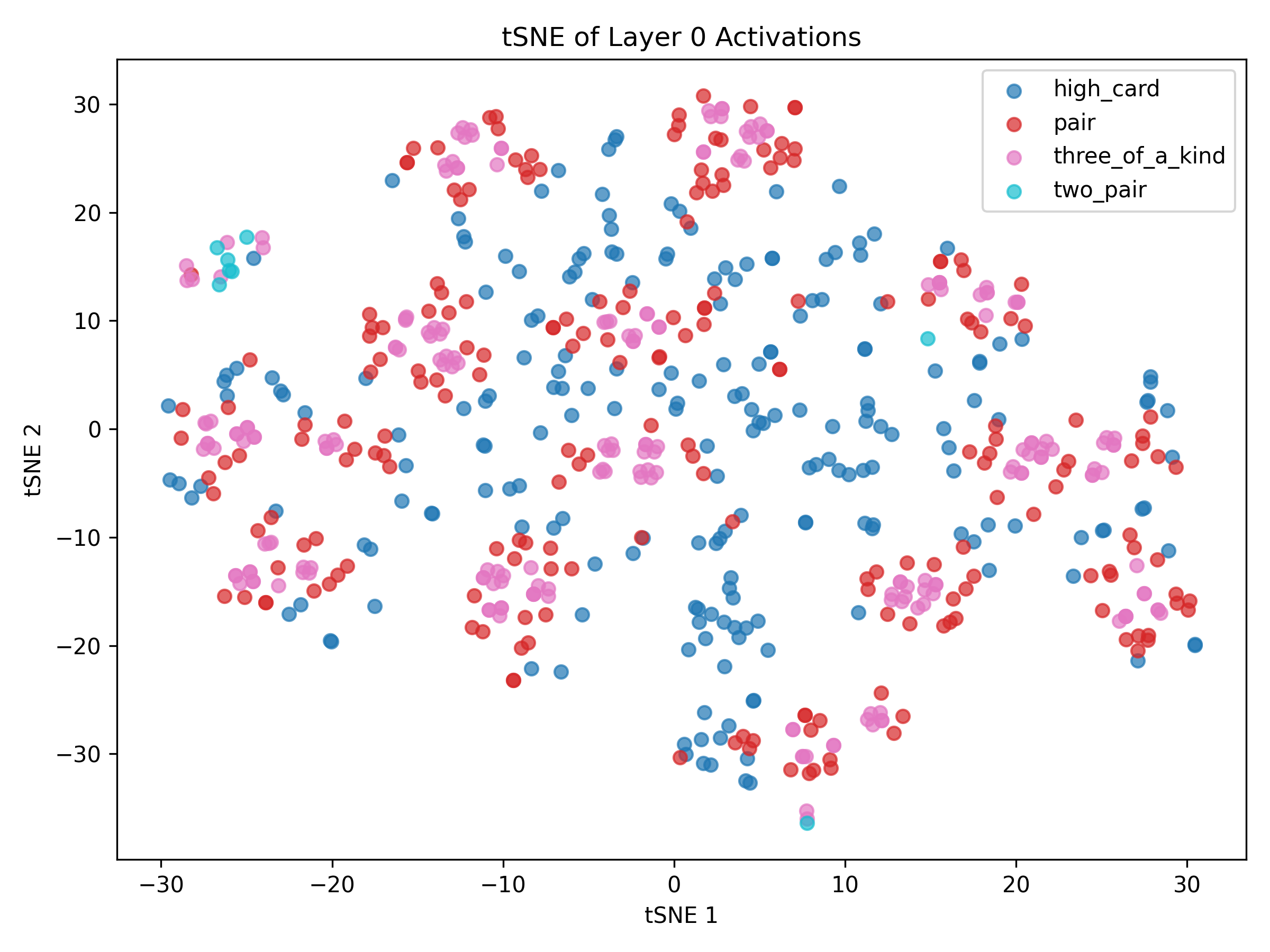}
        \caption{Layer 0}
    \end{subfigure}\hfill
    \begin{subfigure}[b]{0.24\textwidth}\centering
        \includegraphics[width=\textwidth]{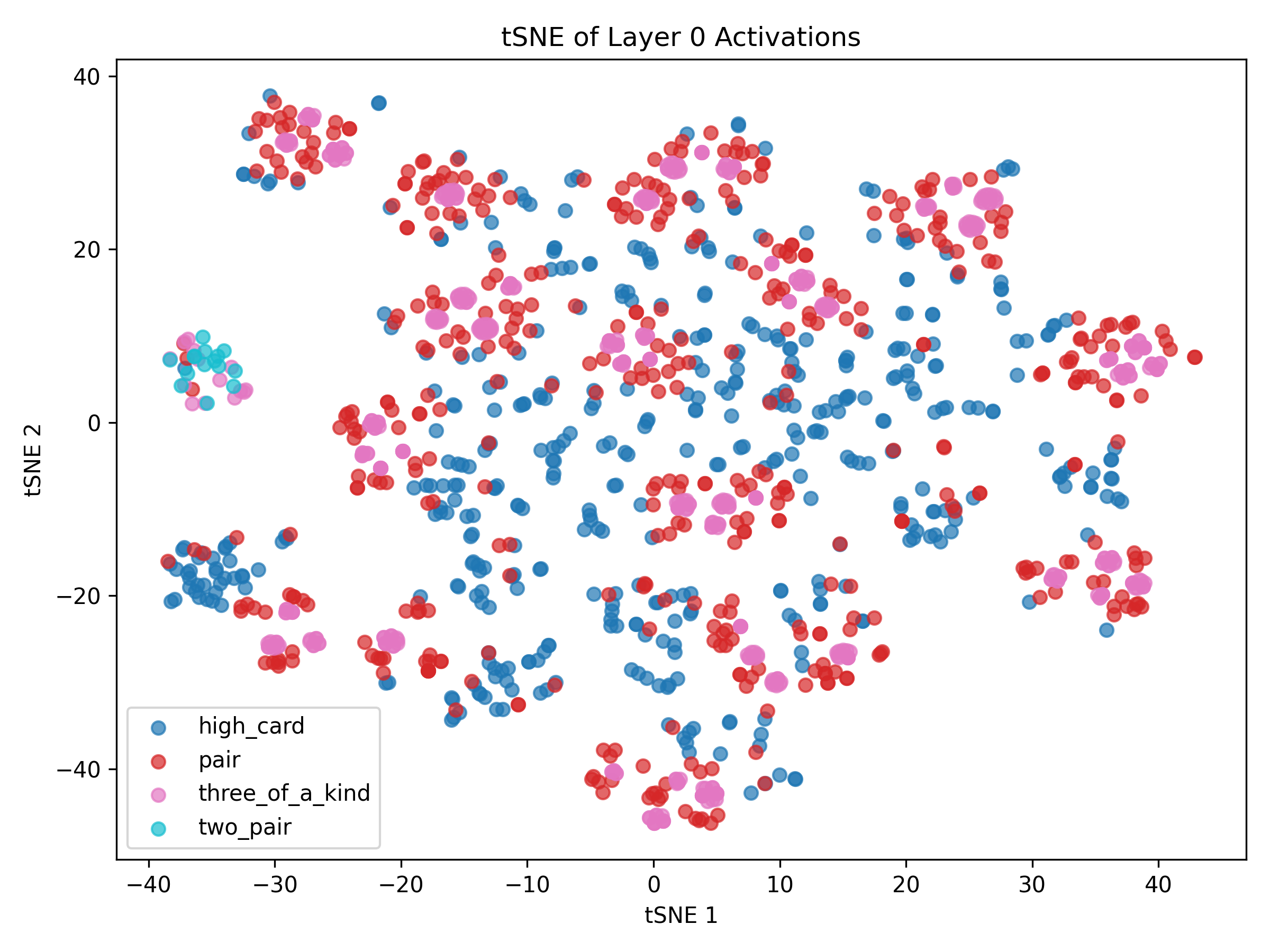}
        \caption{Layer 0}
    \end{subfigure}\hfill
    \begin{subfigure}[b]{0.24\textwidth}\centering
        \includegraphics[width=\textwidth]{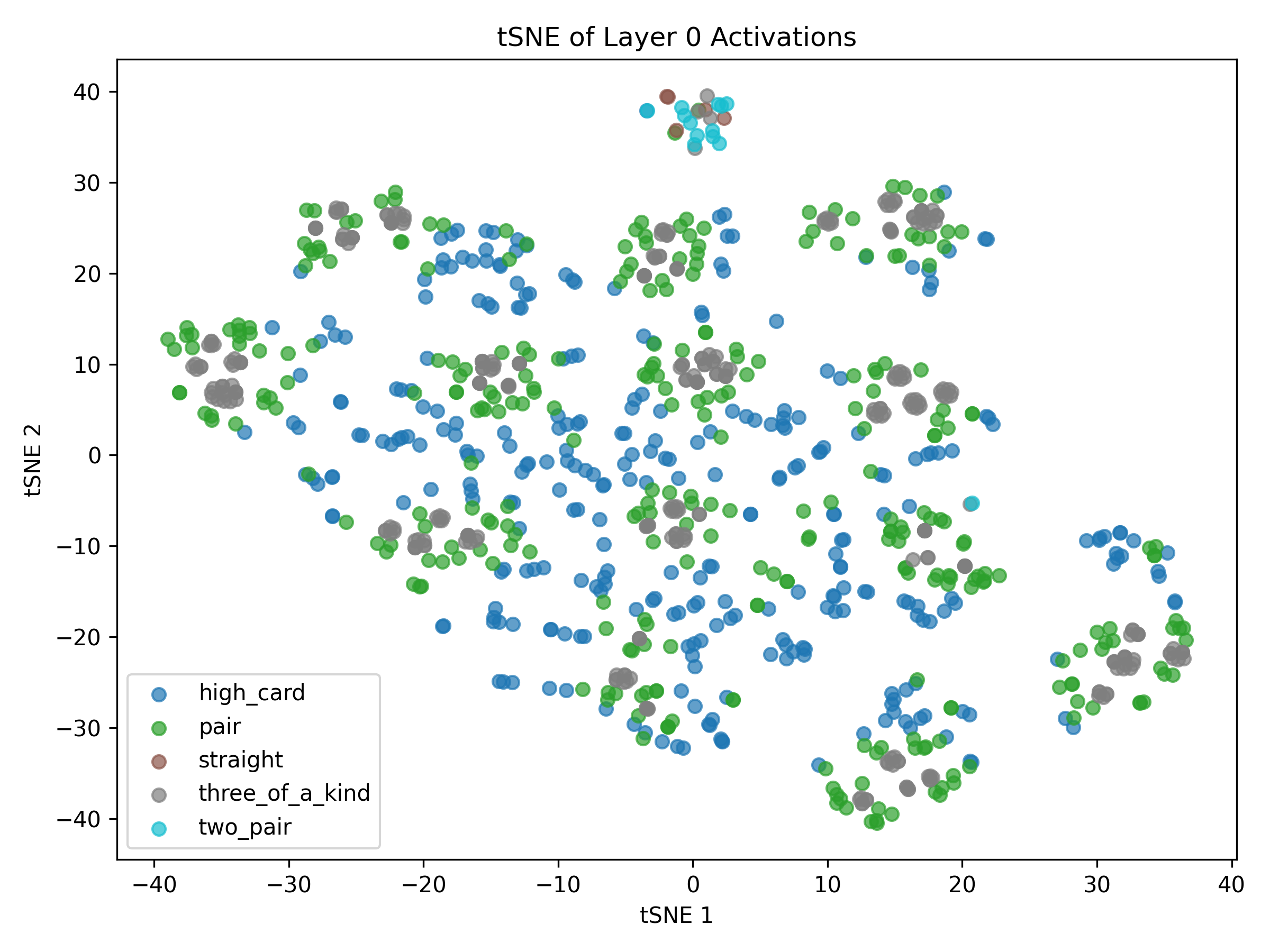}
        \caption{Layer 0}
    \end{subfigure}\hfill
    \begin{subfigure}[b]{0.24\textwidth}\centering
        \includegraphics[width=\textwidth]{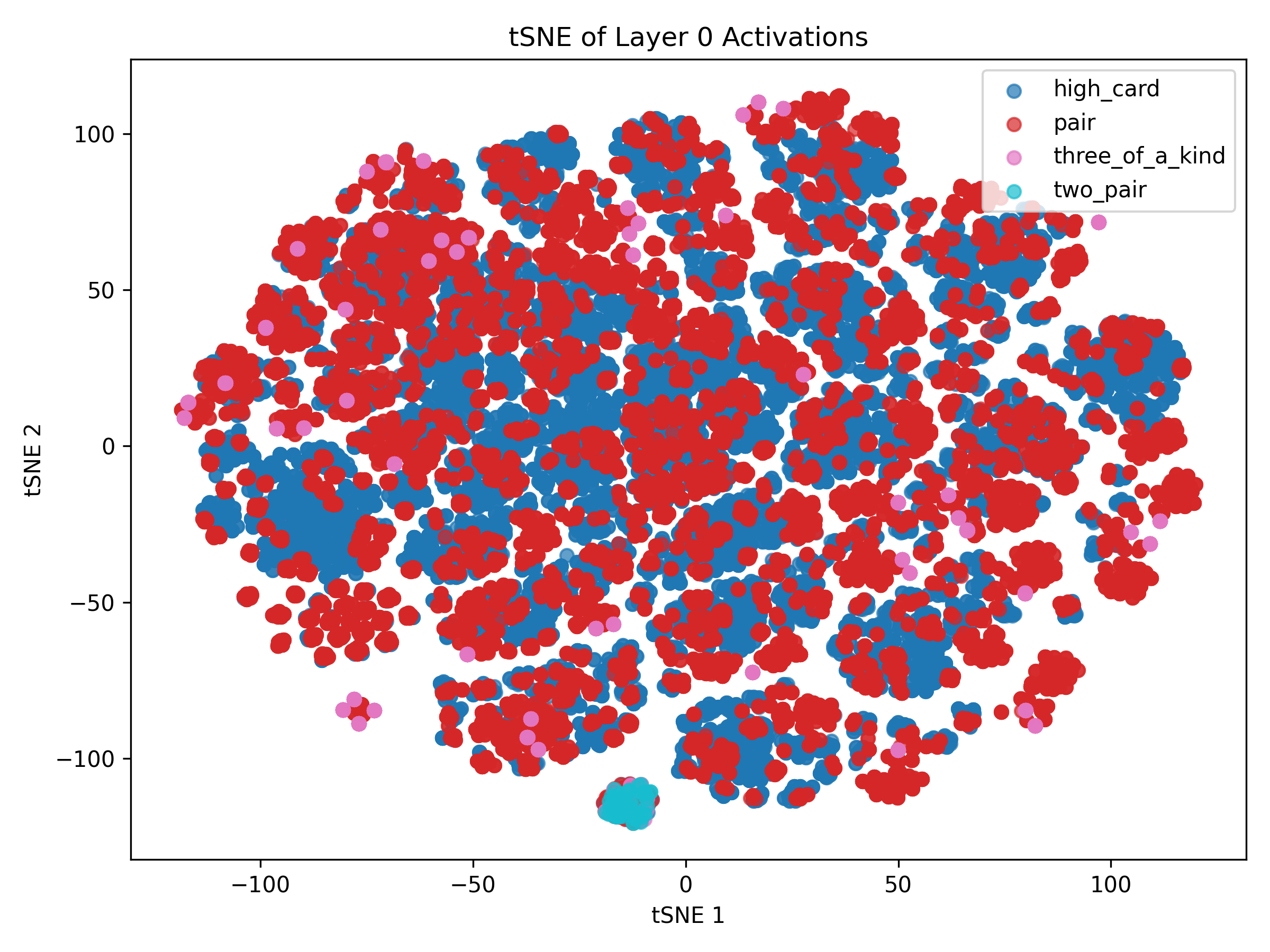}
        \caption{Layer 0}
    \end{subfigure}

    \par\medskip

    \begin{subfigure}[b]{0.24\textwidth}\centering
        \includegraphics[width=\textwidth]{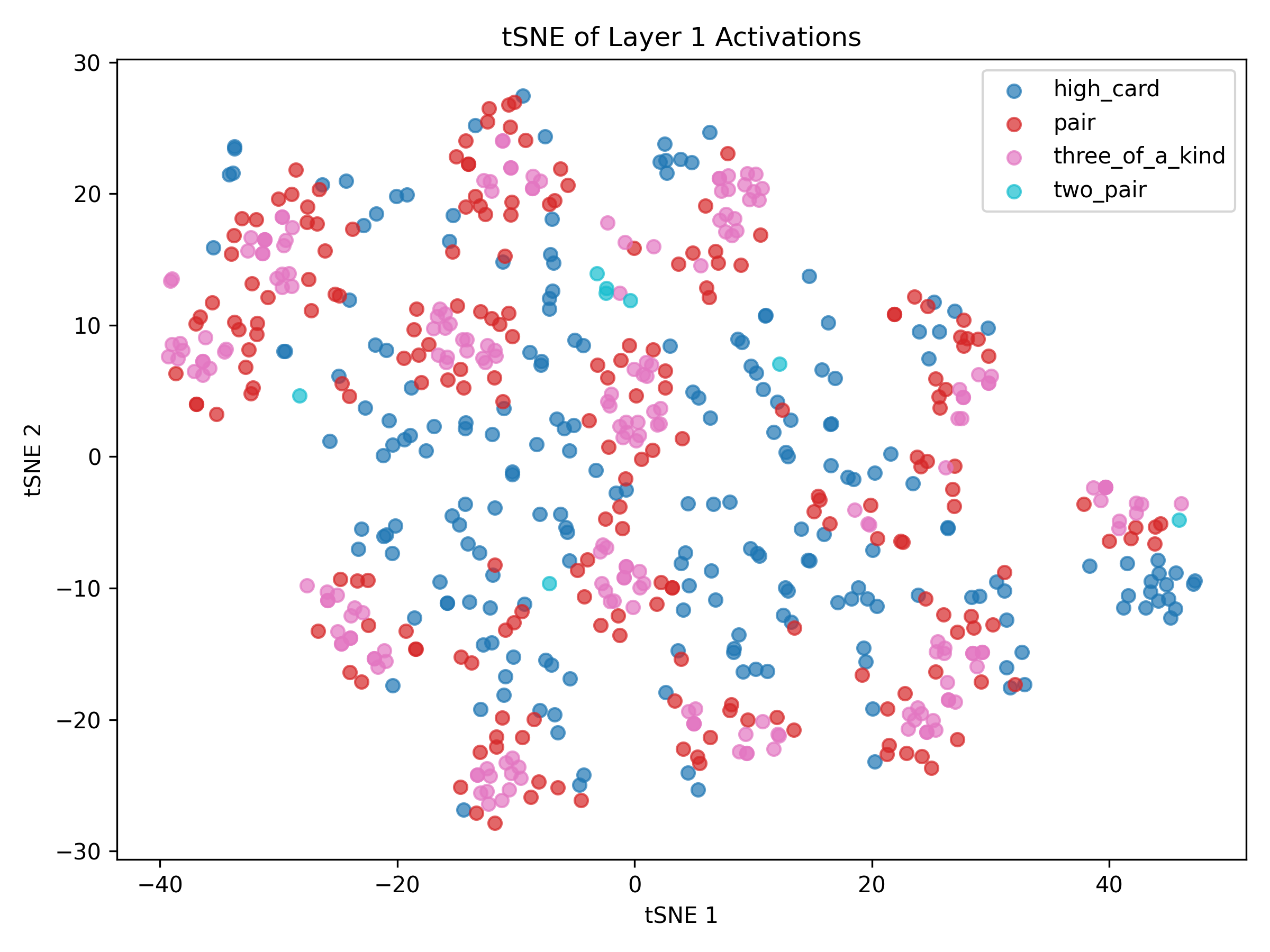}
        \caption{Layer 1}
    \end{subfigure}\hfill
    \begin{subfigure}[b]{0.24\textwidth}\centering
        \includegraphics[width=\textwidth]{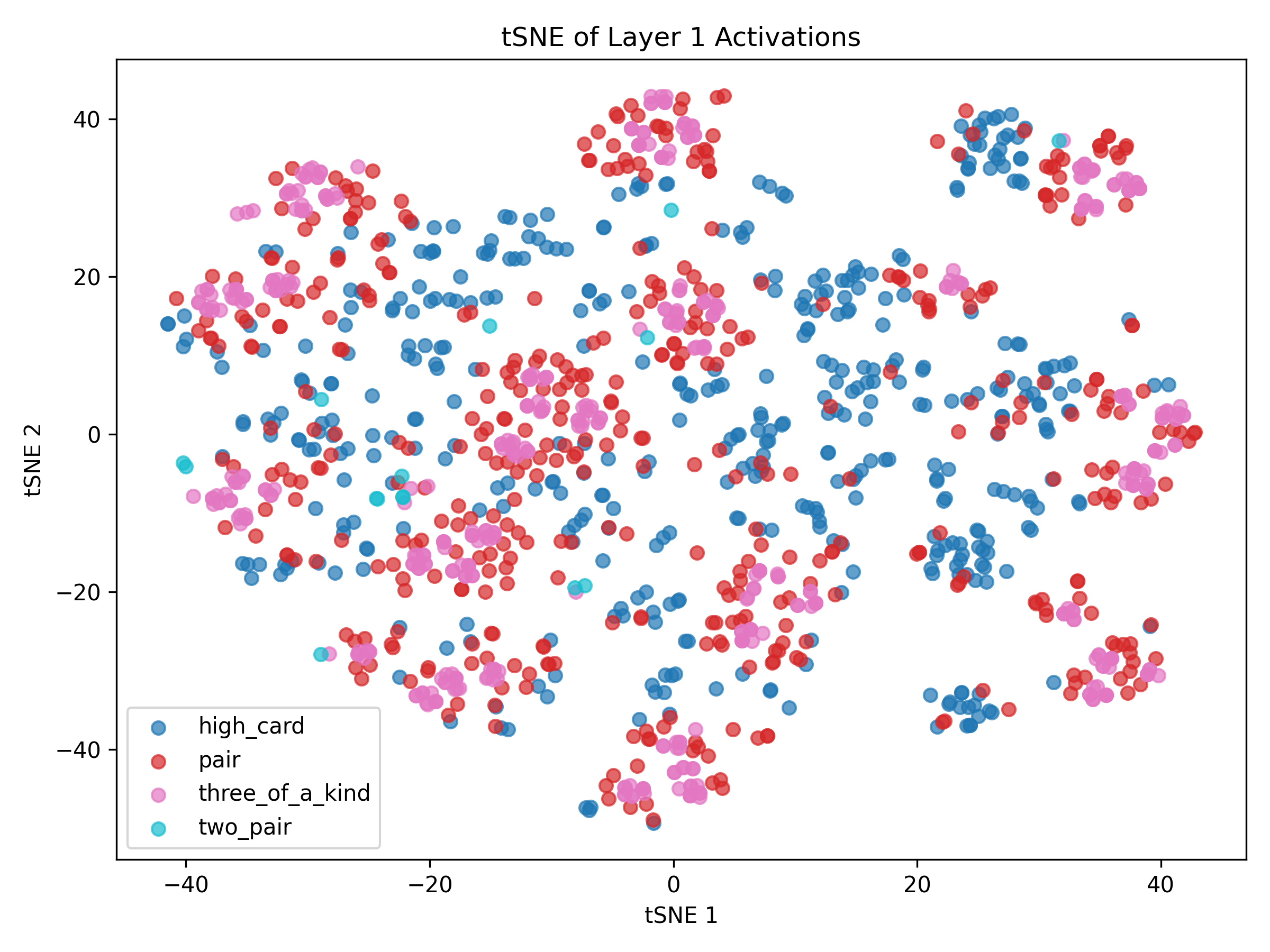}
        \caption{Layer 1}
    \end{subfigure}\hfill
    \begin{subfigure}[b]{0.24\textwidth}\centering
        \includegraphics[width=\textwidth]{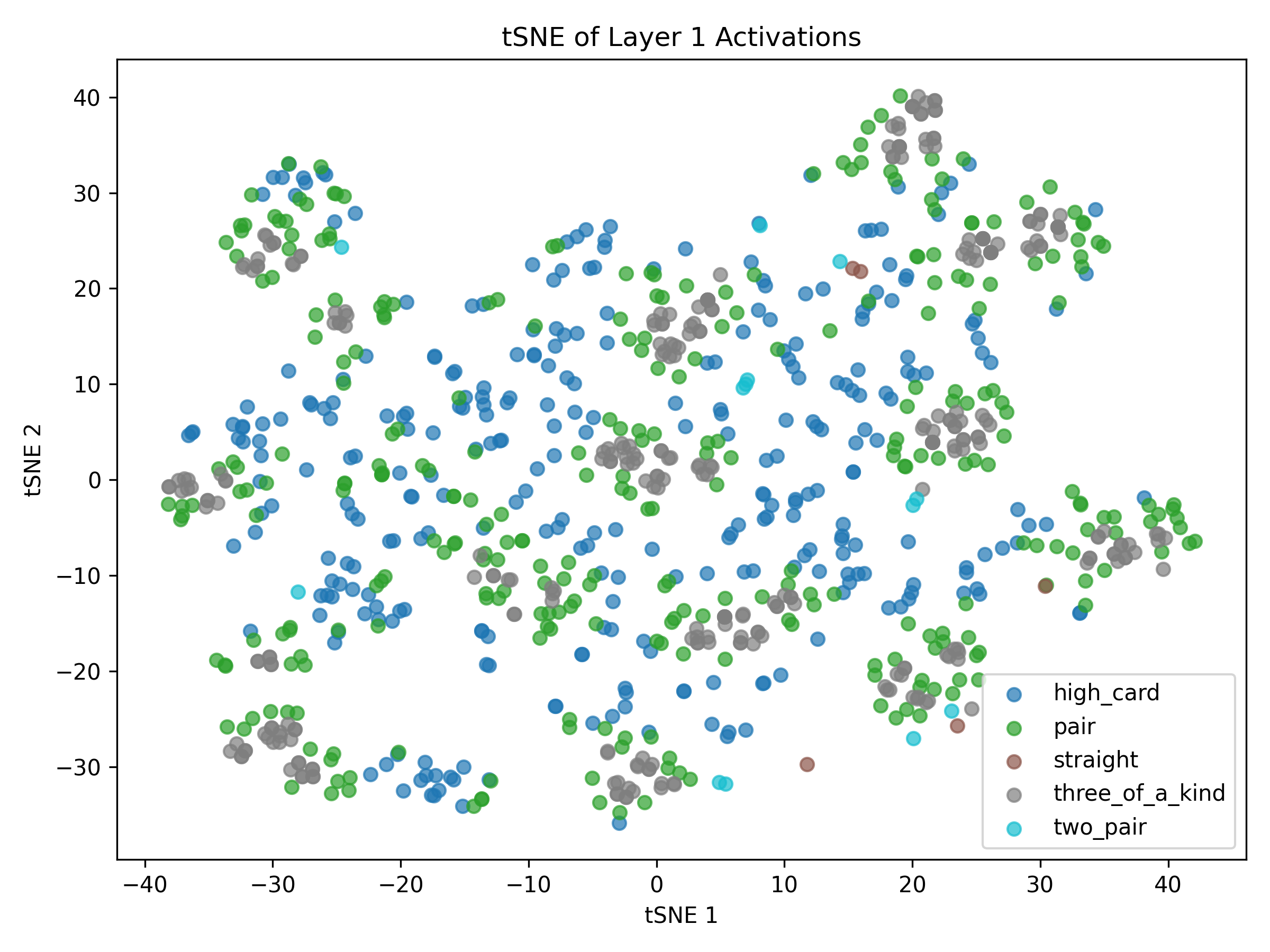}
        \caption{Layer 1}
    \end{subfigure}\hfill
    \begin{subfigure}[b]{0.24\textwidth}\centering
        \includegraphics[width=\textwidth]{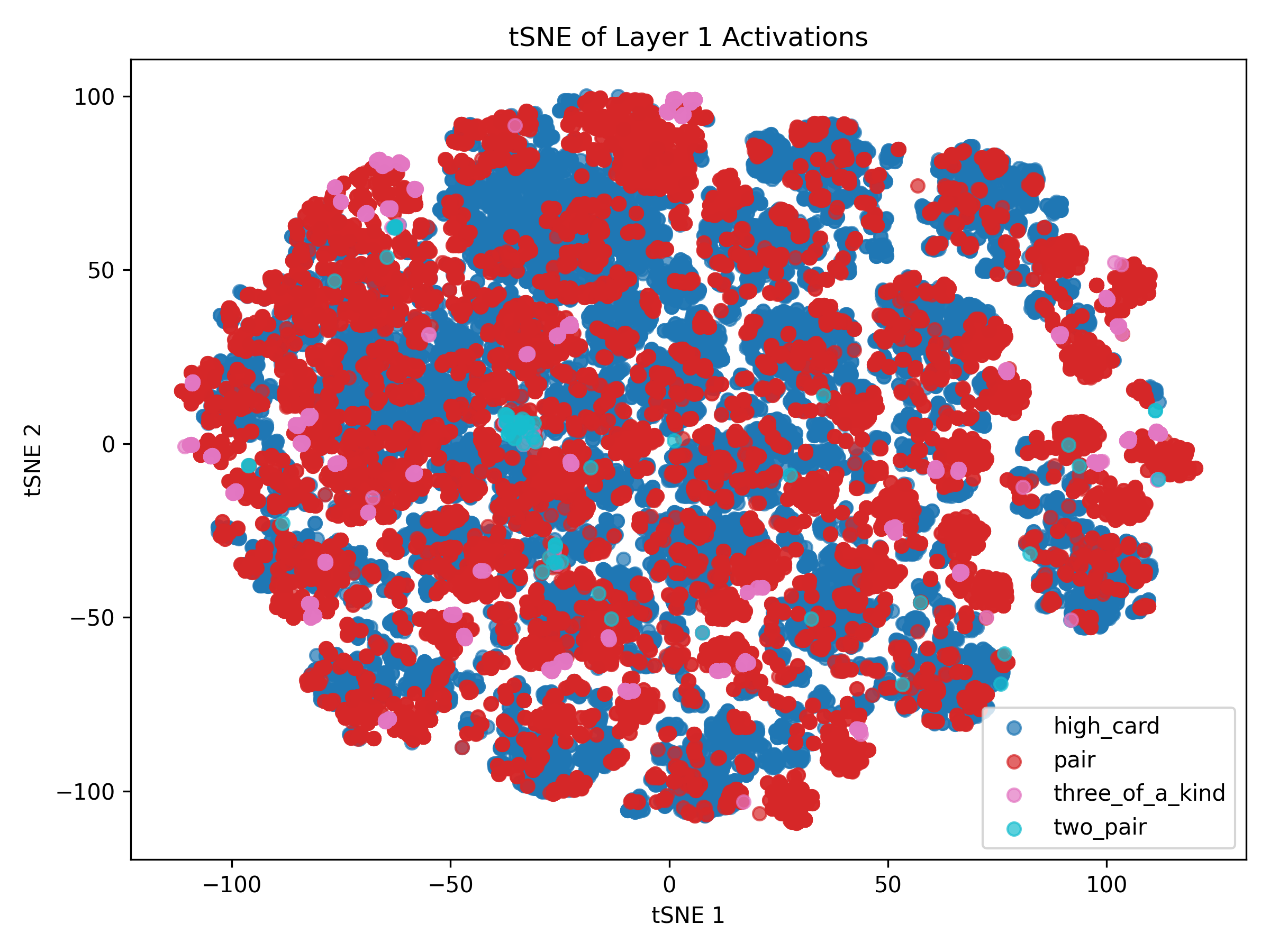}
        \caption{Layer 1}
    \end{subfigure}

    \par\medskip

    \begin{subfigure}[b]{0.24\textwidth}\centering
        \includegraphics[width=\textwidth]{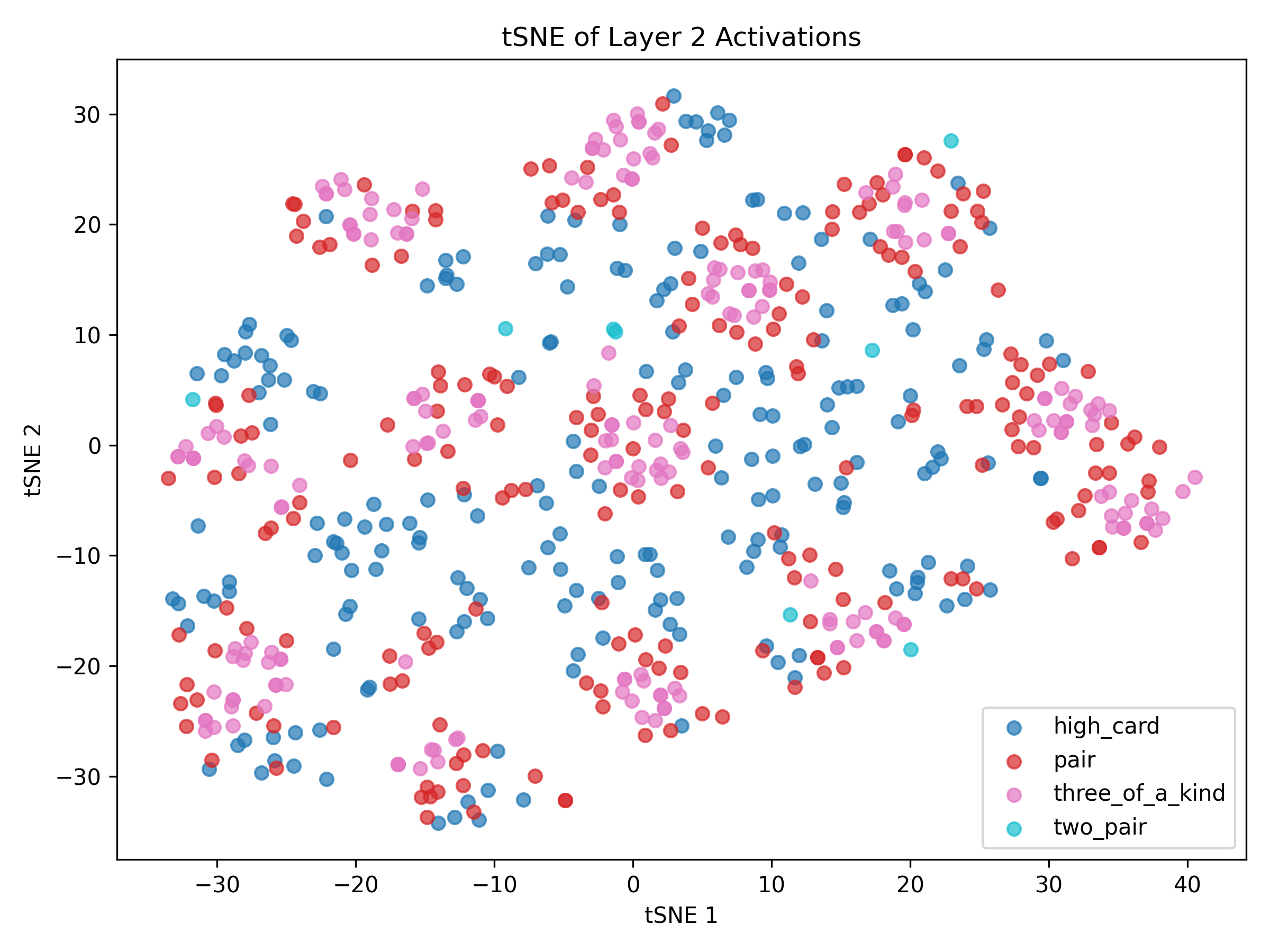}
        \caption{Layer 2}
    \end{subfigure}\hfill
    \begin{subfigure}[b]{0.24\textwidth}\centering
        \includegraphics[width=\textwidth]{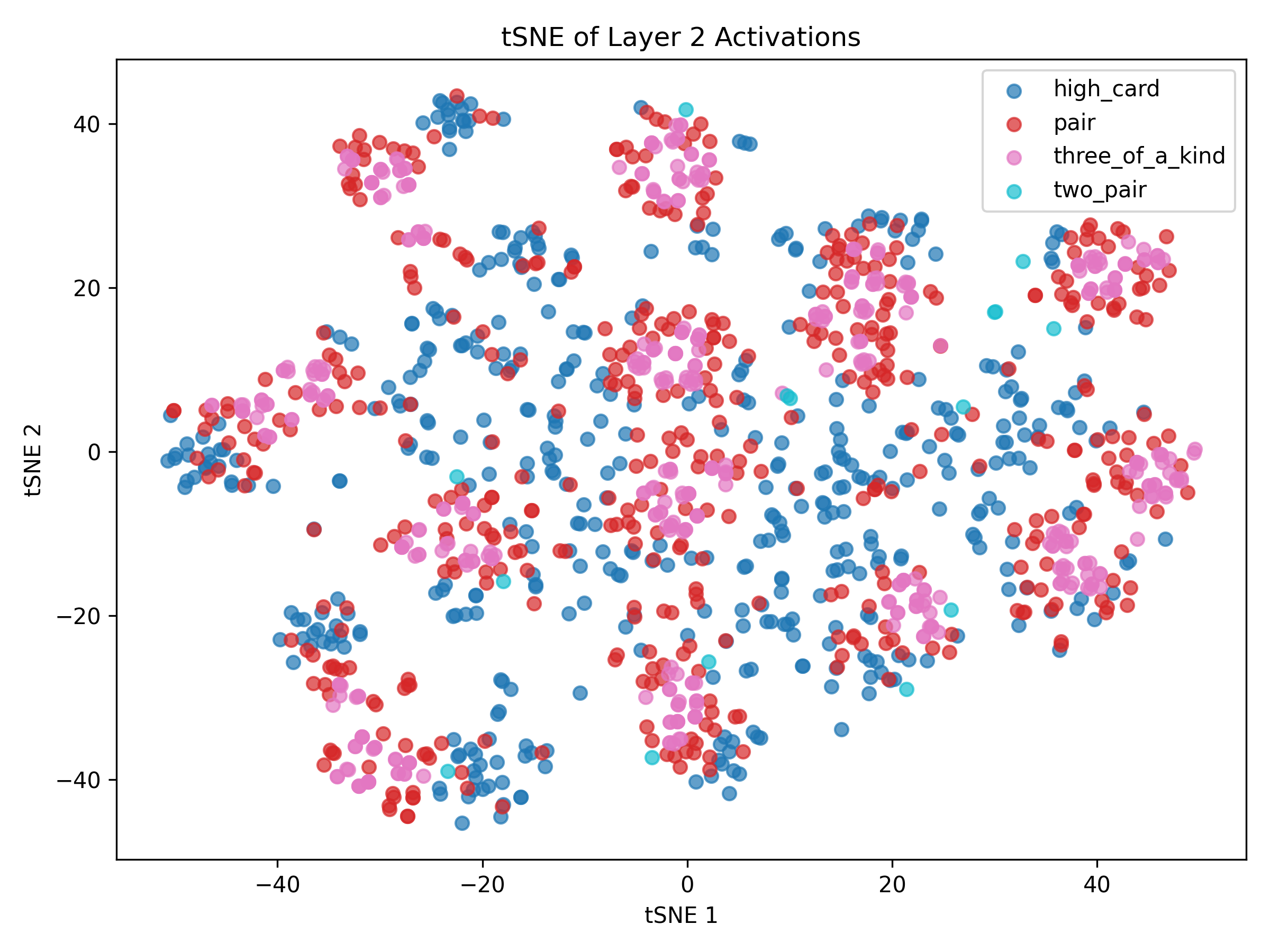}
        \caption{Layer 2}
    \end{subfigure}\hfill
    \begin{subfigure}[b]{0.24\textwidth}\centering
        \includegraphics[width=\textwidth]{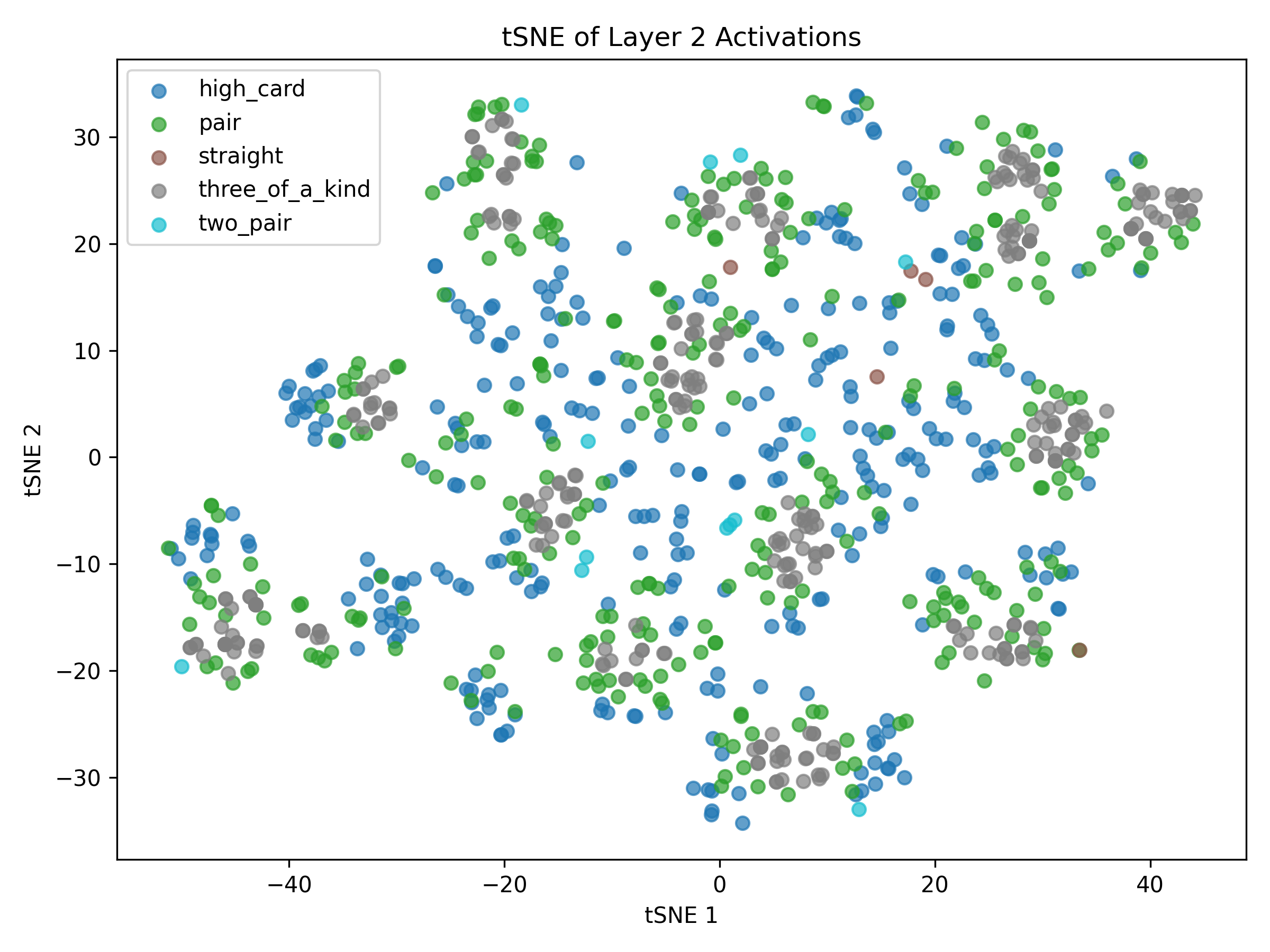}
        \caption{Layer 2}
    \end{subfigure}\hfill
    \begin{subfigure}[b]{0.24\textwidth}\centering
        \includegraphics[width=\textwidth]{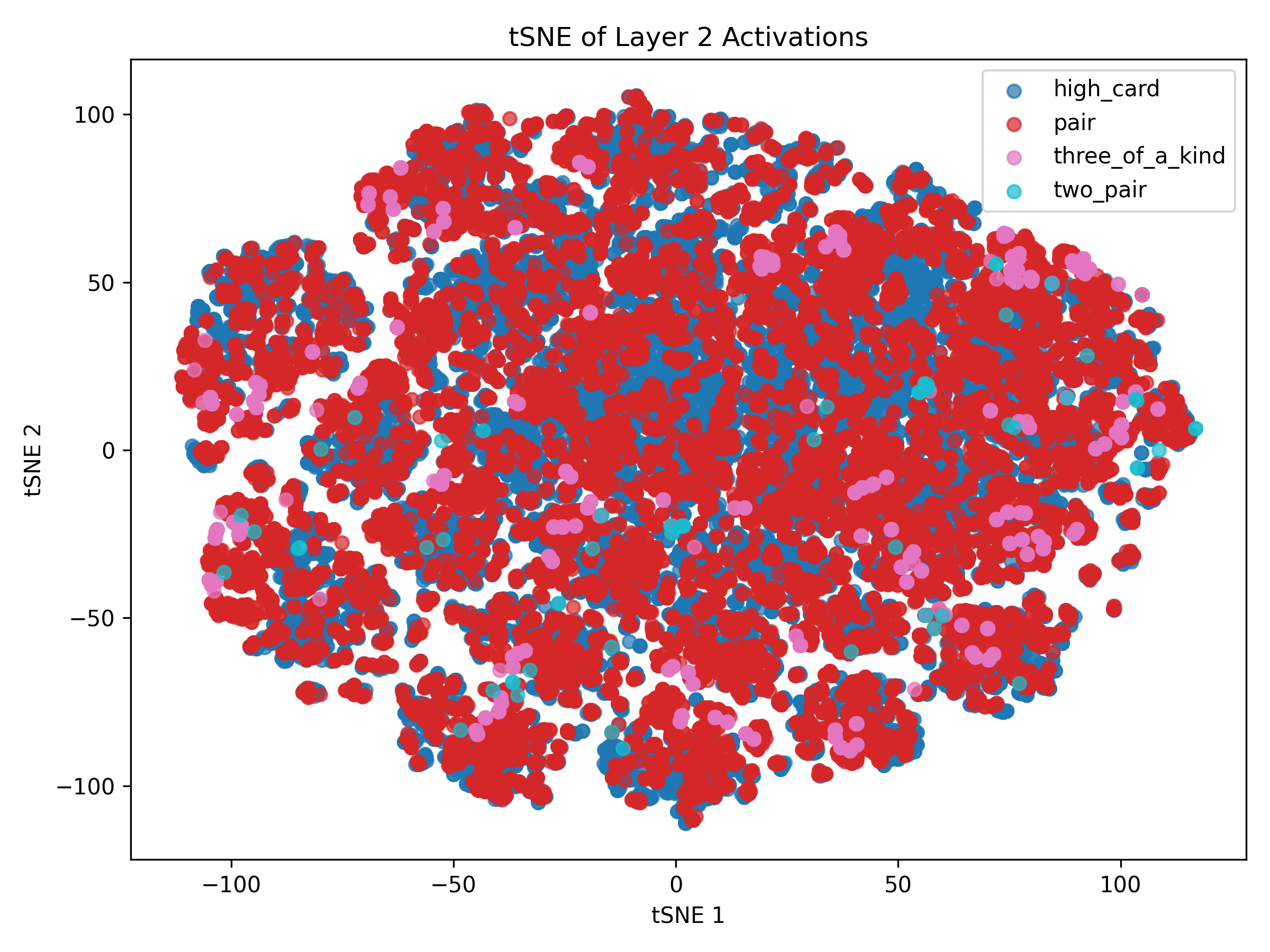}
        \caption{Layer 2}
    \end{subfigure}

    \par\medskip

    \begin{subfigure}[b]{0.24\textwidth}\centering
        \includegraphics[width=\textwidth]{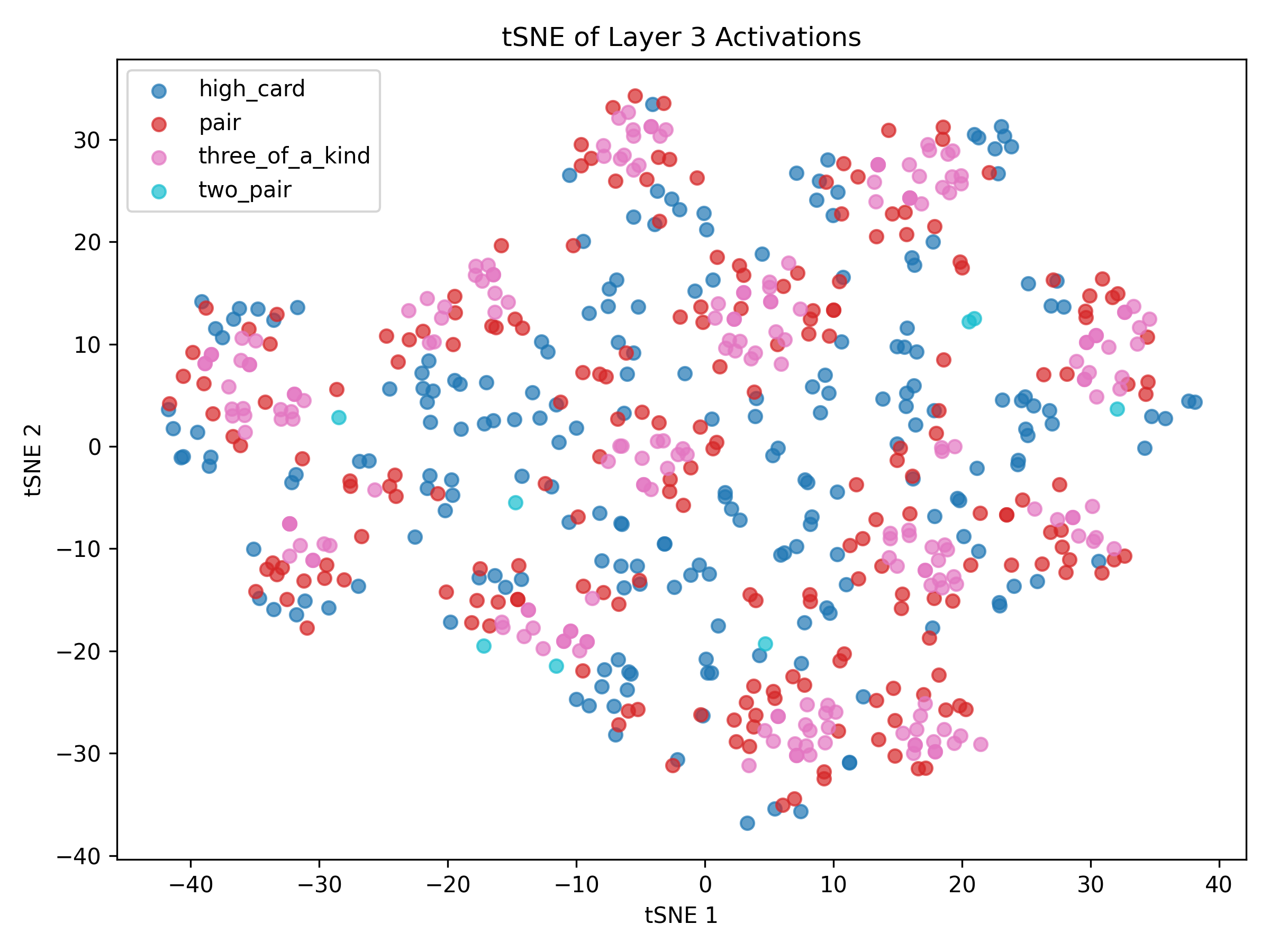}
        \caption{Layer 3}
    \end{subfigure}\hfill
    \begin{subfigure}[b]{0.24\textwidth}\centering
        \includegraphics[width=\textwidth]{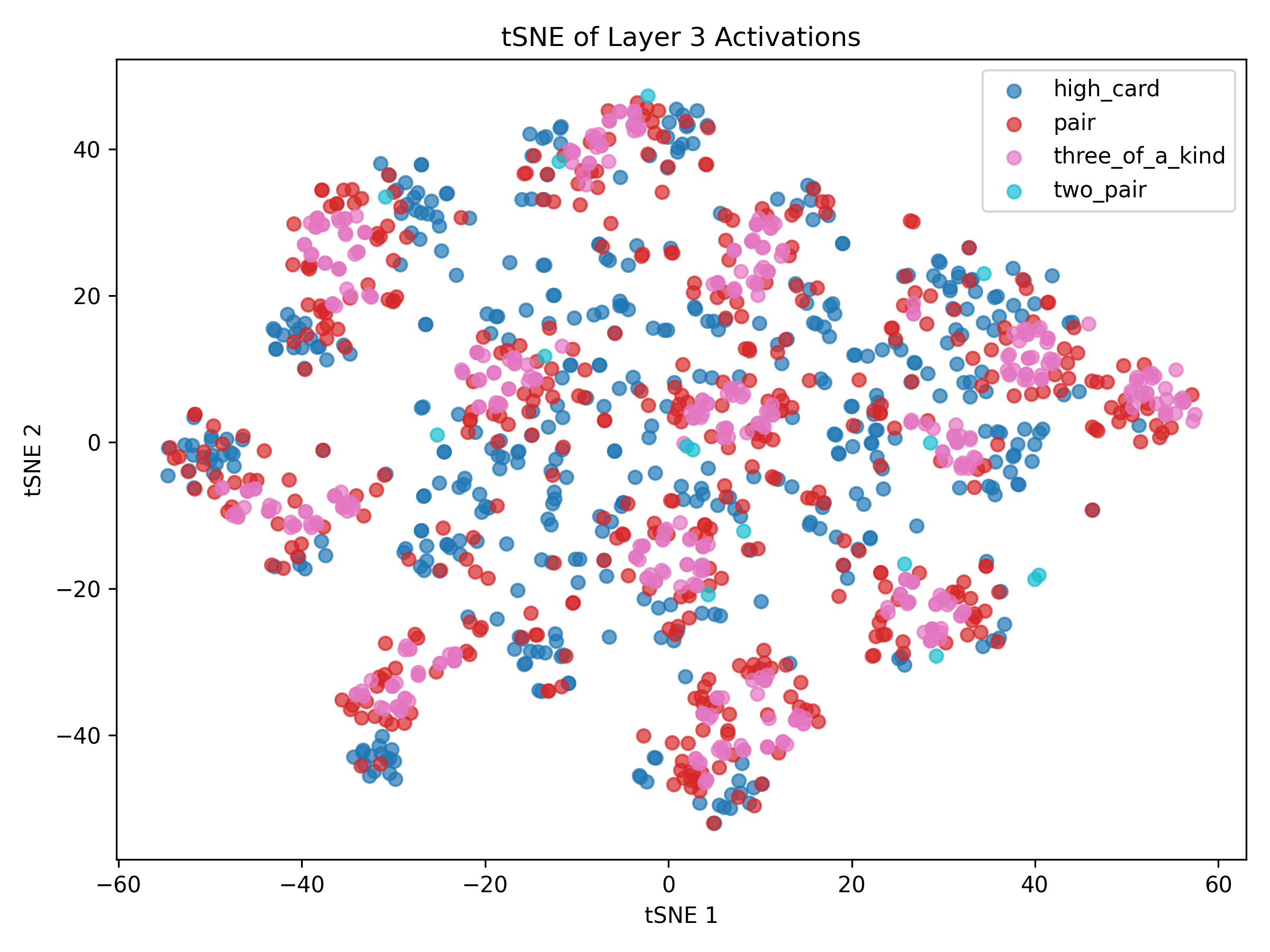}
        \caption{Layer 3}
    \end{subfigure}\hfill
    \begin{subfigure}[b]{0.24\textwidth}\centering
        \includegraphics[width=\textwidth]{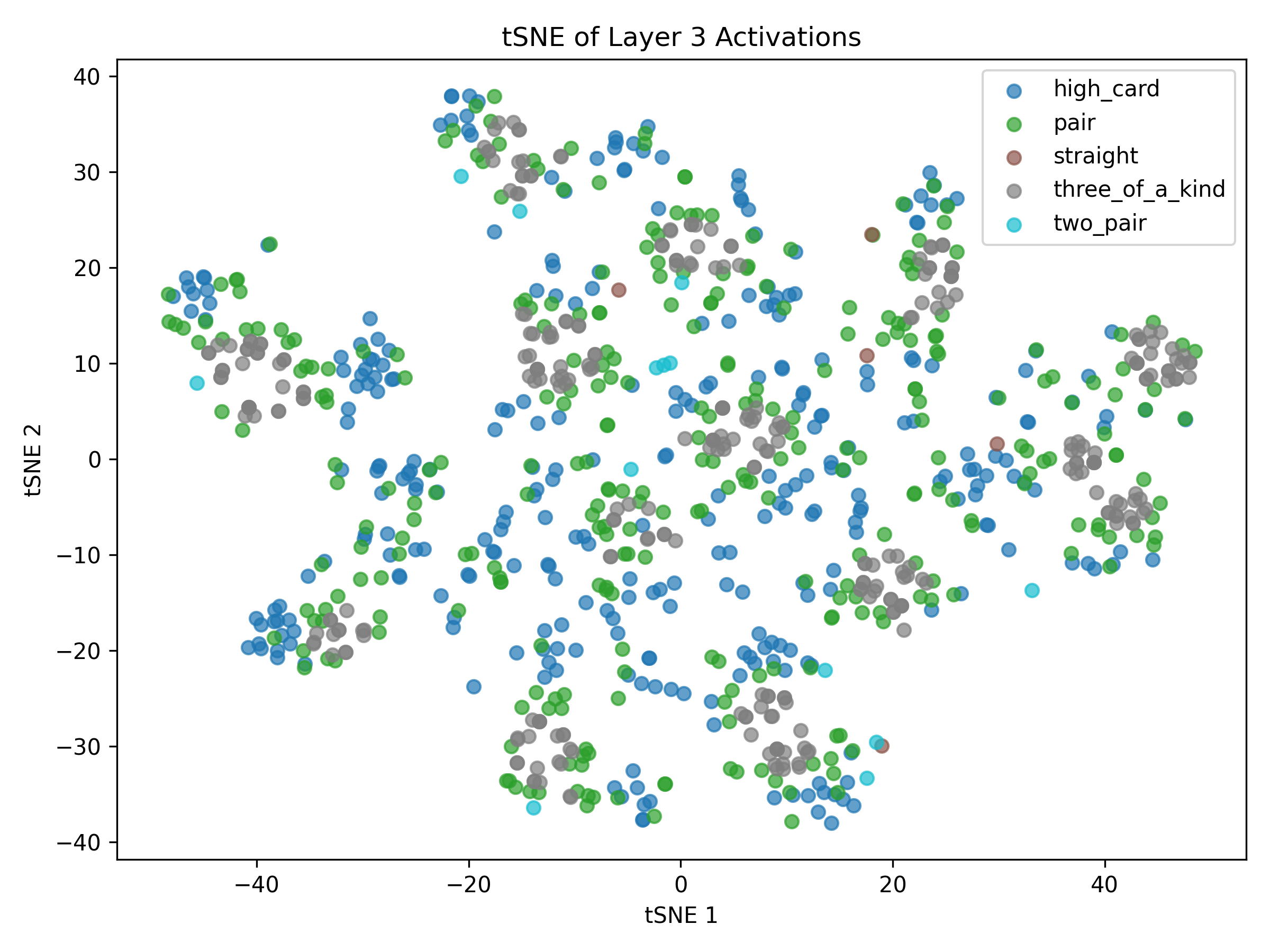}
        \caption{Layer 3}
    \end{subfigure}\hfill
    \begin{subfigure}[b]{0.24\textwidth}\centering
        \includegraphics[width=\textwidth]{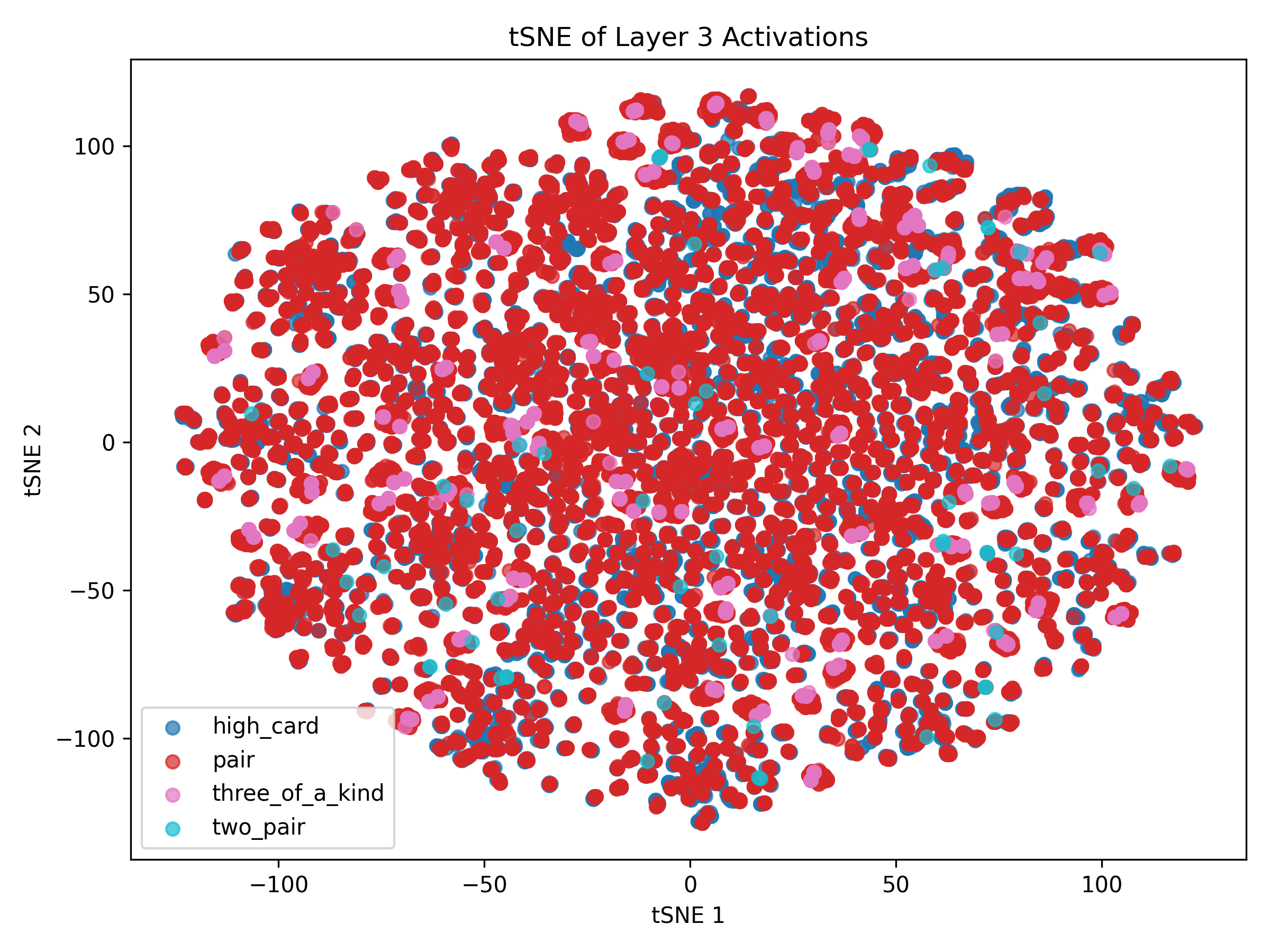}
        \caption{Layer 3}
    \end{subfigure}
    \caption{t-SNE visualizations of activation vectors across transformer Layers 0–3 and
    four different training set sizes. Each subplot embeds per-token activations into two dimensions, colored by hand-rank class. t-SNE reveals fine-grained cluster structure that is
    especially pronounced for conceptually similar hands (e.g., pairs and
    three-of-a-kind). Deeper layers also exhibit tighter, more separated clusters,
    indicating progressive specialization of internal representations.}
\end{figure}
Each subplot below shows a UMAP
two-dimensional embedding of activation vectors colored by hand-rank class. UMAP reveals partial clustering behavior, though with less separation and interpretability than PCA or t-SNE.
\medskip

\begin{figure}[H]
    \centering
    
    \label{fig:umap_grid}

    \begin{subfigure}[b]{0.24\textwidth}\centering
        \includegraphics[width=\textwidth]{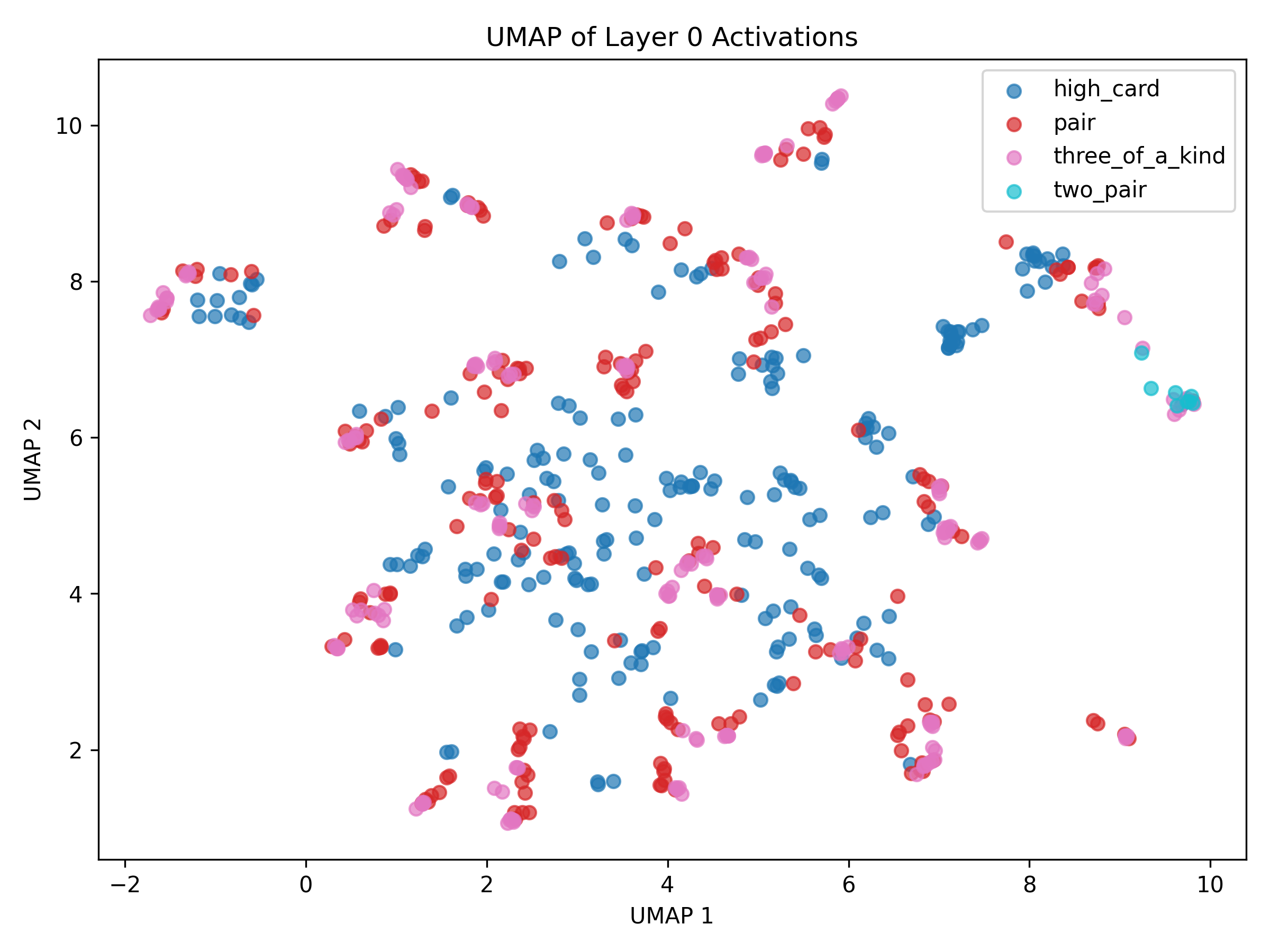}
        \caption{Layer 0}
    \end{subfigure}\hfill
    \begin{subfigure}[b]{0.24\textwidth}\centering
        \includegraphics[width=\textwidth]{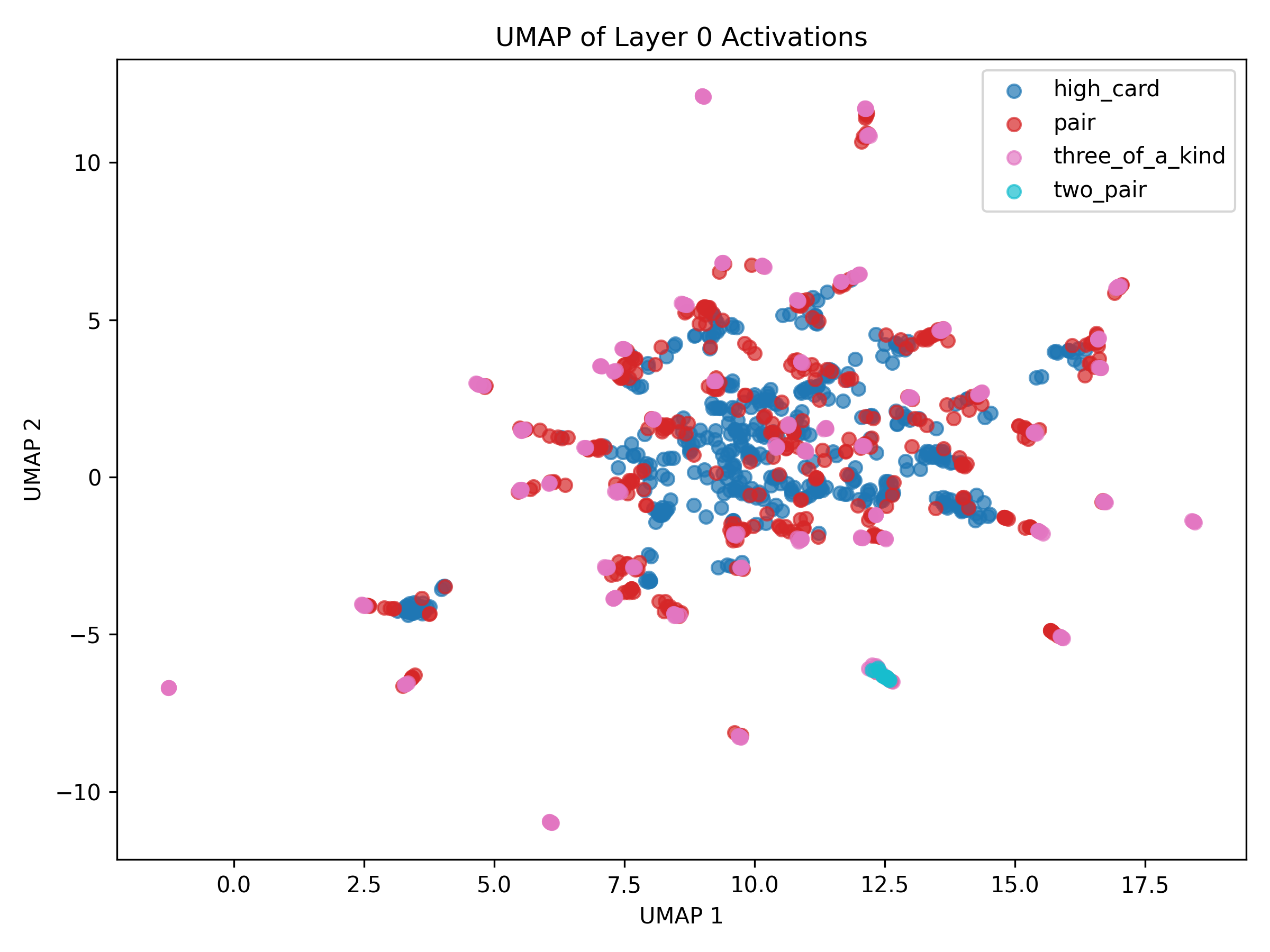}
        \caption{Layer 0}
    \end{subfigure}\hfill
    \begin{subfigure}[b]{0.24\textwidth}\centering
        \includegraphics[width=\textwidth]{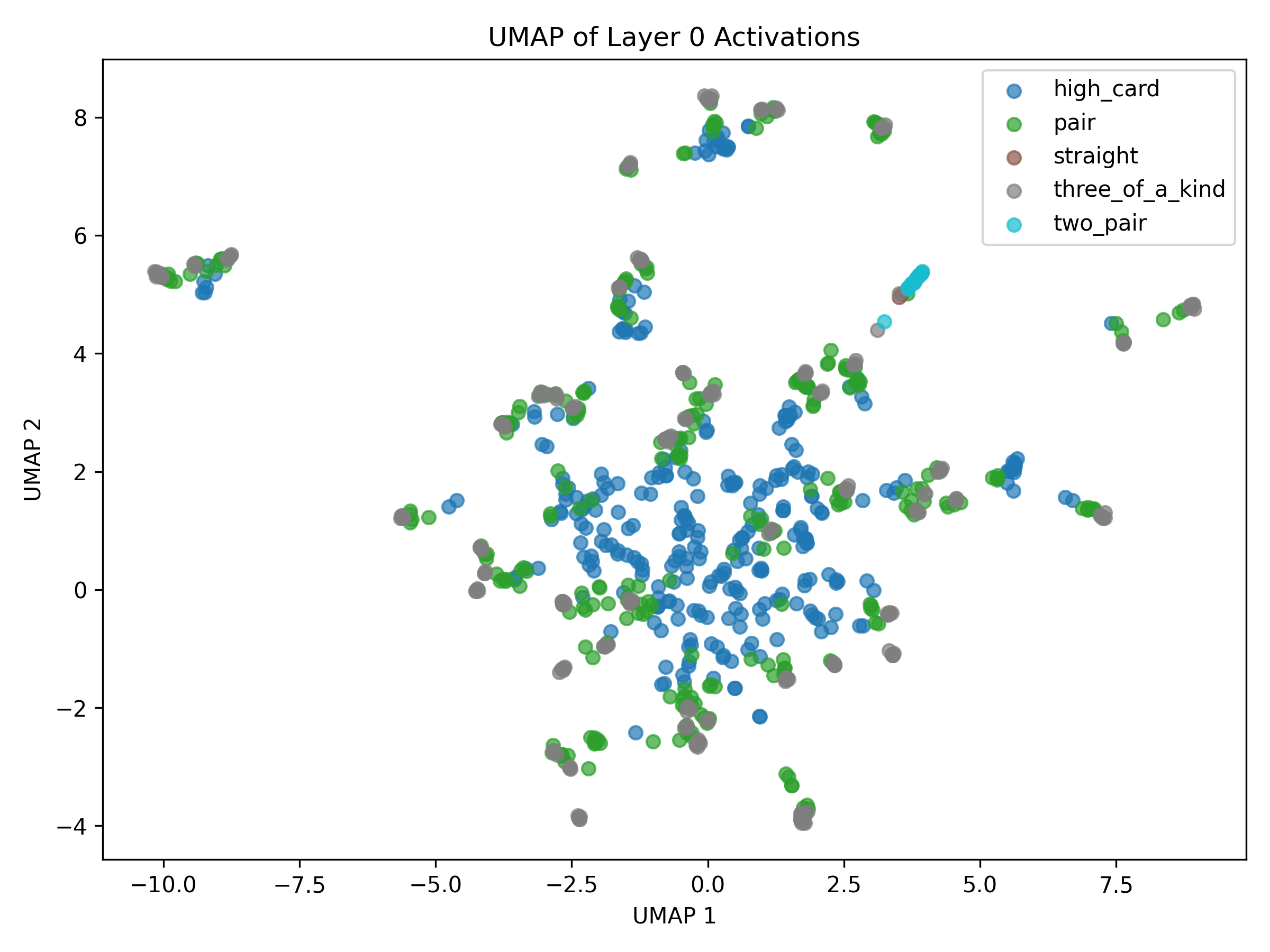}
        \caption{Layer 0}
    \end{subfigure}\hfill
    \begin{subfigure}[b]{0.24\textwidth}\centering
        \includegraphics[width=\textwidth]{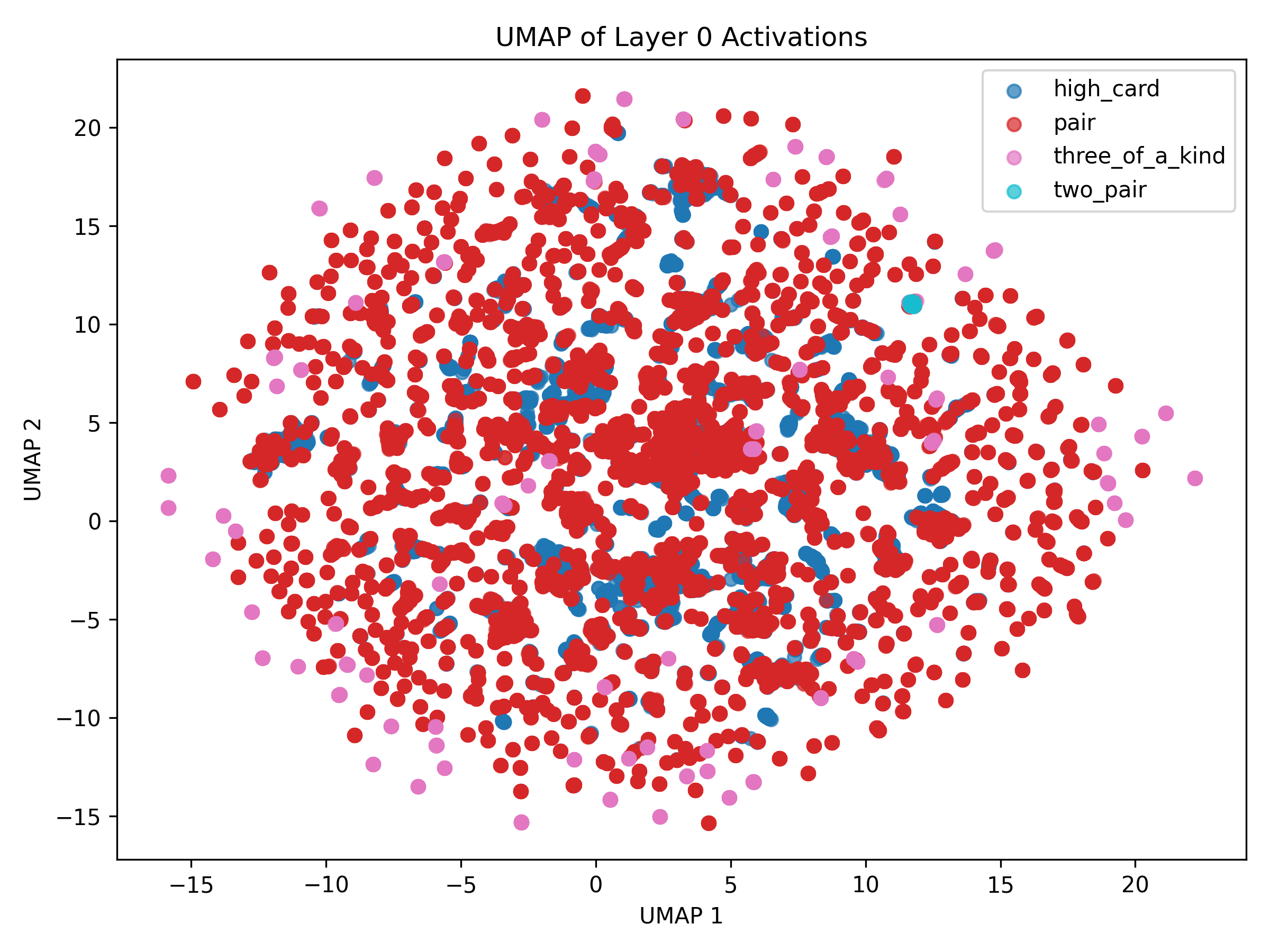}
        \caption{Layer 0}
    \end{subfigure}

    \par\medskip

    \begin{subfigure}[b]{0.24\textwidth}\centering
        \includegraphics[width=\textwidth]{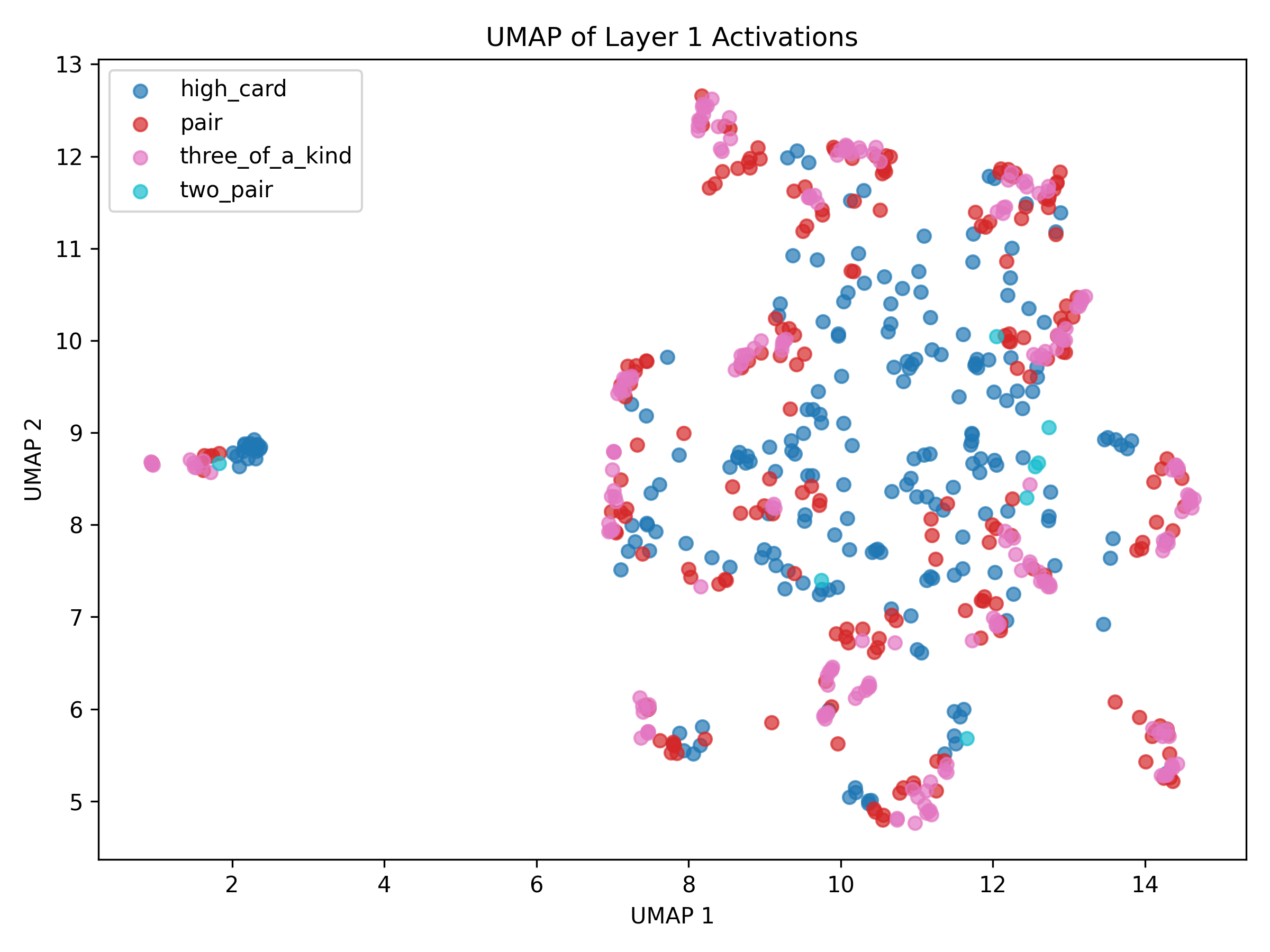}
        \caption{Layer 1}
    \end{subfigure}\hfill
    \begin{subfigure}[b]{0.24\textwidth}\centering
        \includegraphics[width=\textwidth]{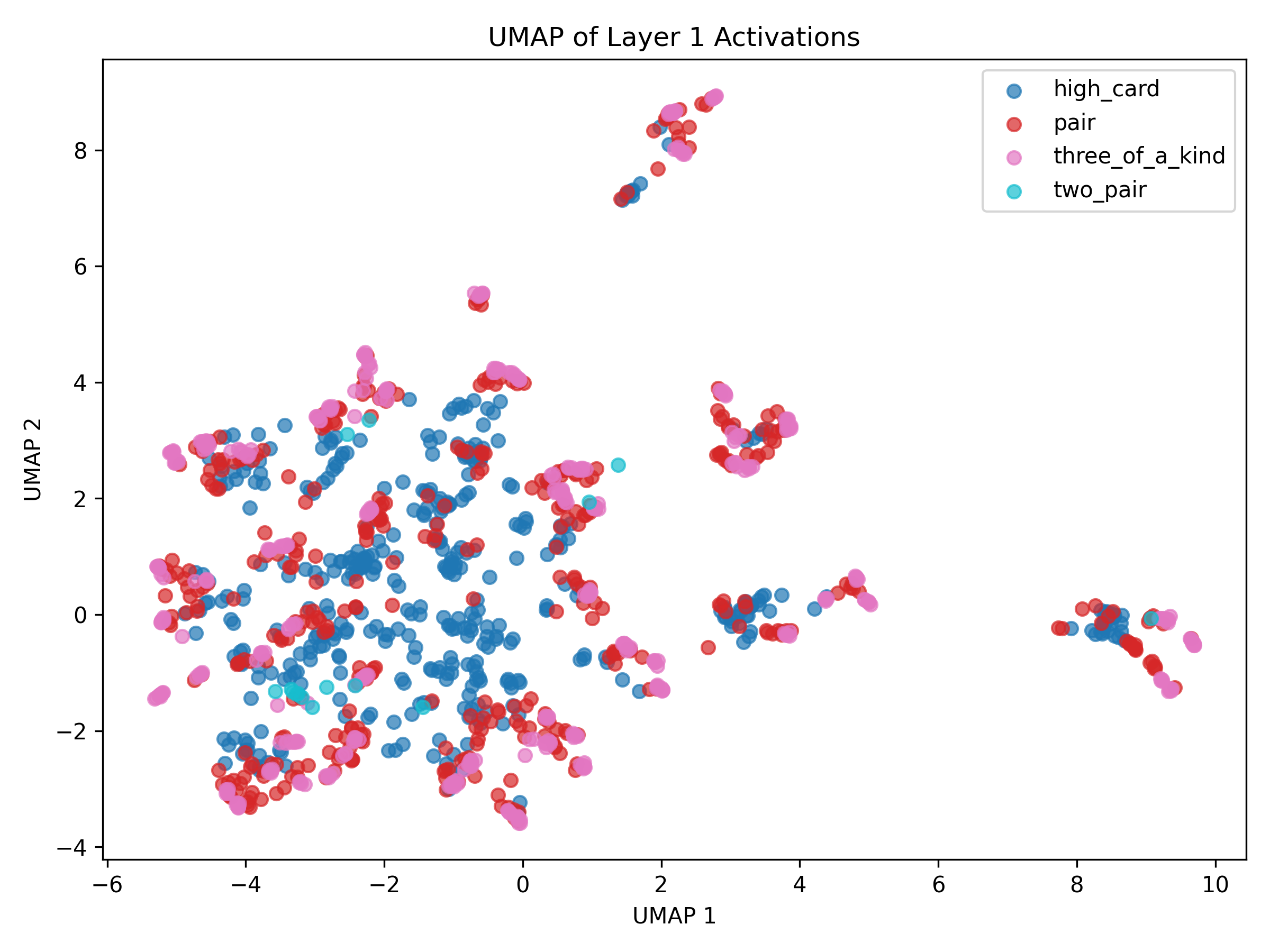}
        \caption{Layer 1}
    \end{subfigure}\hfill
    \begin{subfigure}[b]{0.24\textwidth}\centering
        \includegraphics[width=\textwidth]{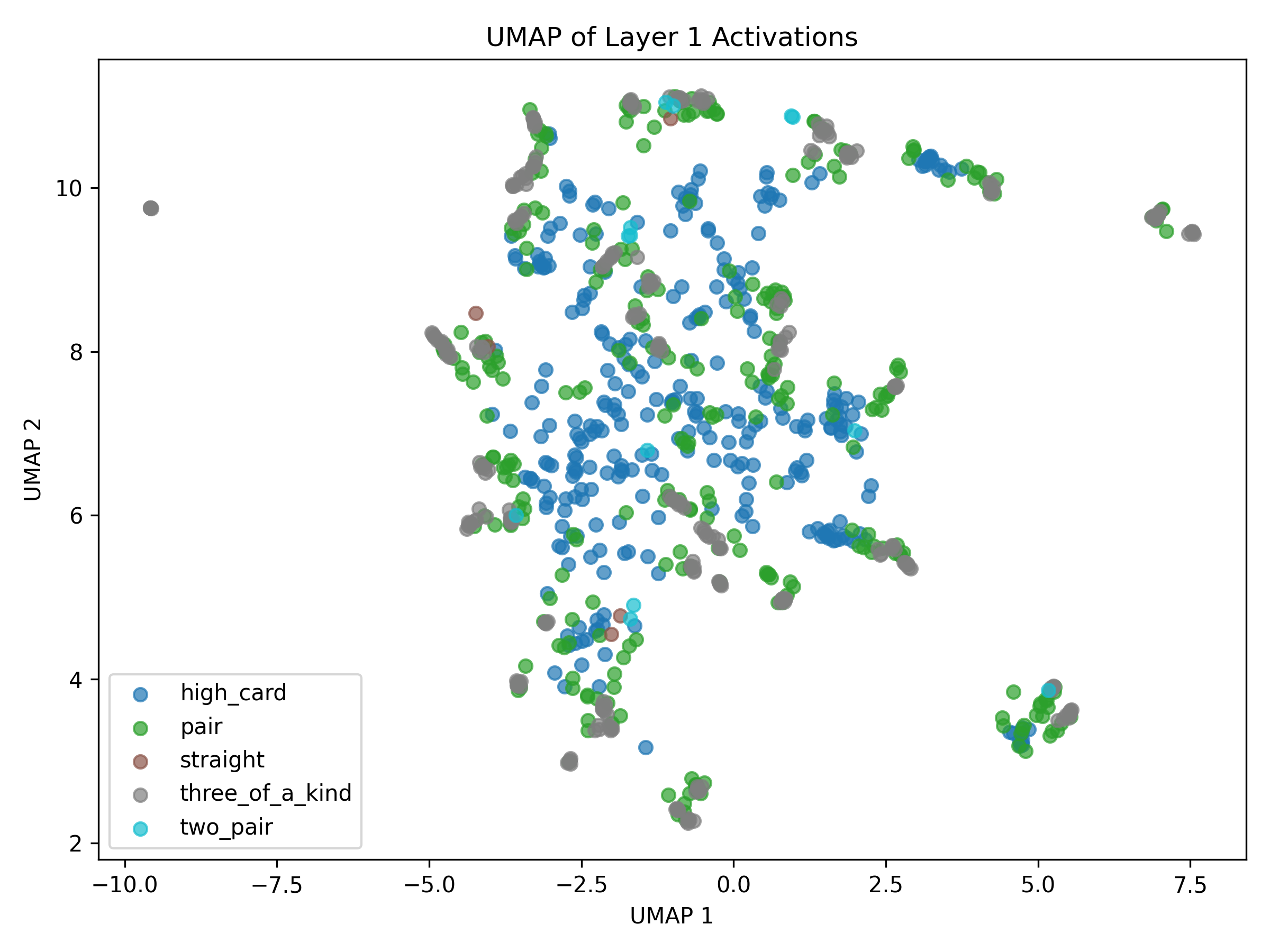}
        \caption{Layer 1}
    \end{subfigure}\hfill
    \begin{subfigure}[b]{0.24\textwidth}\centering
        \includegraphics[width=\textwidth]{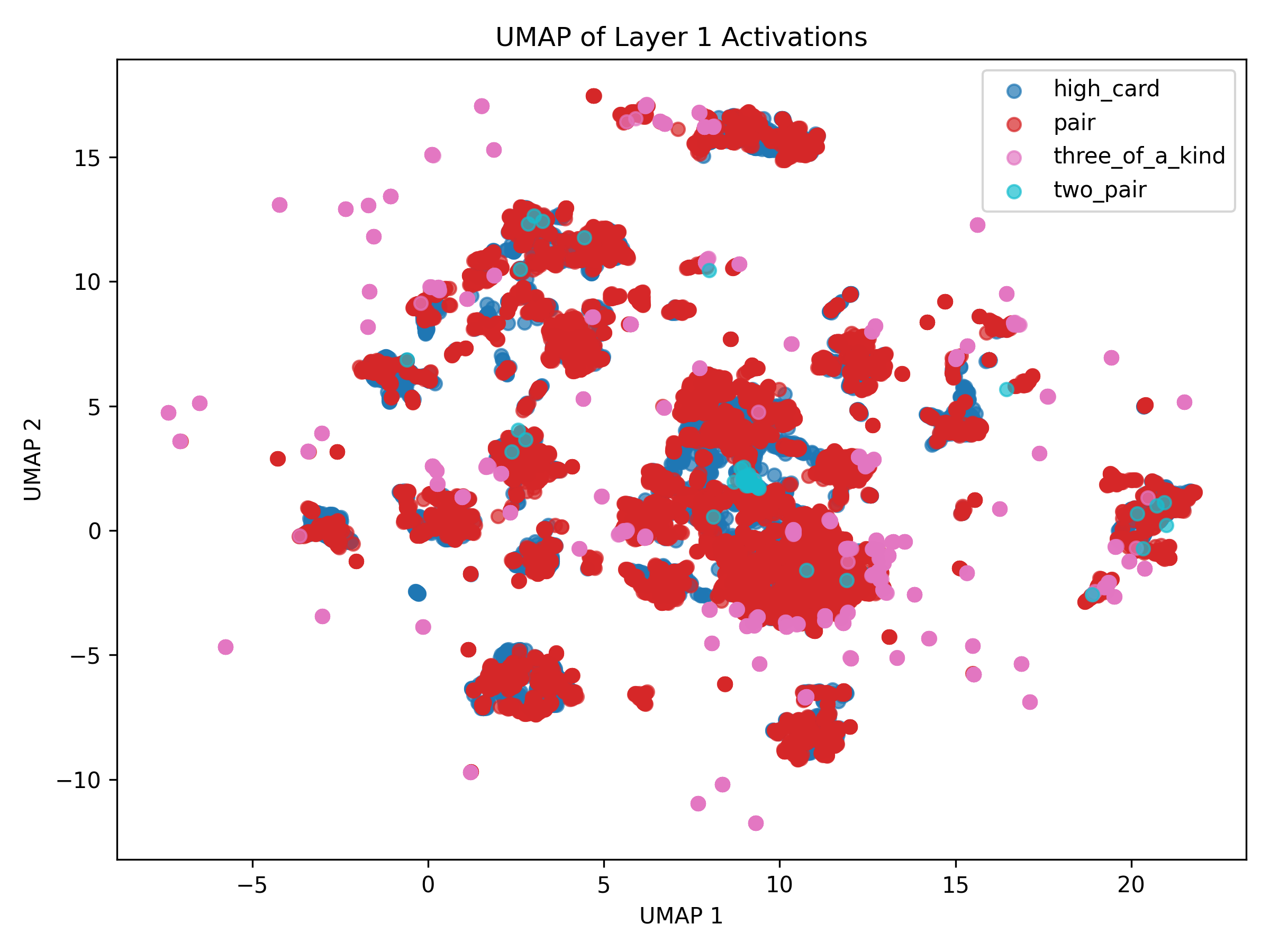}
        \caption{Layer 1}
    \end{subfigure}

    \par\medskip

    \begin{subfigure}[b]{0.24\textwidth}\centering
        \includegraphics[width=\textwidth]{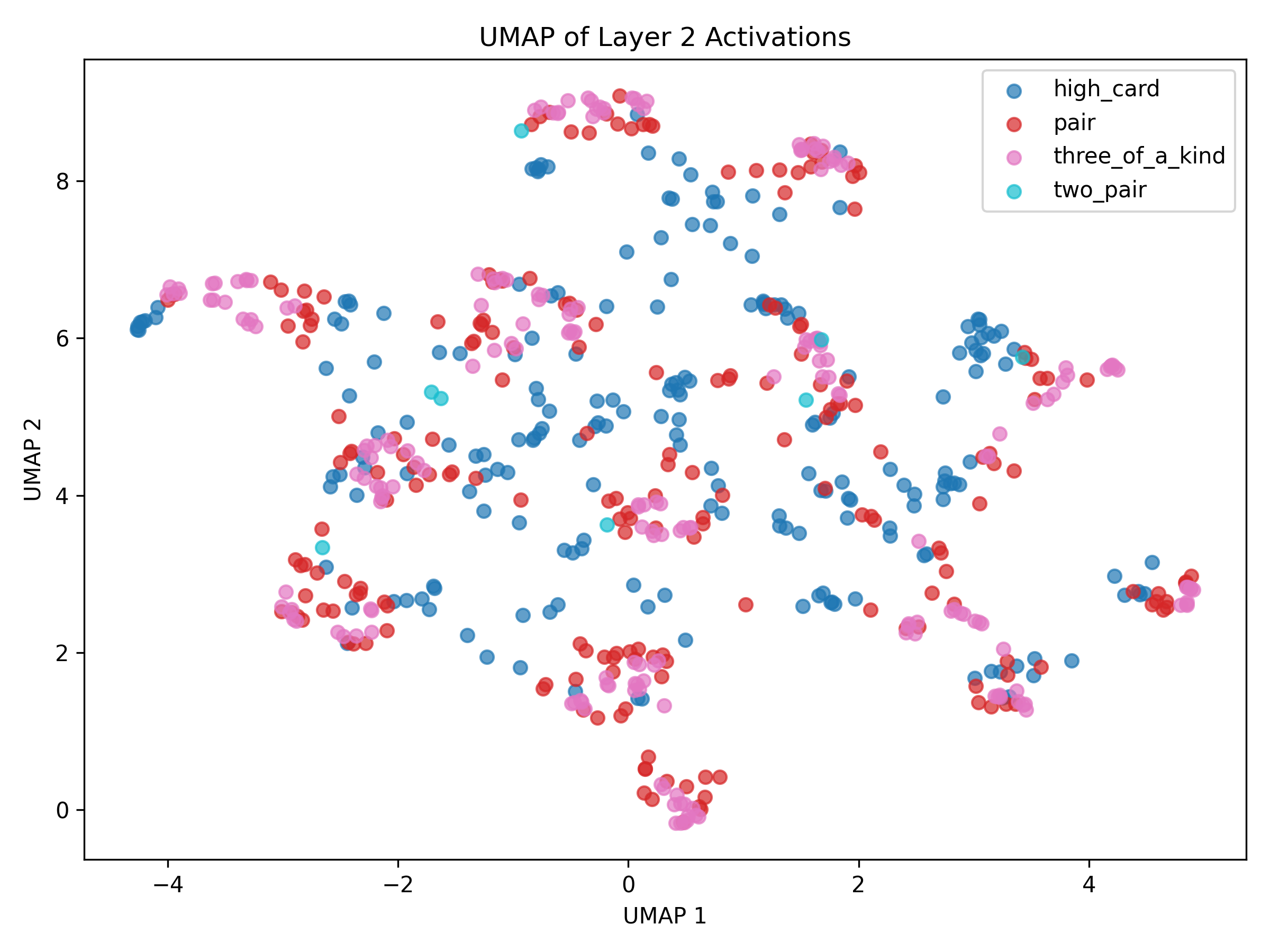}
        \caption{Layer 2}
    \end{subfigure}\hfill
    \begin{subfigure}[b]{0.24\textwidth}\centering
        \includegraphics[width=\textwidth]{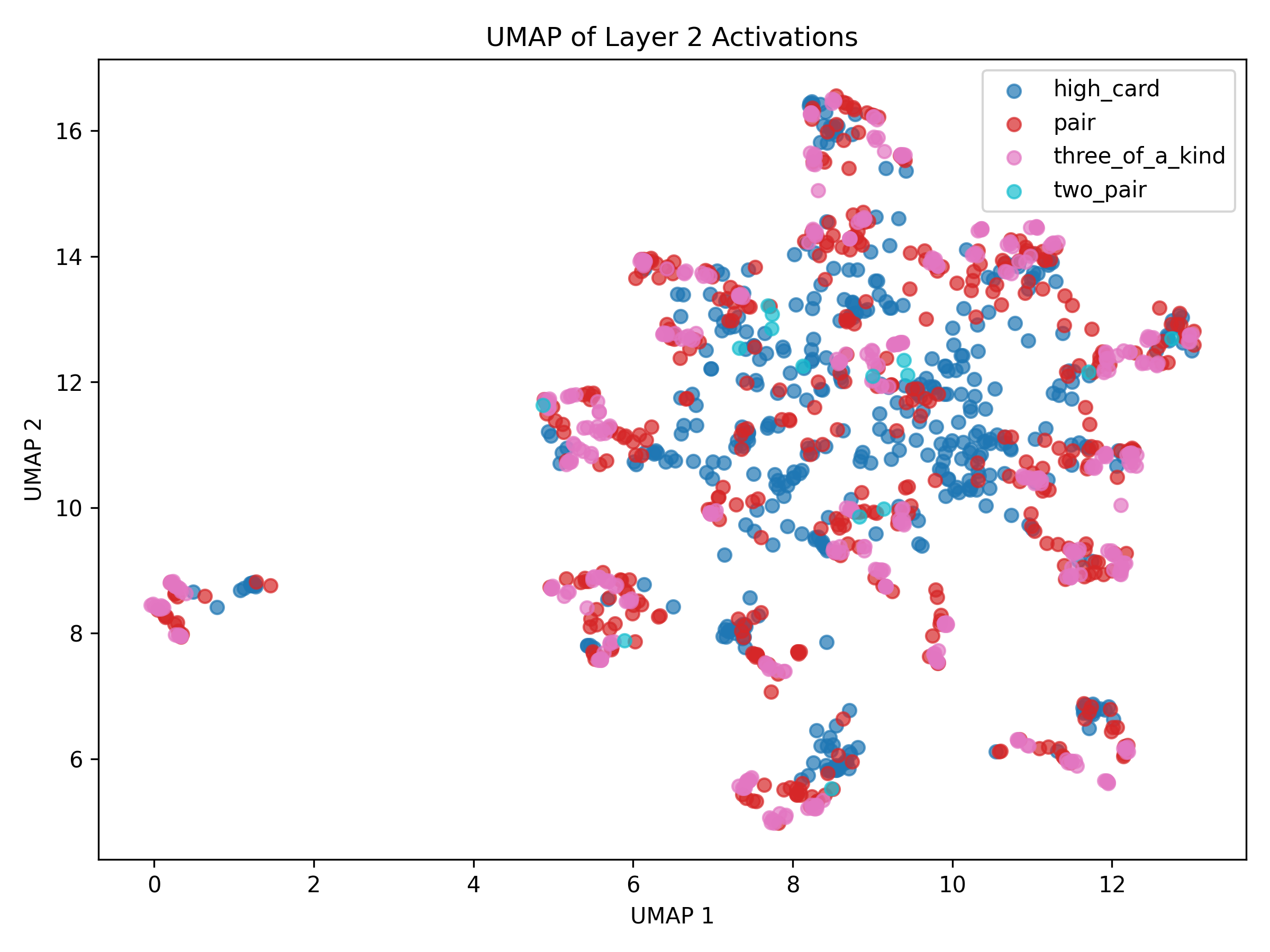}
        \caption{Layer 2}
    \end{subfigure}\hfill
    \begin{subfigure}[b]{0.24\textwidth}\centering
        \includegraphics[width=\textwidth]{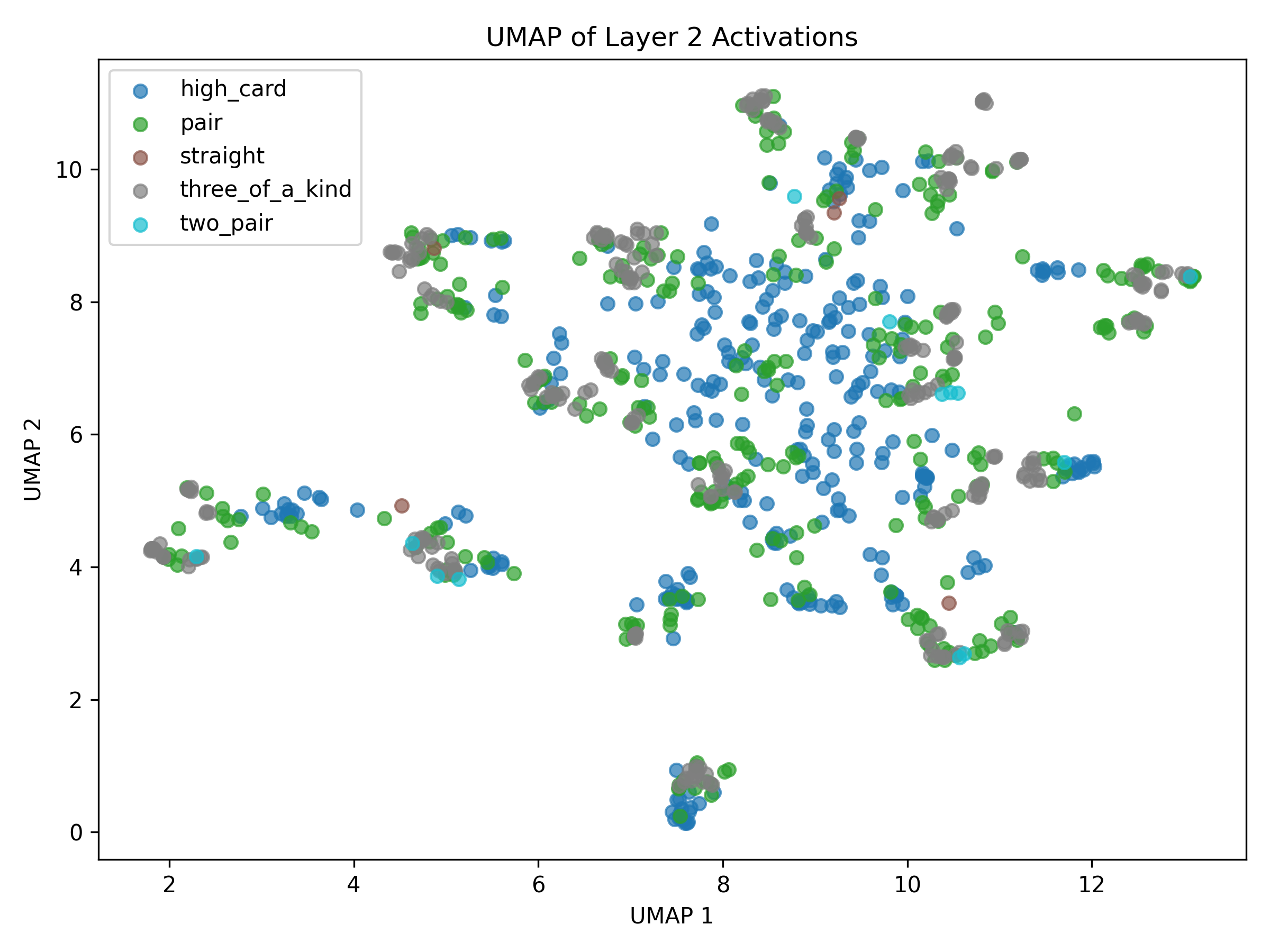}
        \caption{Layer 2}
    \end{subfigure}\hfill
    \begin{subfigure}[b]{0.24\textwidth}\centering
        \includegraphics[width=\textwidth]{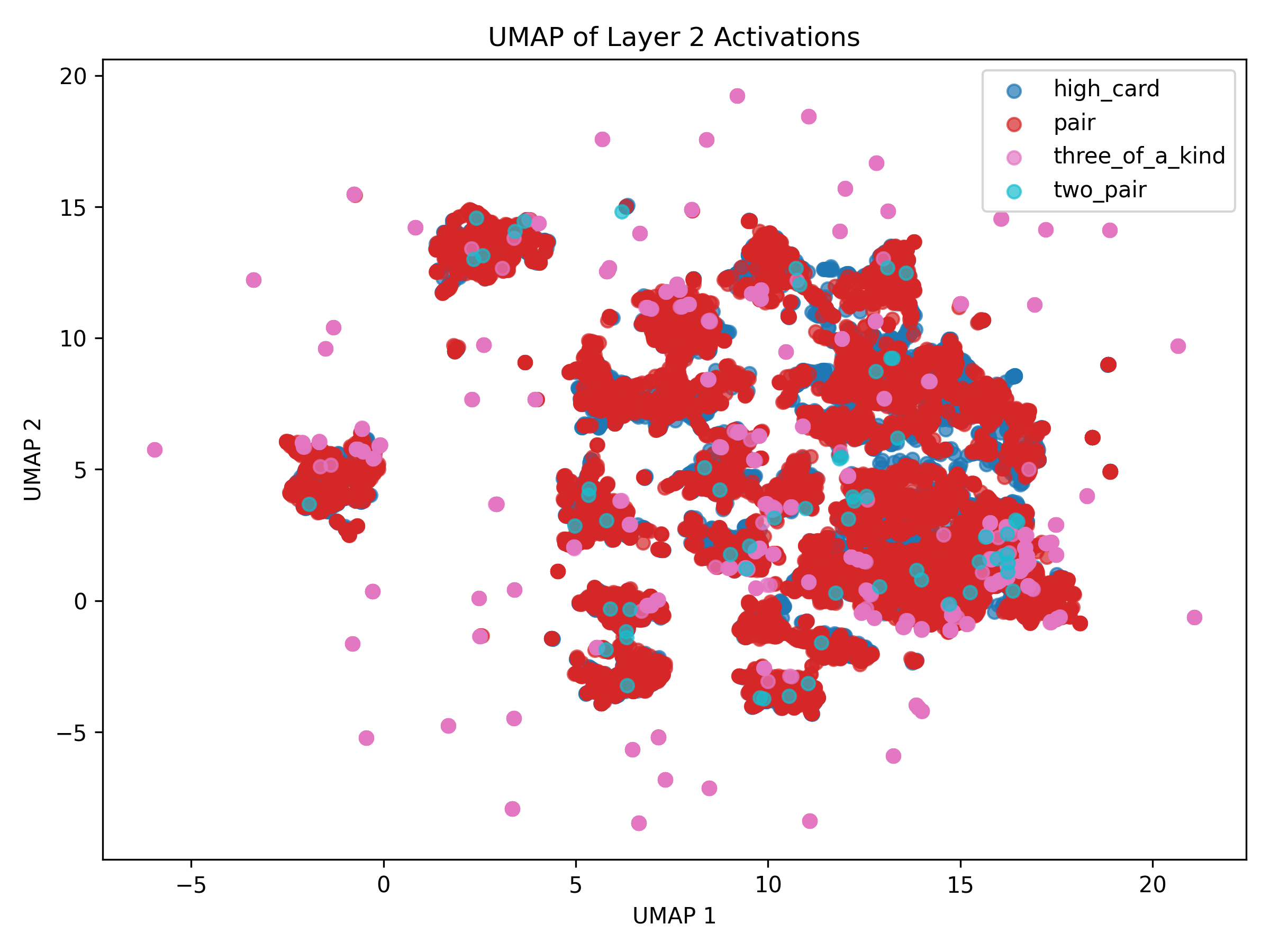}
        \caption{Layer 2}
    \end{subfigure}

    \par\medskip

    \begin{subfigure}[b]{0.24\textwidth}\centering
        \includegraphics[width=\textwidth]{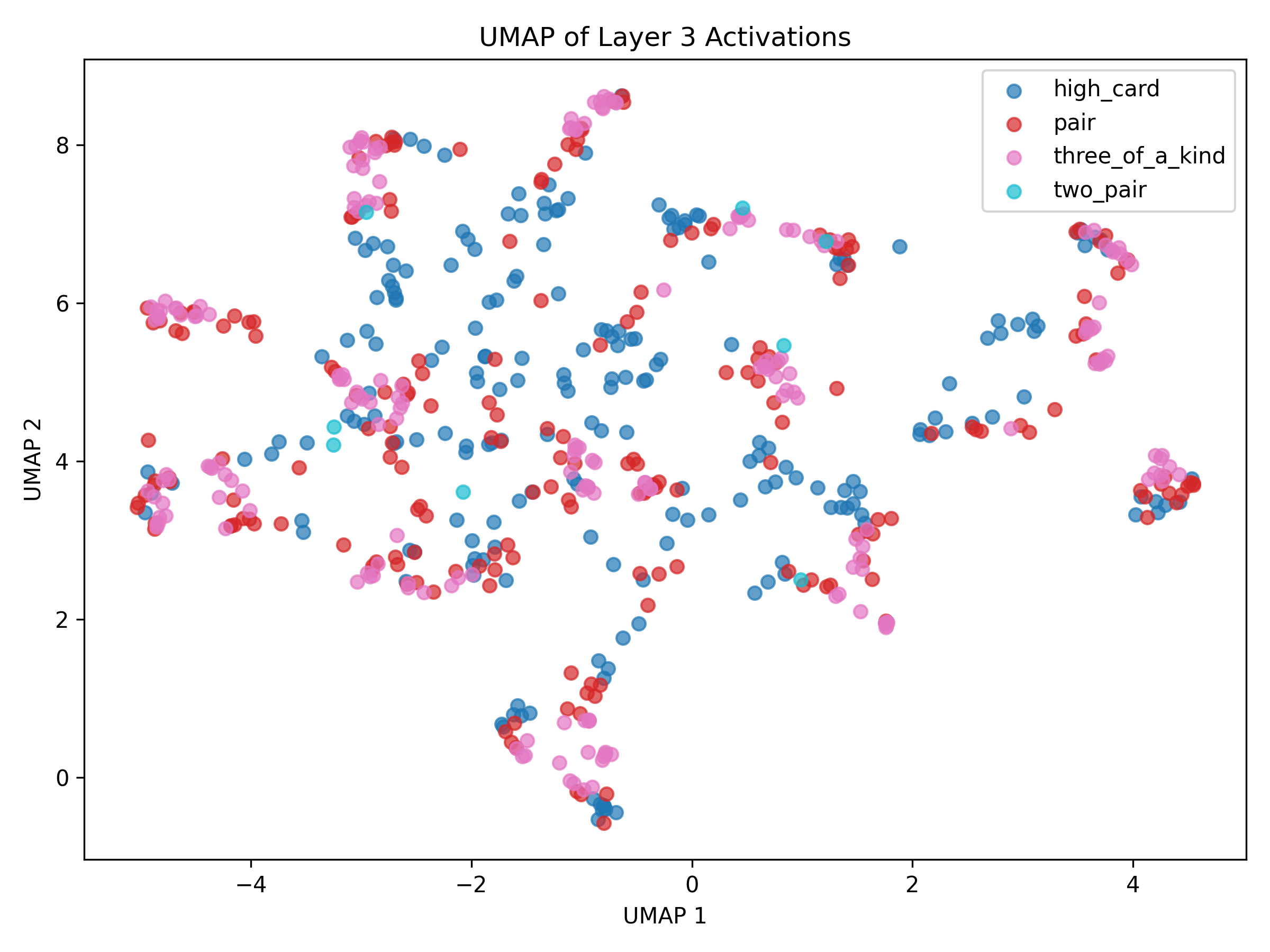}
        \caption{Layer 3}
    \end{subfigure}\hfill
    \begin{subfigure}[b]{0.24\textwidth}\centering
        \includegraphics[width=\textwidth]{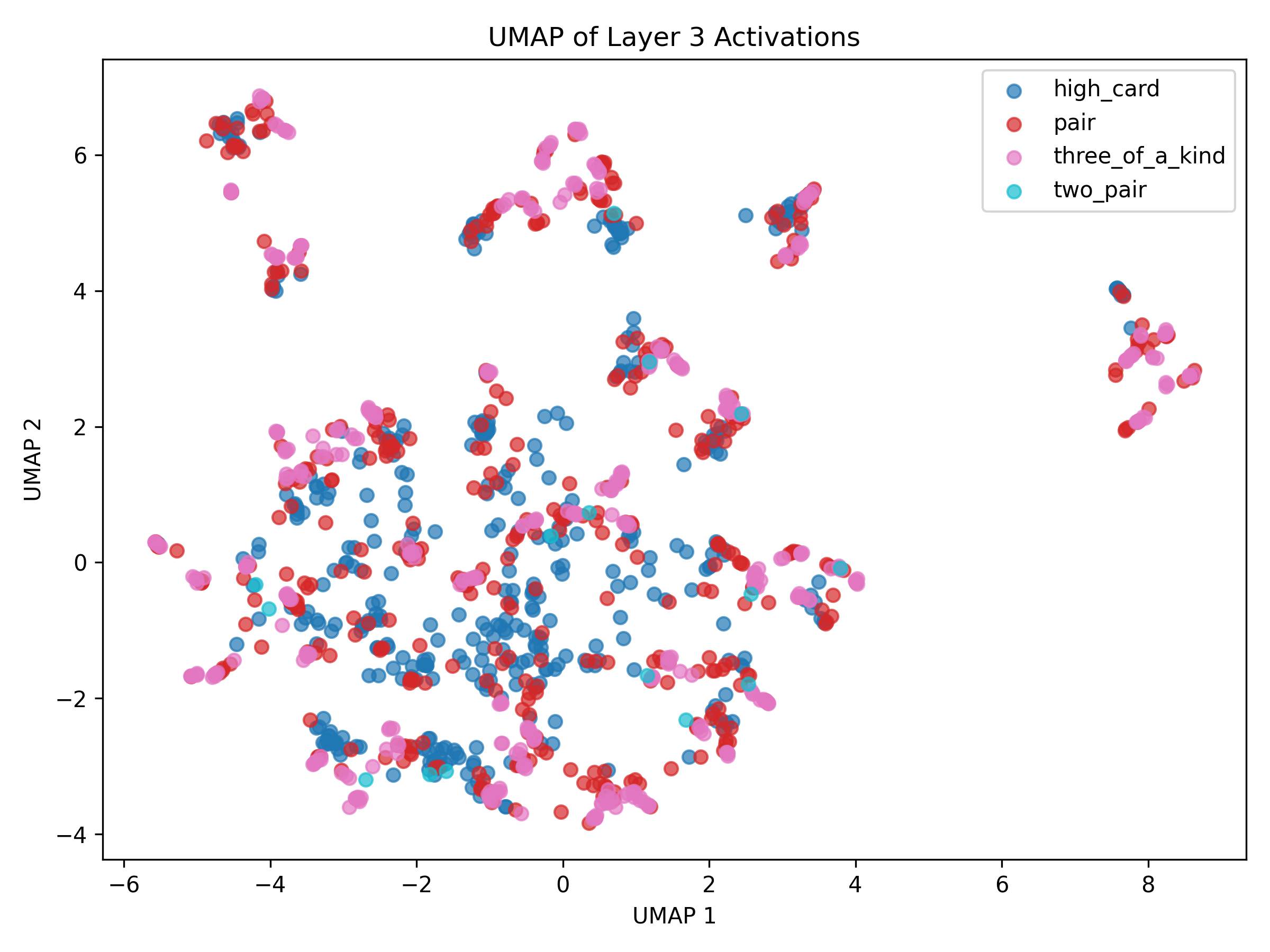}
        \caption{Layer 3}
    \end{subfigure}\hfill
    \begin{subfigure}[b]{0.24\textwidth}\centering
        \includegraphics[width=\textwidth]{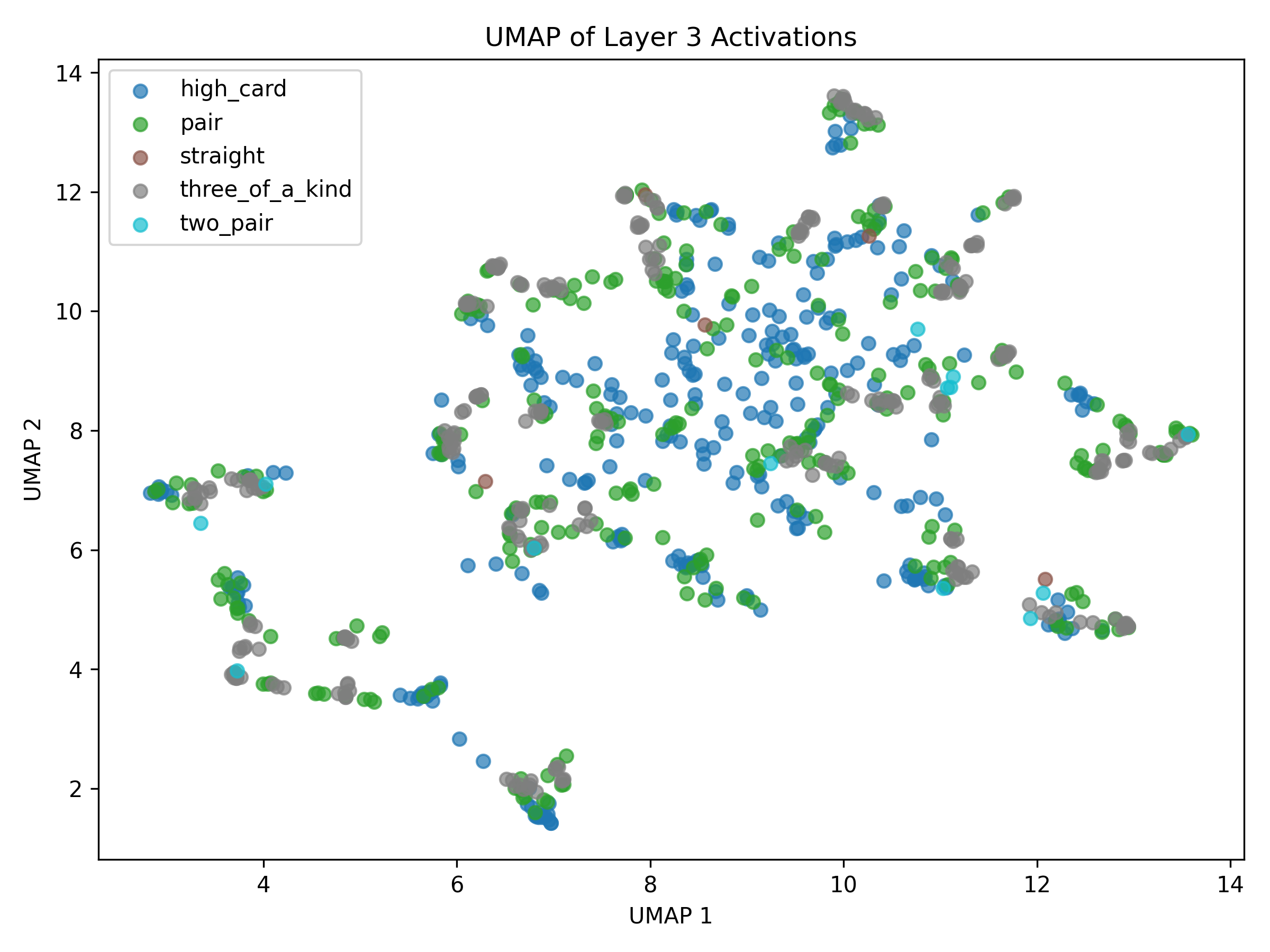}
        \caption{Layer 3}
    \end{subfigure}\hfill
    \begin{subfigure}[b]{0.24\textwidth}\centering
        \includegraphics[width=\textwidth]{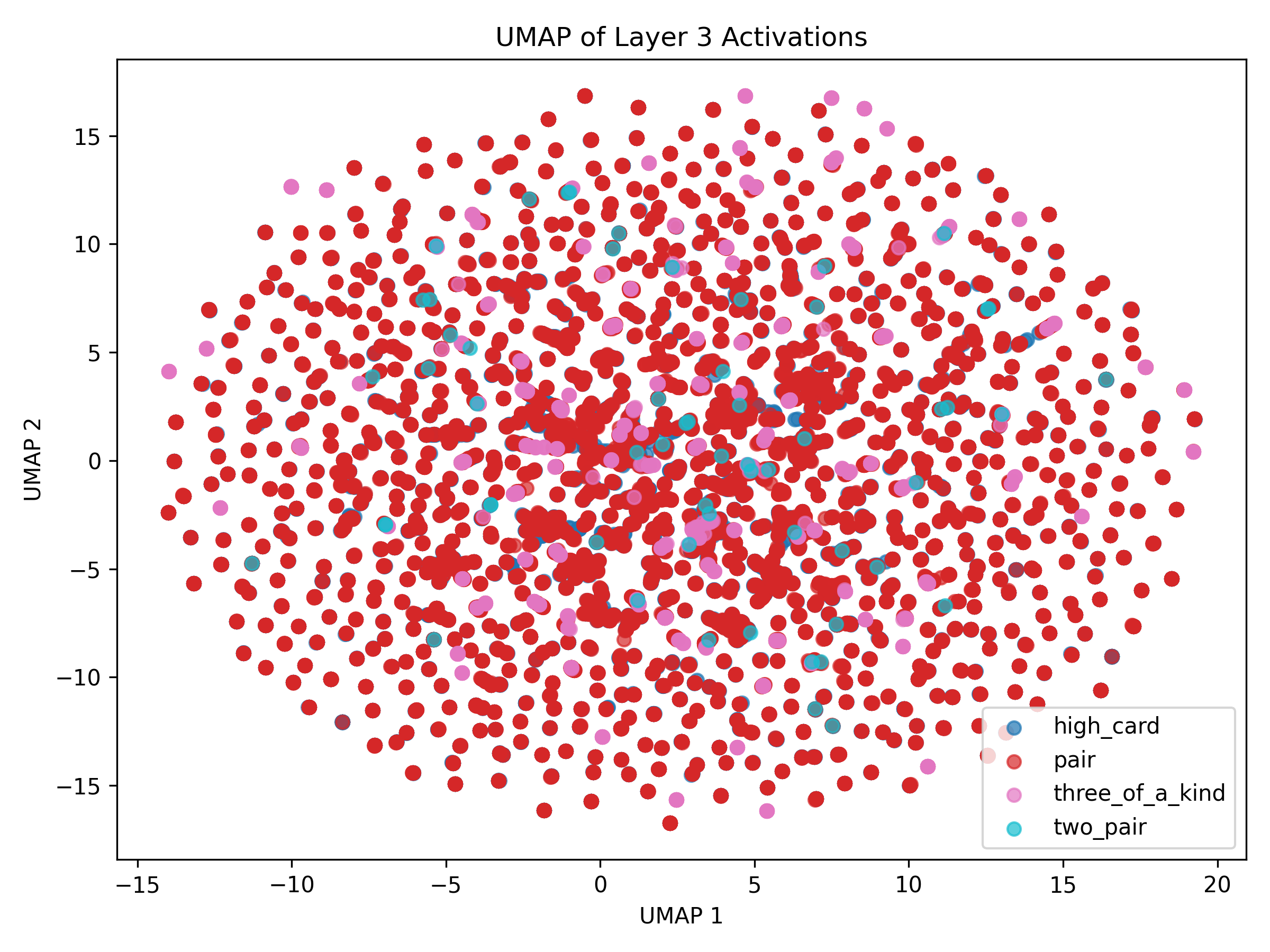}
        \caption{Layer 3}
    \end{subfigure}
    \caption{UMAP projections of activation vectors across transformer Layers 0–3 and
across four training set sizes. Clusters correspond to
semantically related hand-ranks, but the method introduces more distortion in
the results.}
\end{figure}

\end{document}